\documentclass[authoryear]{article}

\usepackage{amssymb}

\usepackage[final,nonatbib]{neurips_2024_modified}
\usepackage[authoryear]{natbib}
\usepackage{amsmath}
\usepackage[flushleft]{threeparttable}

\usepackage[utf8]{inputenc} 
\usepackage[T1]{fontenc}    
\usepackage{hyperref}       
\usepackage{url}            
\usepackage{booktabs}       
\usepackage{amsfonts}       
\usepackage{nicefrac}       
\usepackage{microtype}      
\usepackage{xcolor}         
\usepackage{graphicx}
\usepackage{subcaption}
\usepackage{graphicx}
\usepackage{babel,blindtext}

\usepackage{subcaption}
\usepackage{changepage}
\usepackage{multirow}
\usepackage{array}

\title{Illustrious: an Open Advanced Illustration Model}

\author{%
  Sang Hyun Park\thanks{equal contribution} 
  \quad Jun Young Koh\footnotemark[1]    \quad Junha Lee   \quad Joy Song \\
  \AND
  Dongha Kim \quad Hoyeon Moon \quad Hyunju Lee \quad Min Song\thanks{corresponding author}  \\
  \\
  Onoma AI Research \\
}

\begin{document}
\maketitle

\begin{abstract}
  In this work, we share the insights for achieving state-of-the-art quality in our text-to-image anime image generative model, called Illustrious. To achieve high resolution, dynamic color range images, and high restoration ability, we focus on three critical approaches for model improvement. First, we delve into the significance of the batch size and dropout control, which enables faster learning of controllable token based concept activations. Second, we increase the training resolution of images, affecting the accurate depiction of character anatomy in much higher resolution, extending its generation capability over 20MP with proper methods. Finally, we propose the refined multi-level captions, covering all tags and various natural language captions as a critical factor for model development. Through extensive analysis and experiments, Illustrious demonstrates state-of-the-art performance in terms of animation style, outperforming widely-used models in illustration domains, propelling easier customization and personalization with nature of open source. We plan to publicly release updated Illustrious model series sequentially as well as sustainable plans for improvements on HuggingFace\footnote{https://huggingface.co/OnomaAIResearch/Illustrious-xl-early-release-v0} with a license\footnote{https://huggingface.co/OnomaAIResearch/Illustrious-xl-early-release-v0/blob/main/README.md}.
\end{abstract}

\begin{figure*}[htbp]
    \centering
    \begin{minipage}{0.33\textwidth}
        \includegraphics[width=\textwidth,height=8cm,keepaspectratio]{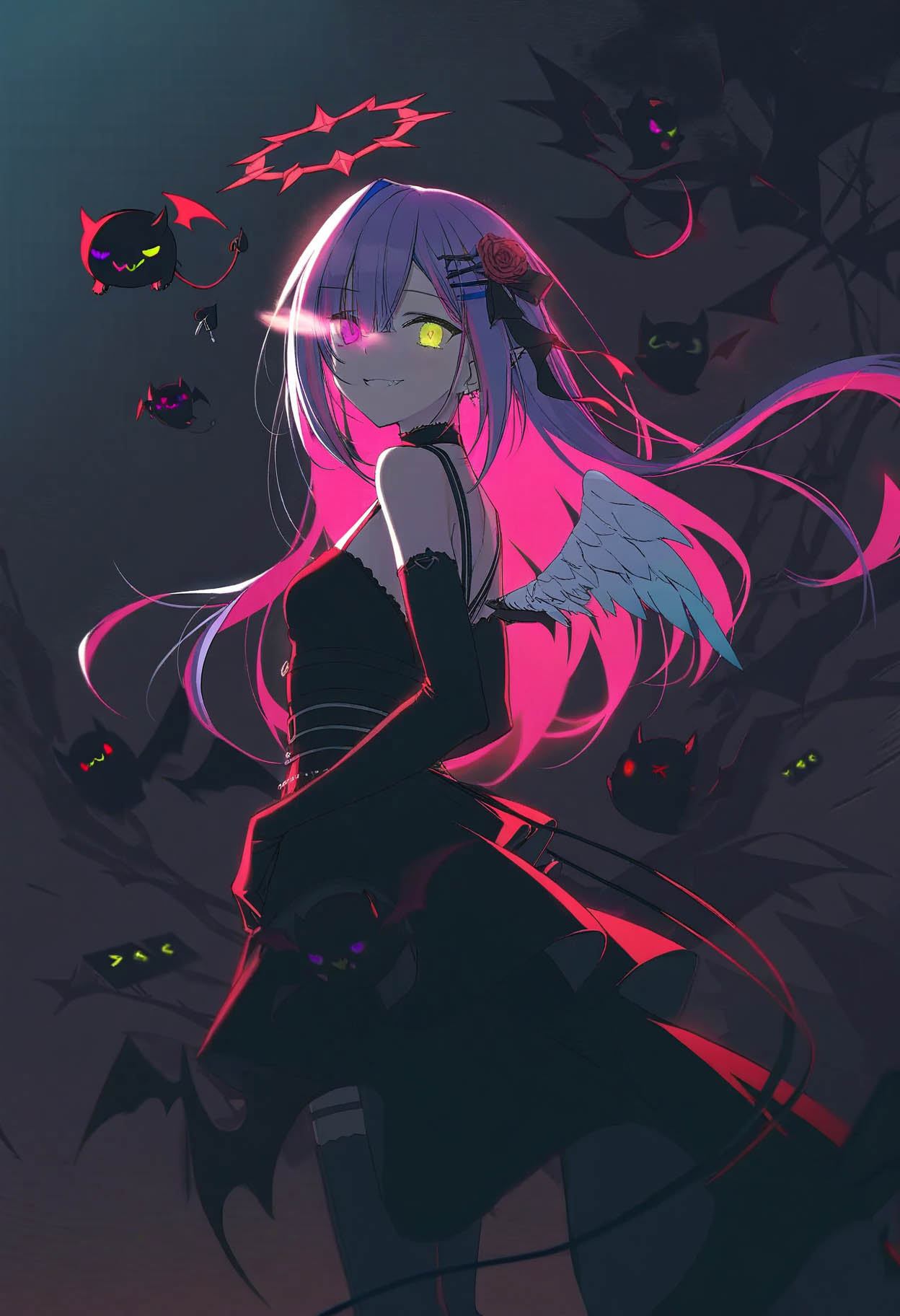}
    \end{minipage}\hfill
    \begin{minipage}{0.33\textwidth}
        \includegraphics[width=\textwidth,height=8cm,keepaspectratio]{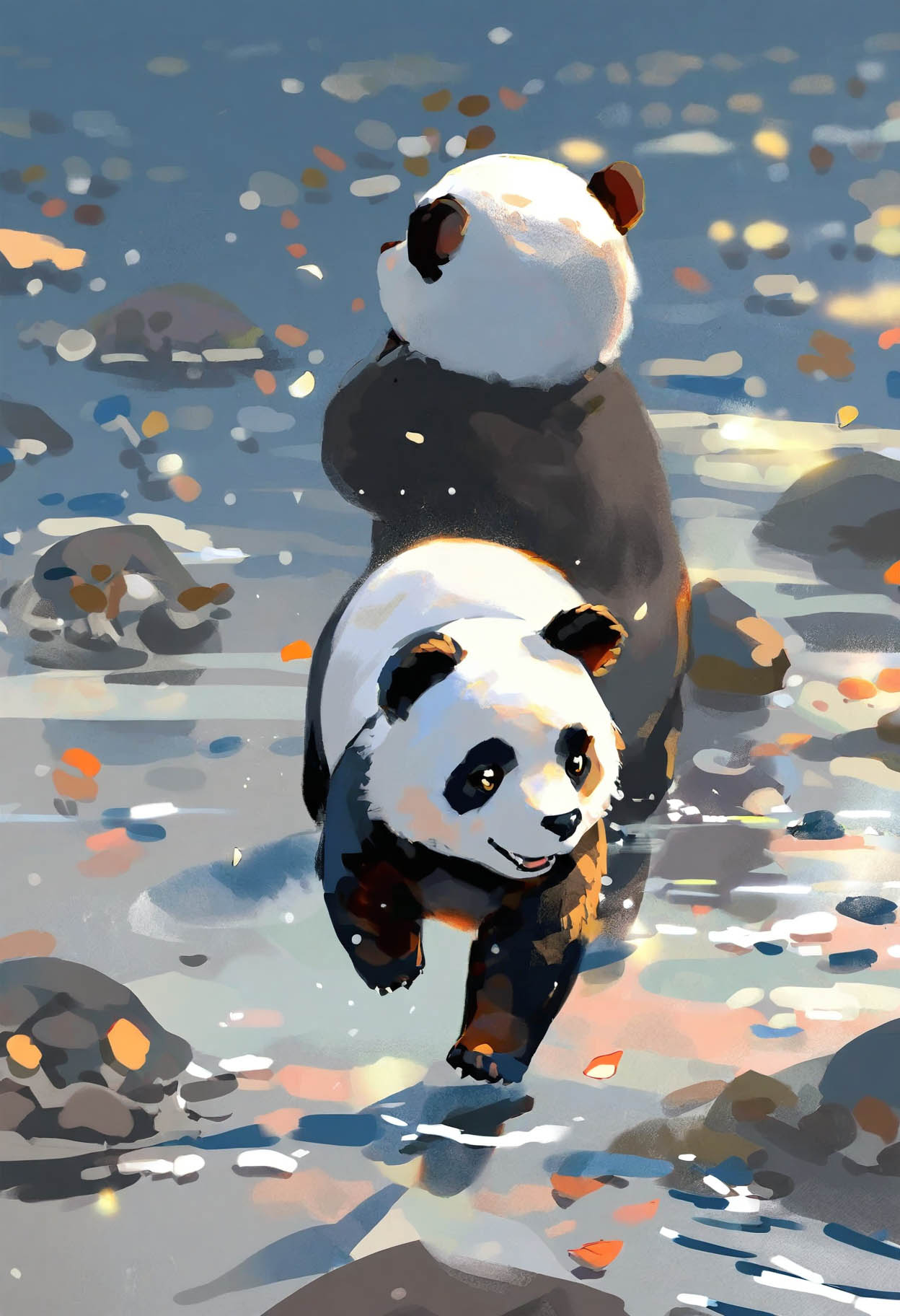}
    \end{minipage}\hfill
    \begin{minipage}{0.33\textwidth}
        \includegraphics[width=\textwidth,height=8cm,keepaspectratio]{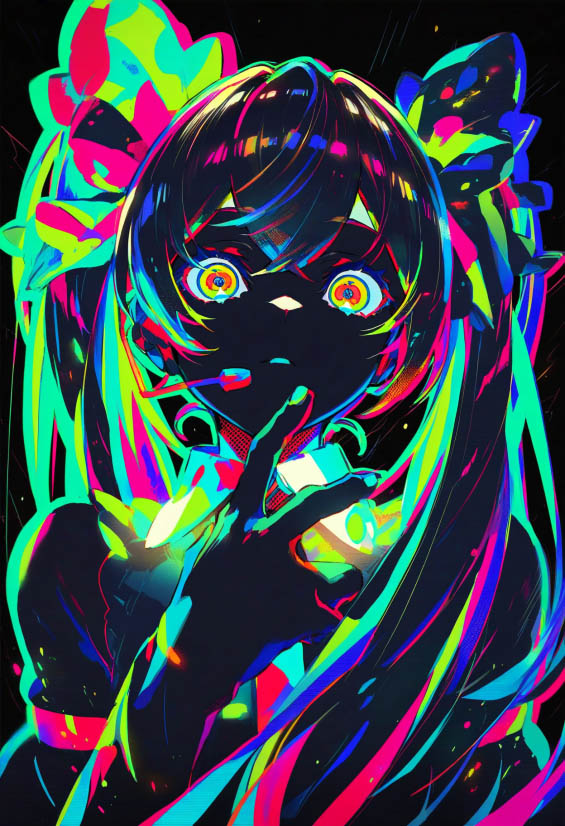}
    \end{minipage}
    
    \begin{minipage}{0.33\textwidth}
        \includegraphics[width=\textwidth,height=8cm,keepaspectratio]{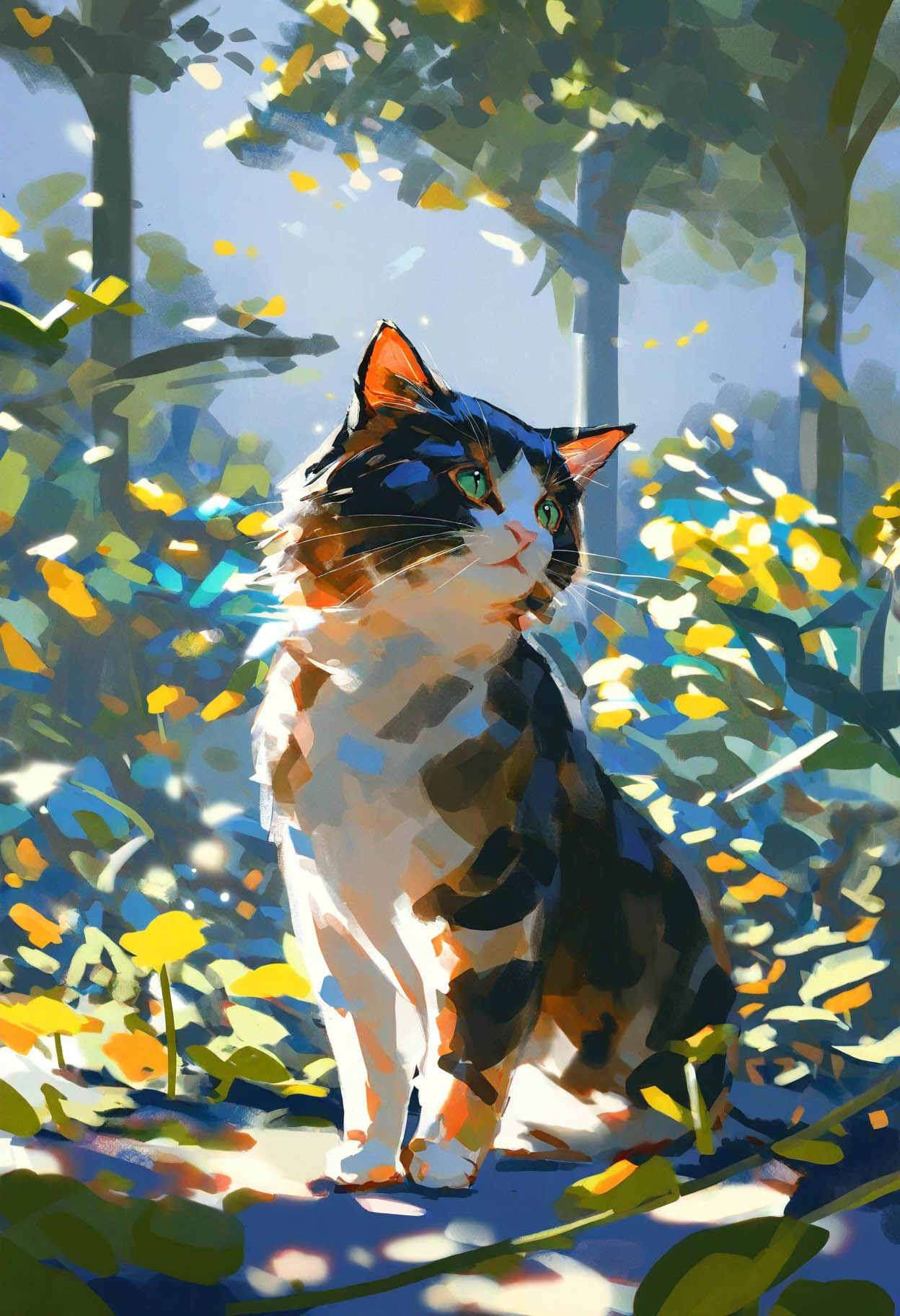}
    \end{minipage}\hfill
    \begin{minipage}{0.33\textwidth}
        \includegraphics[width=\textwidth,height=8cm,keepaspectratio]{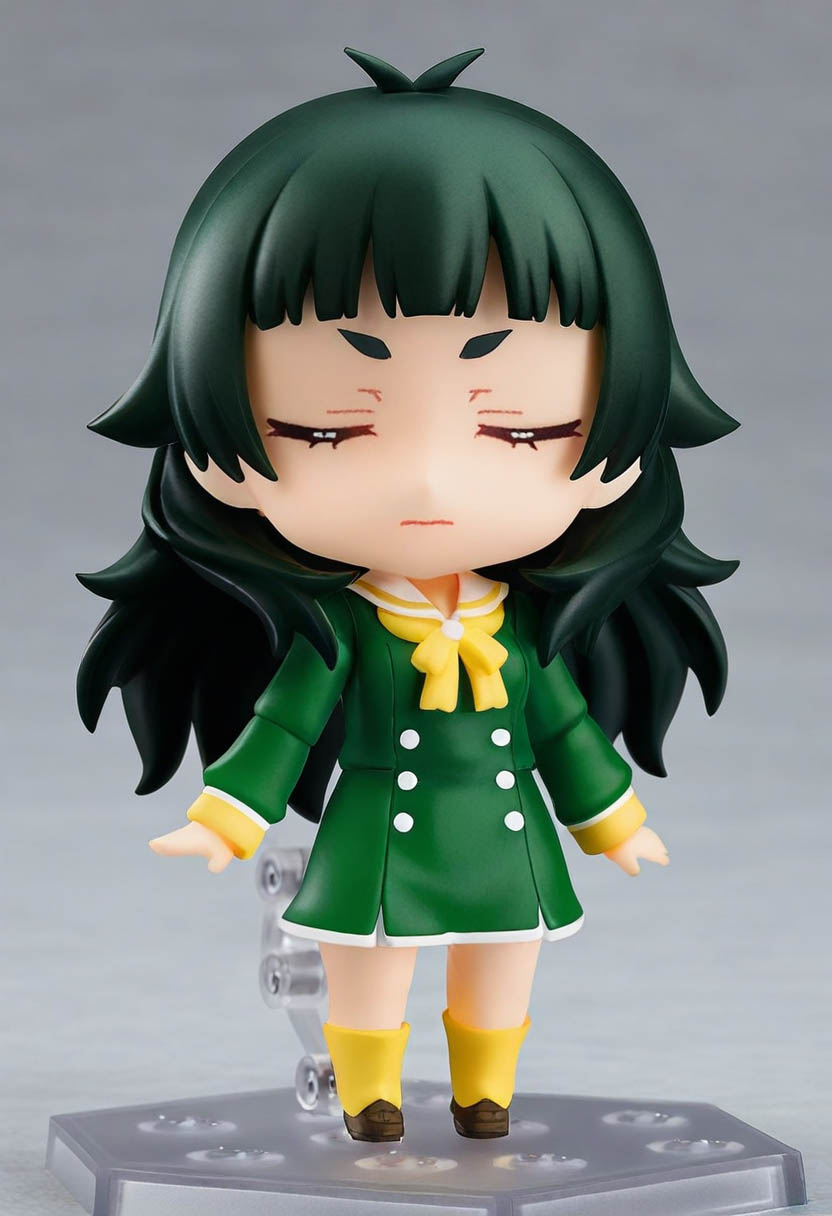}
    \end{minipage}\hfill
    \begin{minipage}{0.33\textwidth}
        \includegraphics[width=\textwidth,height=8cm,keepaspectratio]{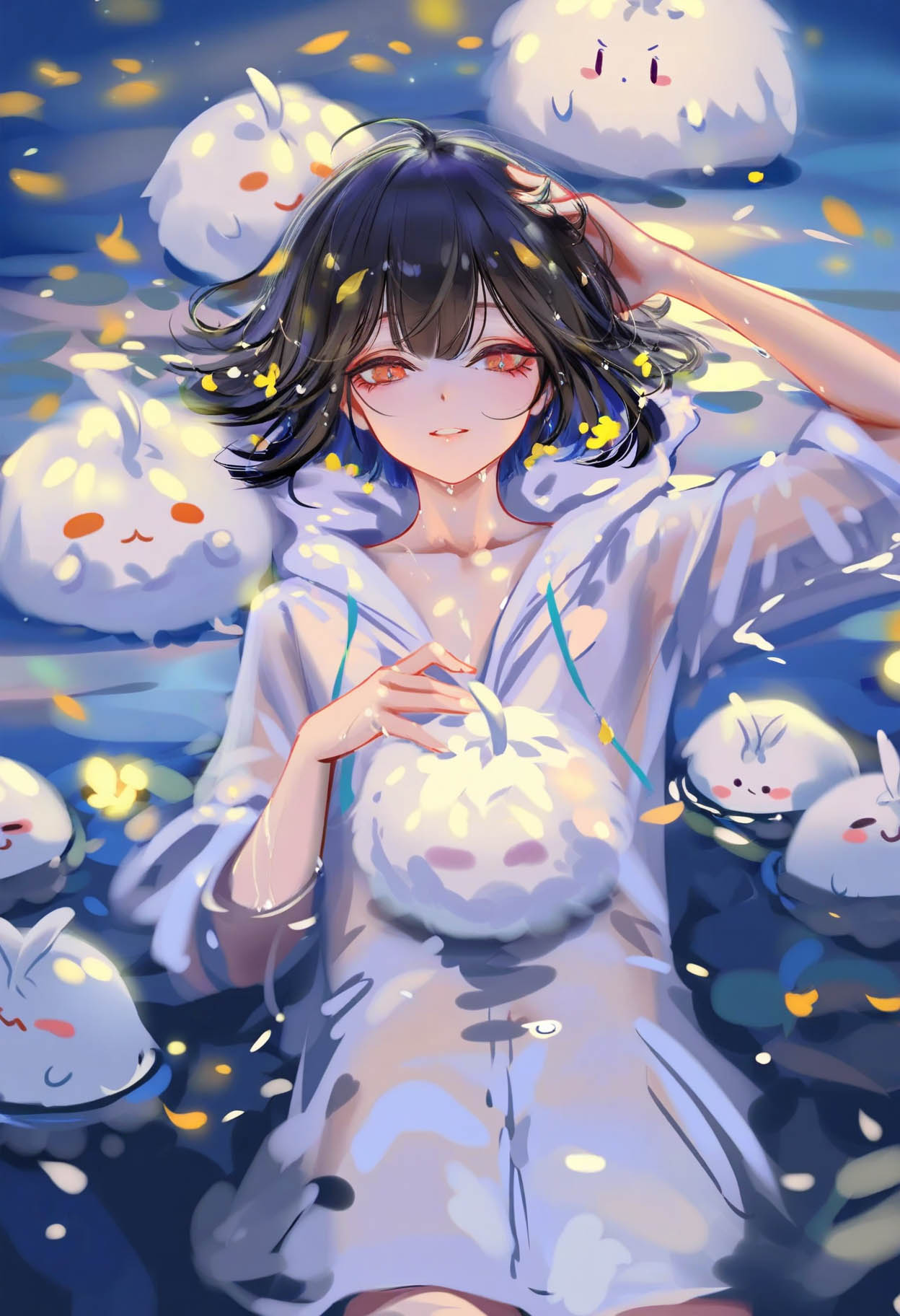}
    \end{minipage}


    \begin{minipage}{0.33\textwidth}
        \includegraphics[width=\textwidth,height=8cm,keepaspectratio]{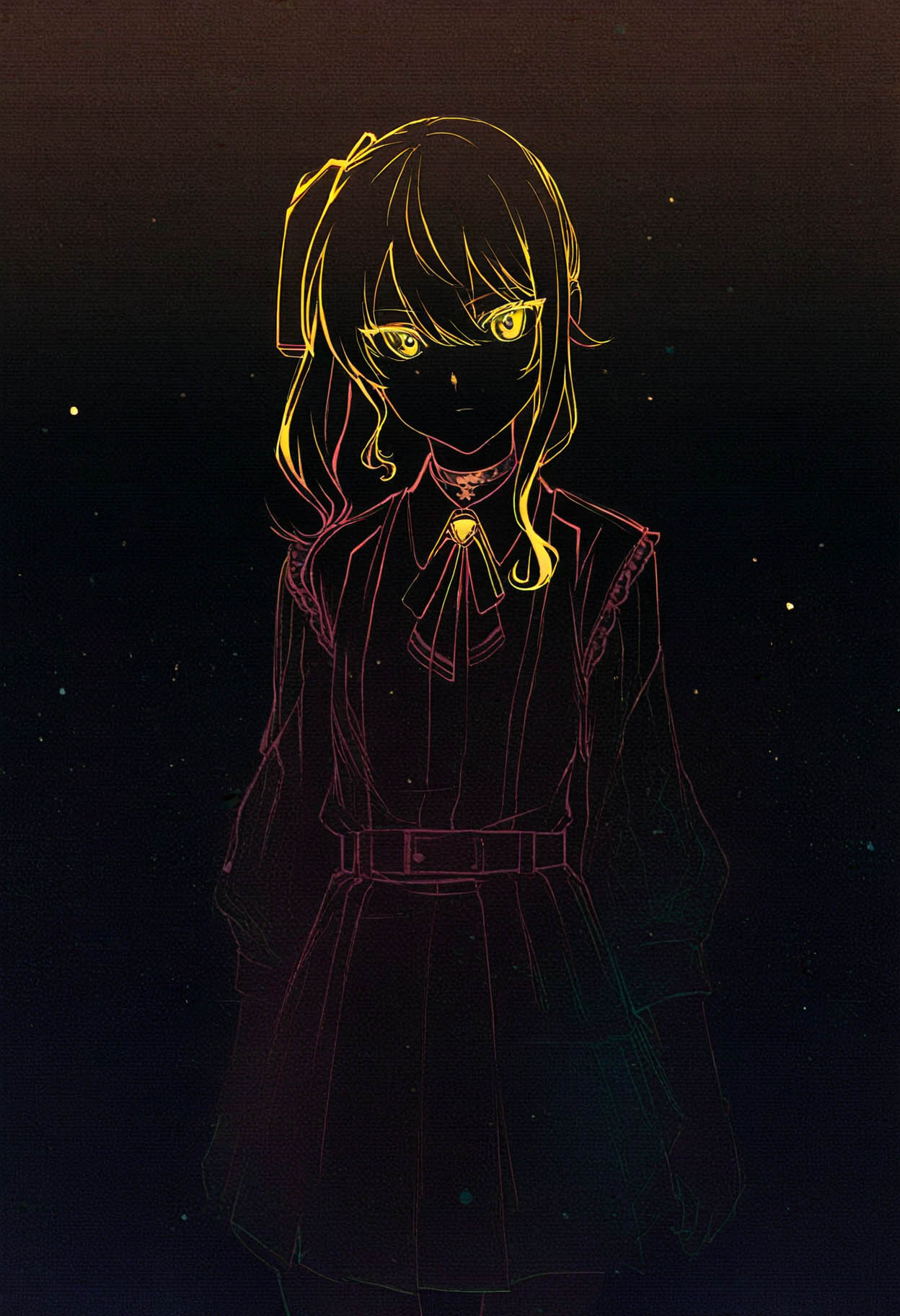}
    \end{minipage}\hfill
    \begin{minipage}{0.33\textwidth}
        \includegraphics[width=\textwidth,height=8cm,keepaspectratio]{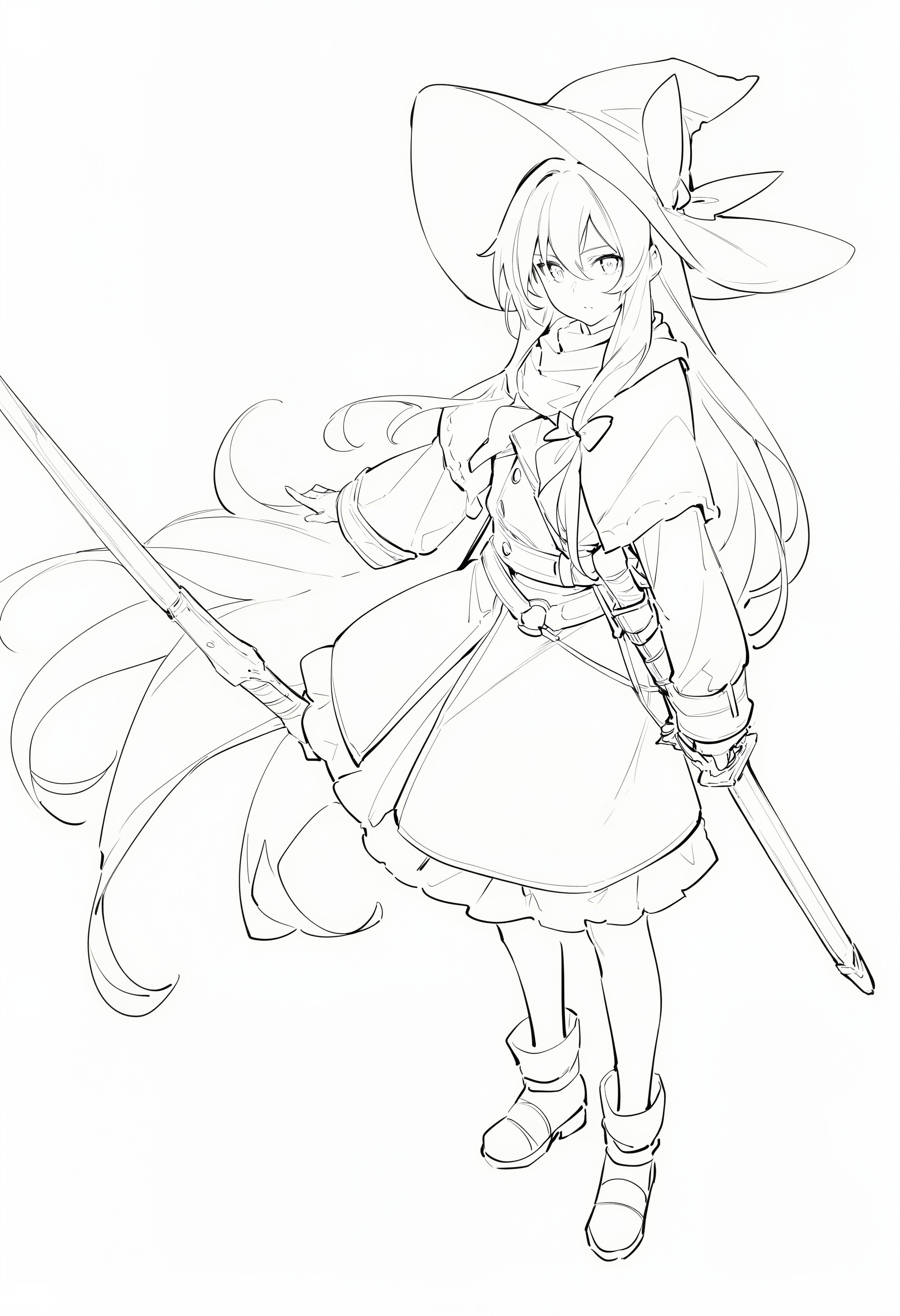}
    \end{minipage}\hfill
    \begin{minipage}{0.33\textwidth}
        \includegraphics[width=\textwidth,height=8cm,keepaspectratio]{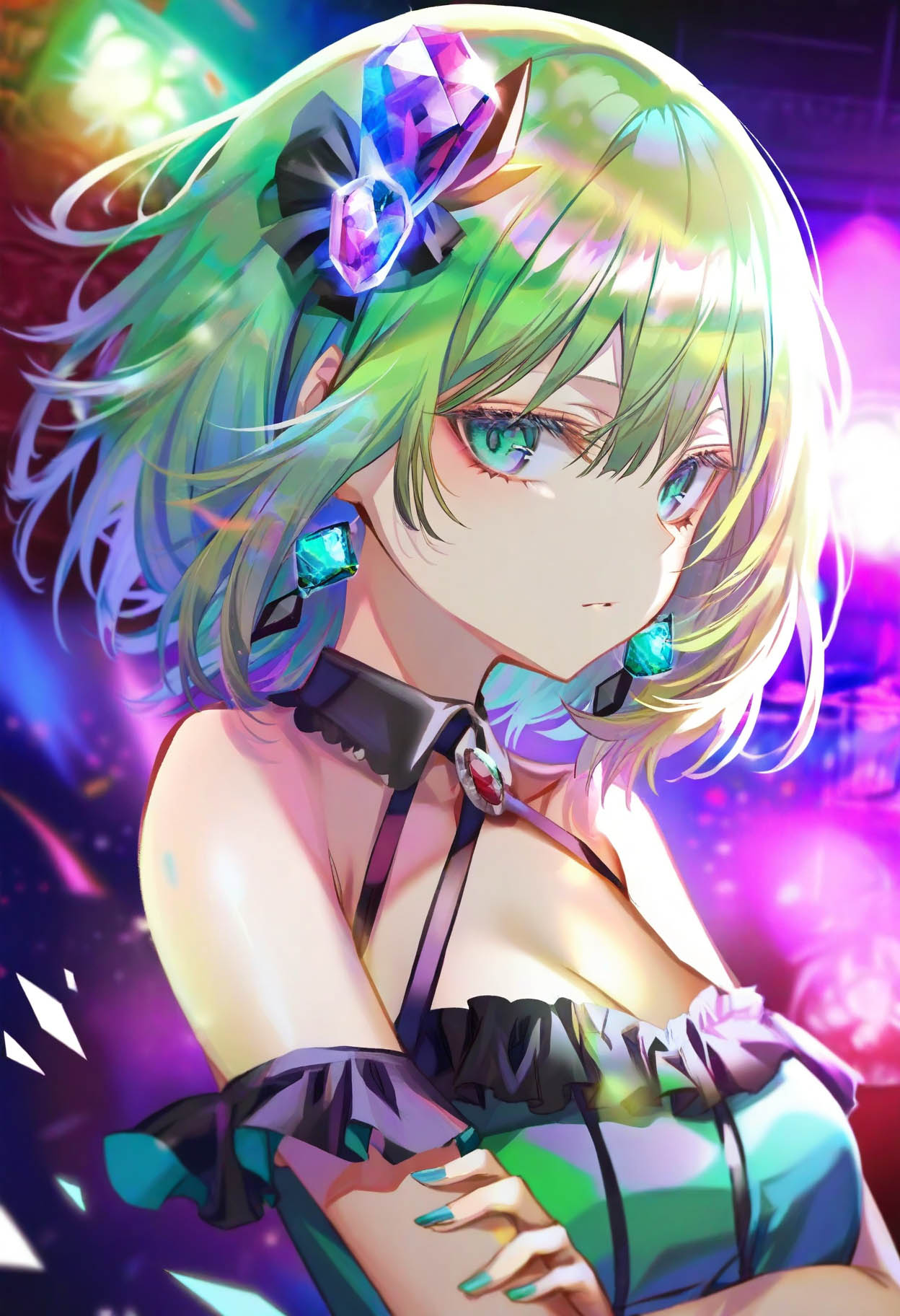}
    \end{minipage}
    \caption{\textbf{High-quality samples from Illustrious.} Our model exhibits vibrant color and contrast on a range of image styles.}
    \label{fig:Samples}
\end{figure*}

\section{Introduction}
Stable Diffusion \cite{SD1.5} has brought groundbreaking advancements to the field of image generation. In particular, SDXL \cite{SDXL}, which was trained with SD XL architecture with dual CLIP text encoder, based on large-scale datasets, has become even more powerful, offering excelling prompt control over generated images. While photorealistic image generation has benefited from large datasets like ImageNet \cite{ImageNet} and OpenImages \cite{openimages}, illustration and animation image generation has comparatively shown slow progress, mainly due to the lack of large-scale finetuned open sourced models and strict dataset requirements.

We introduce a state-of-the-art anime generation model, Illustrious,  which surpasses existing various models in various aspects. By leveraging a large dataset and offering detailed prompt guidance, Illustrious can express a wide range of concepts combinations, as depicted in Figure \ref{fig:multiconcept} that previous models struggled, with accurate control with prompt guidance, such as CFG \cite{CFG}, and capable for producing high-resolution images with anatomical integrity. 

\begin{figure*}
    \centering
    \includegraphics[width=\linewidth]{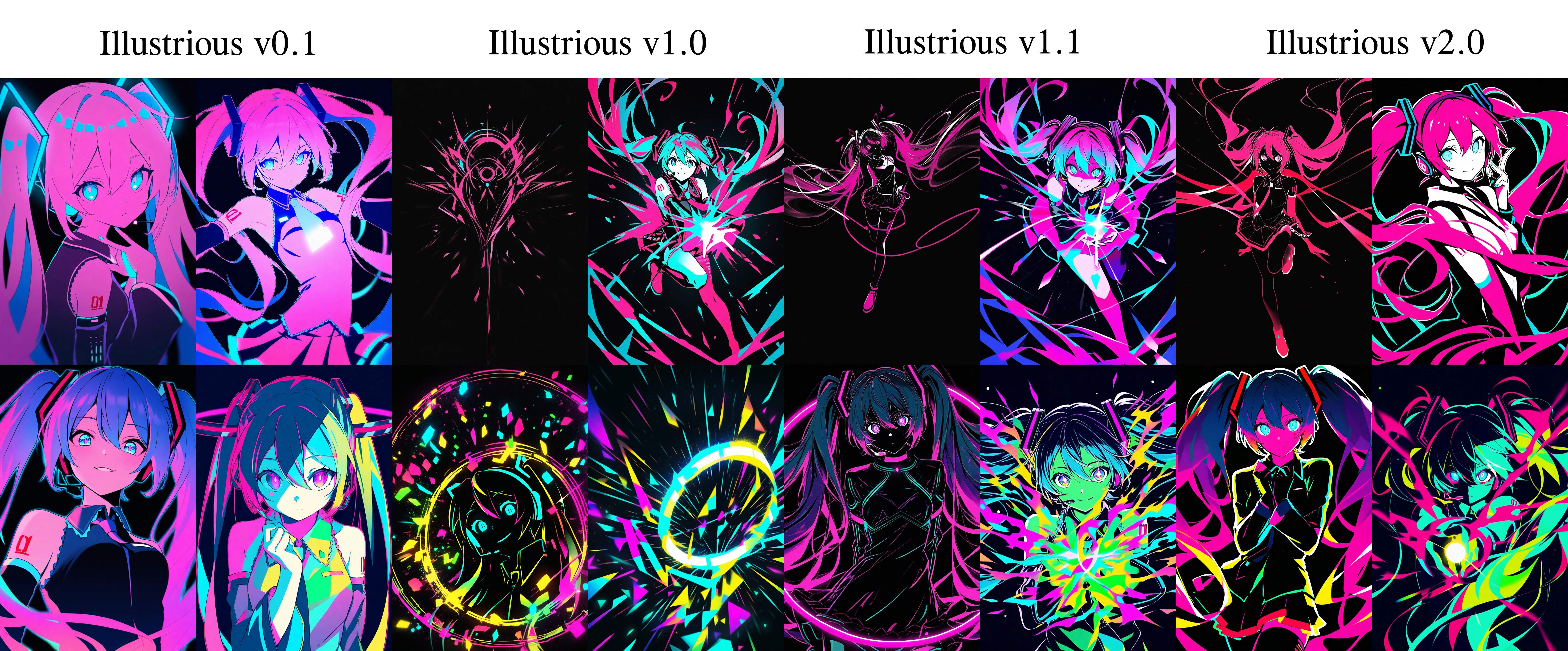}
    \caption{Model Comparison Images}
    \label{fig:image_samples_compare}
\end{figure*}

\section{Preliminary}
\label{preliminary}
\subsection{SDXL}
Stable Diffusion is a latent Text-to-Image diffusion model used as a foundation model in various image domain fields such as classification \cite{diversitydefinitelyneededimproving}, controllable image editing \cite{IPAdapter} \cite{ControlNet}, personalized image generation  \cite{Dreambooth} \cite{CAT} \cite{TextualInversion}, and synthetic data generation \cite{syntheticdatadiffusionmodels} \cite{improvingtextgenerationimages}. According to previous studies by Ho et al. \cite{DDPM} and Song et al. \cite{SDE} \cite{DDIM}, the diffusion model has arose as a powerful image generation model \cite{diffusionbeatsgan}, placing the U-Net \cite{UNet} backbone as a dominant architecture. In addition to this popular U-Net backbone, SD / SDXL applies improved upscaling layers, and cross-attention for text-to-image synthesis to a Transformer-based architecture. Unlike SD1.5 and SD2.0, which uses CLIP ViT-L, OpenCLIP ViT-H respectively, SDXL employs dual text encoders: CLIP ViT-L and OpenCLIP ViT-bigG. With the addition of a second text encoder, SDXL has significantly improved its understanding of text descriptions for images compared to previous models. The change resulted in the parameter size of the text encoder of 817M and 2.6B parameters in the U-Net.

\begin{table*}[htbp]
\caption{Finetuned Model}
\begin{center}
\resizebox{\textwidth}{!}{ 
\begin{tabular}{|cc|c|c|c|c|c|c|}
\toprule
\multicolumn{1}{c|}{\textbf{Finetuned Model}}  & \textbf{Base Model}  & \textbf{step} &  \textbf{batch size} & \textbf{Dataset Size} & \textbf{Prompt Style} &  \textbf {Annotation Method} & \textbf{Resolution}   \\ 
\midrule
\multicolumn{1}{c|}{Animagine XL V3.1} & SDXL 1.0 &  91,030 & 96 & 2.1M  &  Tag  based & Original Prompt  & 1024 x 1024 \\
\multicolumn{1}{c|}{Kohaku XL Delta} & SDXL 1.0 &  28,638 & 128 & 3.6M &  Tag based & Original Prompt & 1024 x 1024 \\
\multicolumn{1}{c|}{Kohaku XL Zeta} & SDXL 1.0 &  16,548 & 128 & 8.4M &  Tag based & Original Prompt & 1024 x 1024 \\
\multicolumn{1}{c|}{SanaeXL anime V1.0} & SDXL 1.0 &  - & - & 7.8M &  Tag based & Original Prompt & 1024 x 1024 \\
\multicolumn{1}{c|}{Neta Art XL} & SDXL 1.0 & - & - &  -  & Tag based & Original Prompt + CogVLM  \cite{CogVLM} + WaifuTagger &  1024 x 1024 \\ 
\multicolumn{1}{c|}{Arti Waifu Diffusion 2.0} & SDXL 1.0 & - & - &  2.5M & Tag based & Original Prompt + Tag Ordering &  1024 x 1024 \\ 
\bottomrule
\multicolumn{1}{c|}{Illustrious v0.1} & SDXL 1.0 & 781,250  & 192 & 7.5M  & Tag based  & Original Prompt + Reorganized / Manual Filtering & 1024 x 1024 \\ 
\multicolumn{1}{c|}{Illustrious v1.0} & SDXL 1.0 & 625,000  & 128 & 10M  & Tag based  & Original Prompt + Reorganized / Manual Filtering & \textbf{1536 x 1536} \\ 
\multicolumn{1}{c|}{Illustrious v1.1} & SDXL 1.0 & 93,750  & 512 & 12M  & Tag based  & Multi-level Captions & \textbf{1536 x 1536} \\ 
\multicolumn{1}{c|}{Illustrious v2.0} & SDXL 1.0 & 78,125  & 512 & 20M  & Tag based  & Multi-level Captions & \textbf{1536 x 1536} \\ 
\bottomrule
\end{tabular}
} 
\label{tab2:finetunedmodel}
\end{center}
\end{table*} 

\subsection{Illustration / Animation Domain}
Danbooru dataset  \cite{danbooru} \cite{danbooru2023} is a public large-scale anime image dataset with over 8 million images contributed and annotated in detail by communities. Annotation of images covers aspects such as characters, scenes, copyrights, and artists. Along with the Danbooru dataset, most available datasets are either processed versions of the Danbooru dataset \cite{DAF} or face datasets \cite{AnimeCeleb} \cite{CartoonFaceRecognition} used for benchmarking purposes. We note that open sourced Illustrious model variants are being released under a research focused, non-commercial / no-closed source derivative public license, solely for open-source progresses. 

\subsection{Next-generation Text-to-Image Generative Models}
With the advancement of AI technology in recent years, AI-based generative models have attracted a great amount of attention within the illustration field. In particular, next-generation models such as Hunyuan-DiT \cite{Hunyuan-DiT}, Kolors \cite{kolors}, Stable Diffusion 3 (SD 3) \cite{SD3}, Flux \cite{blackforestlabs2024flux}, and AuraFlow \cite{fal2024auraflow} utilize additional as well as alternative text encoders to correctly interpret natural language input from users, increasing the sophistication of their ability to generate various, correct compositions of visual content.

\subsubsection{Features of next-generation models}
Hunyuan-DiT is a text-to-image diffusion transformer with a fine-grained understanding of both English and Chinese. It has redesigned the transformer structure, text encoder, and positional encoding. The model supports multi-turn, multi-modal dialogue with users, allowing it to generate and refine images based on contextual input.
Another text-to-image generation model, Kolors, uses GLM \cite{GLM}, instead of T5 to improve comprehension of captions in order to improve the performance of natural language processing. Kolors uses the U-Net architecture and improves performance through a two-stage learning strategy: conceptual learning and learning for quality improvement.
SD3 trains a rectified flow model by enhancing existing noise sampling techniques. This approach has demonstrated superior performance compared to traditional diffusion methods in high-resolution text-to-image synthesis.
Flux is based on a hybrid architecture of multi-modal and parallel diffusion transformer \cite{scalableTransformer} blocks, scaled to 12B parameters, with various technologies \cite{flowmatching} \cite{Roformer} \cite{ScalingViT}.
AuraFlow replaced the MMDiT block with a large DiT encoder block to improve model performance and computational efficiency. Performance was improved by using zero-shot LR transitions, and all data was re-captioned to reduce noise in the dataset.

\subsubsection{Text Encoder}
Currently, the text encoder seemly plays a crucial role in text-to-image generative models. A commonly used text encoder in generative models is CLIP \cite{CLIP}. OpenCLIP \cite{OpenCLIP} provides various versions of CLIP. Despite existence of various CLIP model variants, trained in various datasets \cite{WIT}  \cite{LAION-5B} \cite{DataComp}, the CLIP-only model has not shown significant success on complex compositions and glyph generations.
For instance, SD1.5 and DALL-E2 \cite{DALLE2} use CLIP as their text encoder, however possibly due to limitation of CLIP itself proposed in various researches, \cite{limitationCLIP1} \cite{limitationCLIP2} \cite{lycoris}, it is unknown whether SD XL architecture is fundamentally limited in complex compositions, and glyph generations.

One valid solution has been proposed by various models such as Imagen \cite{IMAGEN}, PixArt  \cite{Pixart-alpha}, eDiFF-I \cite{ediff-i} Hunyuan-DiT \cite{Hunyuan-DiT}, Auraflow, and Flux. Through the utilization of the Transformer T5 \cite{T5}, this  solution enables delivering more fine-grained local information to  their text encoder. Stable Diffusion 3 \cite{SD3} also demonstrated the potential to interpret and generate complex prompts using the T5-XXL model. Remarkably, the CLIP-escaping architectures, like Kolors \cite{kolors}, which use GLM \cite{GLM} has noted CLIP-dependent architecture as significant cause of limitation.

The Illustrious model is built upon SD XL architecture without changes, may share the noted limitations.

\subsection{Data Ethics}

Text-to-image diffusion models are often trained under the pretext of 'aesthetic' considerations. However, this practice sometimes involves unethical data usage, such as obscuring the names of the artists whose works are used in training, thereby enabling the generation of specific styles without crediting the original artists. We believe it is crucial not to exploit or distort the data, even if this leads to a model with a default style that may appear dull or unclear.

To ensure ethical use of data, it is essential to clearly distinguish styles by associating them with the names of the artists and making this information transparent. Moreover, to safeguard artists from potential exploitation, we recommend that any transformative use of data and model to be accompanied by clear specification of training methodologies, modifications, and other relevant details, under fair public AI license terms \cite{fpai2024}.
\begin{table*}[htbp]
\caption{Baseline Model}
\begin{center}
\begin{threeparttable}
\resizebox{\textwidth}{!}{ 
\begin{tabular}{|cc|c|c|c|c|c}
\toprule
\multicolumn{1}{c|}{\textbf{Model}}  & \textbf{Parameter Size}    &   \textbf{Dataset} & \textbf{Resolution} &  \textbf{Domain} & \textbf{Prompt Style}  & \textbf{Accessibility}  \\ 
\midrule
\multicolumn{1}{c|}{Stability AI Stable Diffusion 1.5} & 980M & LAION & 512 $\times$ 512 & Arbitrary & Natural Language & Open Source   \\
\multicolumn{1}{c|}{Stability AI Stable Cascade} &  1.4B   & - & 1024 $\times$ 1024 & Arbitrary  & Natural Language & Open Source \\
\multicolumn{1}{c|}{Stability AI Stable Diffusion XL} &   2.5B   & - & 1024 $\times$ 1024 & Arbitrary & Natural Language  & Open Source        \\
\multicolumn{1}{c|}{Stability AI Stable Diffusion 3} &   2B, 8B   & ImageNet, CC12M \cite{CC12M} & 1024 $\times$ 1024 & Arbitrary & Natural Language  & Open Source \tnote{a}        \\
\multicolumn{1}{c|}{Midjourney V4} & - & COCO \cite{COCO}, Visual Genome \cite{VisualGenome}, Flickr 30K \cite{Flickr30k} & 1024 $\times$ 1024 & Arbitrary & Natural Language & Closed Source        \\
\multicolumn{1}{c|}{OpenAI DALLE-3} & -  & LAION   & 1024 $\times$ 1024 &  Arbitrary & - & Closed Source \\
\multicolumn{1}{c|}{Hunyuan DiT} & 1.5B & - & - & Arbitrary & Natural Language &  Open Source       \\
\multicolumn{1}{c|}{Playground V3.0 \cite{PGv3}} & 24B &  - & 1024 $\times$ 1024 & Arbitrary & Natural Language &  Closed Source       \\
\multicolumn{1}{c|}{Flux} & 12B &  - & 2.0MP & Arbitrary & Natural Language &  Open Source       \\
\multicolumn{1}{c|}{Novel AI Image Generator \cite{NAI3}} & - &  Danbooru & - & Illustrate Picture & Tag based &  Closed Source       \\
\bottomrule
\multicolumn{1}{c|}{Illustrious} & 2.5B & Danbooru, Synthetic datasets* &  \bf{2048 $\times$ 2048} & Illustrate Picture & Tag based\tnote{b} &  Open Source\tnote{c}  \\ 
\bottomrule

\end{tabular}
}
\label{table:comparison}
\begin{tablenotes}
\scriptsize 
\item[a] SD3 model variants are currently separated source
\item[b] Illustrious datasets and prompt styles vary by version
\item[c] Illustrious model variants are currently separated source
\end{tablenotes}
\end{threeparttable}
\end{center}
\end{table*}

\begin{figure}[htb]
    \centering
    \begin{subfigure}[b]{0.28\textwidth}
        \centering
        \includegraphics[width=\textwidth]{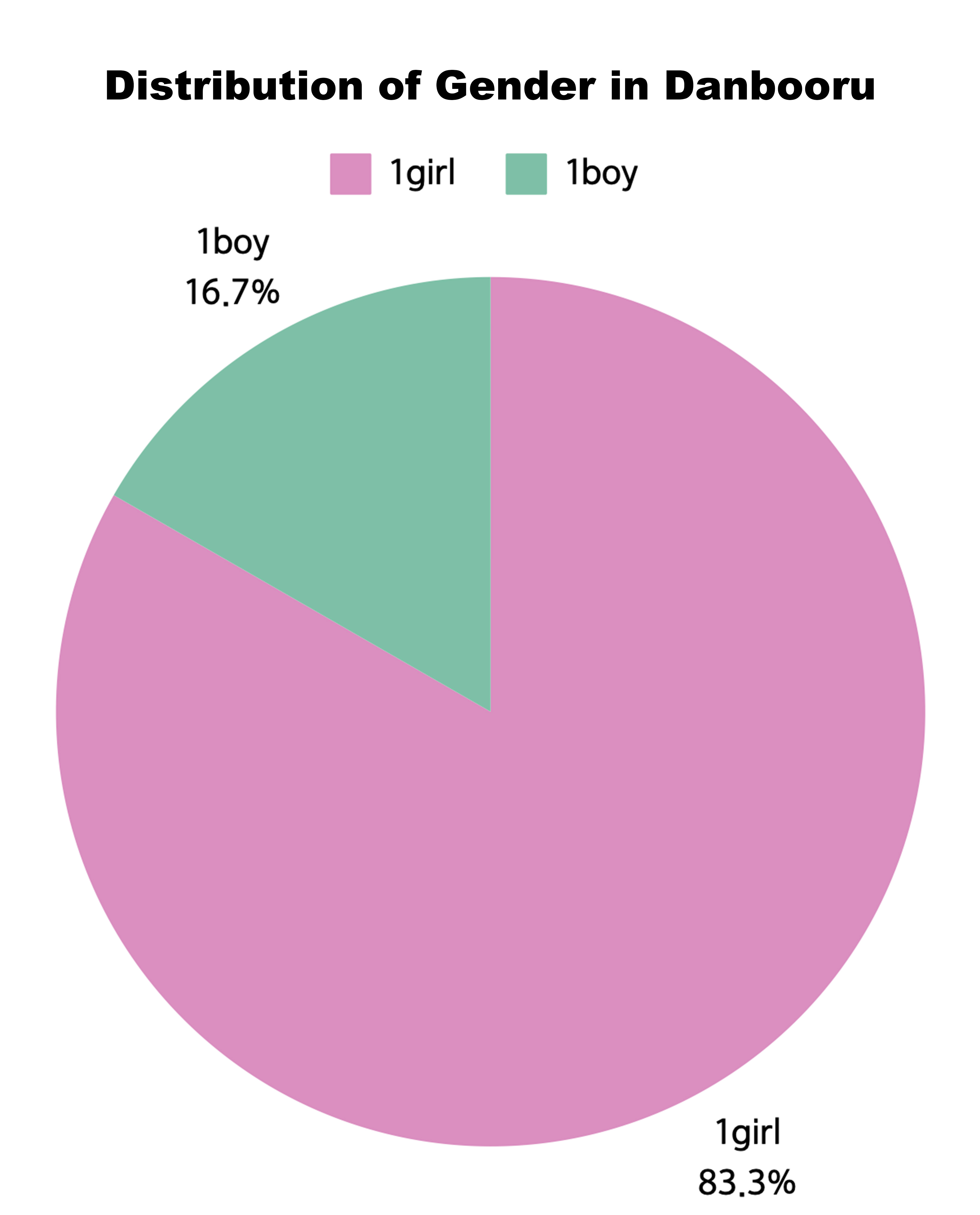}
        \caption{\scriptsize \textbf{Gender distribution of Danbooru dataset.}}
        \label{fig:gender_distribution}
    \end{subfigure}
    \hspace{1em} 
    \begin{subfigure}[b]{0.28\textwidth}
        \centering
        \includegraphics[width=\textwidth]{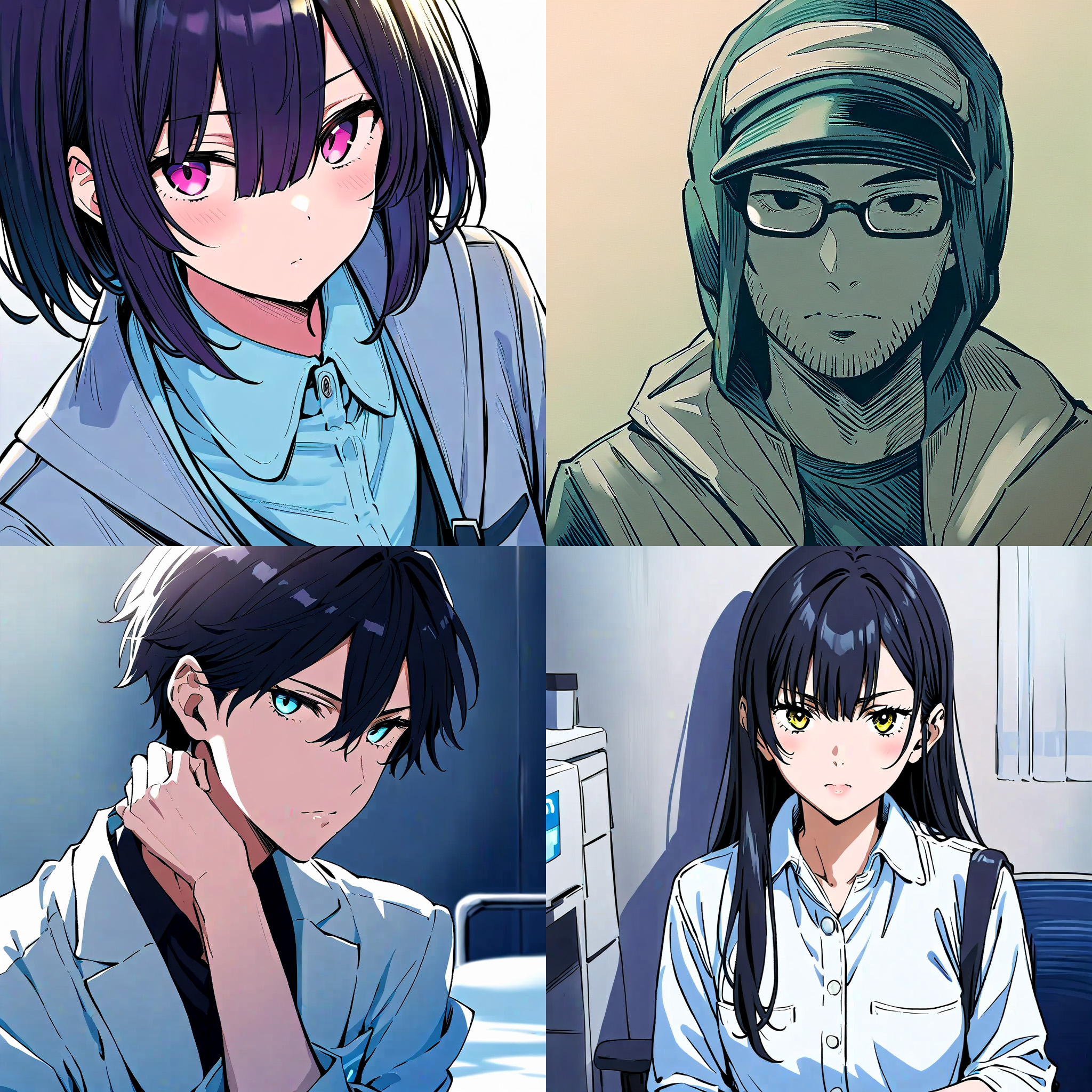}
        \caption{\scriptsize \textbf{Example of "doctor" and "boy", showing model's weak understanding.}}
        \label{fig:doctorproblem}
    \end{subfigure}
    \hspace{1em} 
    \begin{subfigure}[b]{0.28\textwidth}
        \centering
        \includegraphics[width=\textwidth]{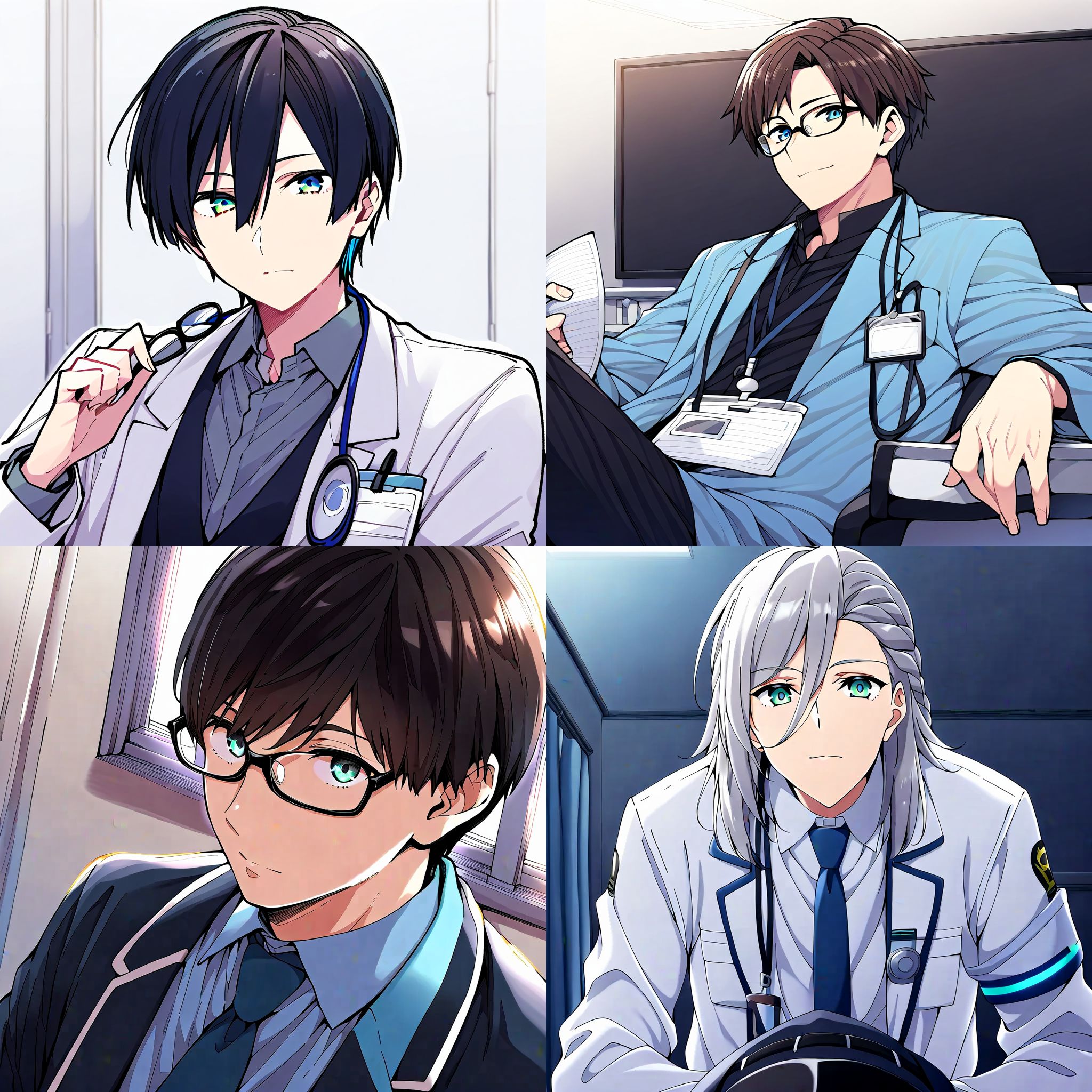}
        \caption{\scriptsize \textbf{The resolved understanding of "doctor" and "boy" in future epochs.}}
        \label{fig:doctorproblem2}
    \end{subfigure}
    
    \caption{\footnotesize Comparison of gender distribution and example generations from the model showing bias and weak understanding of gender-specific terms. The used prompt was \textbf{"1boy, doctor, masterpiece, looking at viewer"}.}
    \label{fig:combined_figure}
\end{figure}

\section{Methodology}
\label{others}
\subsection{Dataset}
\subsubsection{Dataset Bias}

Danbooru dataset contains a noticeably larger representation of female characters compared to male characters. This imbalance mirrors broader trends in the source material including anime and manga, where female characters are often more prominently featured in the form of images and character designs. Such gender imbalance in anime and manga datasets can lead to biased model performance, with models trained on this dataset potentially performing better on tasks involving female characters while underperforming on tasks related to male characters or other underrepresented categories, as shown in Figure \ref{fig:gender_distribution}. This disproportionate representation can hinder the model's generalizability and faireness across different character types. We observed significant discrepancies in v0.1 model, which was later solved by \textbf{removing unfocused annotations in datasets}.

The dataset presents various issues due to its tag-based structure. Oftentimes, multiple meanings overlap with the same tokens or are used interchangeably, leading to confusion and ambiguity. A prominent example is the token "doctor," which can refer to both a character and a profession. In this case, one concept completely overlaps with the other, as shown in Figure \ref{fig:doctorproblem}.
Despite the fact that some images feature multiple characters, many in the dataset have very few tags or lack detailed annotations. This sparsity can make it difficult for models to learn critical concepts, since important features or attributes of the image may not be captured.
The dataset contains extremely high-resolution images that could not be properly downsampled using any existent methods, leading to poor concept comprehension by the model. The Illustrious v0.1 model initially struggled with issues related to absurd aspect ratio, extremely high resolution images, and comic-like datasets. Therefore, careful pruning and refocusing of the dataset is necessary.

Based on insights and analysis on the v0.1 model, we expanded the dataset by including synthetic dataset based on generated images and captions to resolve the issues shown in Figure \ref{fig:doctorproblem2}.
\subsubsection{Data Preprocessing}

We initially adopted the tag ordering approach developed by NovelAI Team\footnote{https://docs.novelai.net/image/tags.html}, which we believe that it functions as an instruction-tuning mechanism. Tags were separated and reordered following a specific schema:
\\ \\
person count ||| character names ||| rating ||| general tags ||| artist ||| score range based rating ||| year modifier
\\ \\
In the v0.1 model training, we split the tags using the "," convention, later occasionally replacing it with spaces based on a certain probability, combined with natural language prompts. Over time, we observed that the score range varied both temporally and across rating categories. To address this, we employed a percentile-based moving window method to determine the score range.

The score criteria and the year modifiers are defined numerically and range-based, depending on post counts:

\begin{table*}[htbp]
\centering
\begin{minipage}{0.45\linewidth}
\centering
\caption{Score Criteria}
\begin{threeparttable}
\begin{tabular}{|cc}
\toprule
\multicolumn{1}{c|}{\textbf{Score Criteria}}  & \textbf{Percentage}  \\ 
\midrule
\multicolumn{1}{c|}{Worst quality} & \textasciitilde 8\% \\
\multicolumn{1}{c|}{Bad quality} & \textasciitilde 20\% \\
\multicolumn{1}{c|}{Average quality} & \textasciitilde 60\% \\
\multicolumn{1}{c|}{Good quality} & \textasciitilde 82\% \\
\multicolumn{1}{c|}{Best quality} & \textasciitilde 92\% \\
\multicolumn{1}{c|}{Masterpiece} & \textasciitilde 100\% \\
\bottomrule
\end{tabular}
\label{table:score_criteria}
\end{threeparttable}
\end{minipage}
\hspace{0.05\linewidth} 
\begin{minipage}{0.45\linewidth}
\centering
\caption{Year Modifier}
\begin{threeparttable}
\begin{tabular}{|cc}
\toprule
\multicolumn{1}{c|}{\textbf{Tag}}  & \textbf{Year}  \\ 
\midrule
\multicolumn{1}{c|}{Oldest} & \textasciitilde 2017 \\
\multicolumn{1}{c|}{Old} & \textasciitilde 2019 \\
\multicolumn{1}{c|}{Modern} & \textasciitilde 2020 \\
\multicolumn{1}{c|}{Recent} & \textasciitilde 2022 \\
\multicolumn{1}{c|}{Newest} & \textasciitilde 2023 \\
\bottomrule
\end{tabular}
\label{table:year_modifier}
\end{threeparttable}
\end{minipage}
\end{table*}

In subsequent epochs, we slightly modified the shuffling behavior by introducing aesthetic modifiers and filtering based on aesthetic scoring and file compression size metrics. The details of these aesthetic modifiers will be disclosed in future model releases.

For images used in the dataset, when the image size exceeded 4MP, we employed a mixed NEAREST/LANCZOS resizing method to maintain the aspect ratio. Images smaller than 768 $\times$ 768 were pruned from the dataset. 
Notably, few extremely high resolution images, \textasciitilde 40MP and those with uncommon aspect ratios (>1:10) were also removed.

However, the significant amount of high-resolution data remain problematic during the resize process, regardless of the resize method. Thus, we limited the higher-resolution dataset with minimal resizing, which is used for high-resolution training from v1.0 training. This allows for native high resolution generation, while minimizing down-sampling artifacts in smaller resolution.

\begin{figure}[htb]
    \centering
    \begin{subfigure}[b]{0.45\textwidth}
        \centering
        \includegraphics[width=\textwidth]{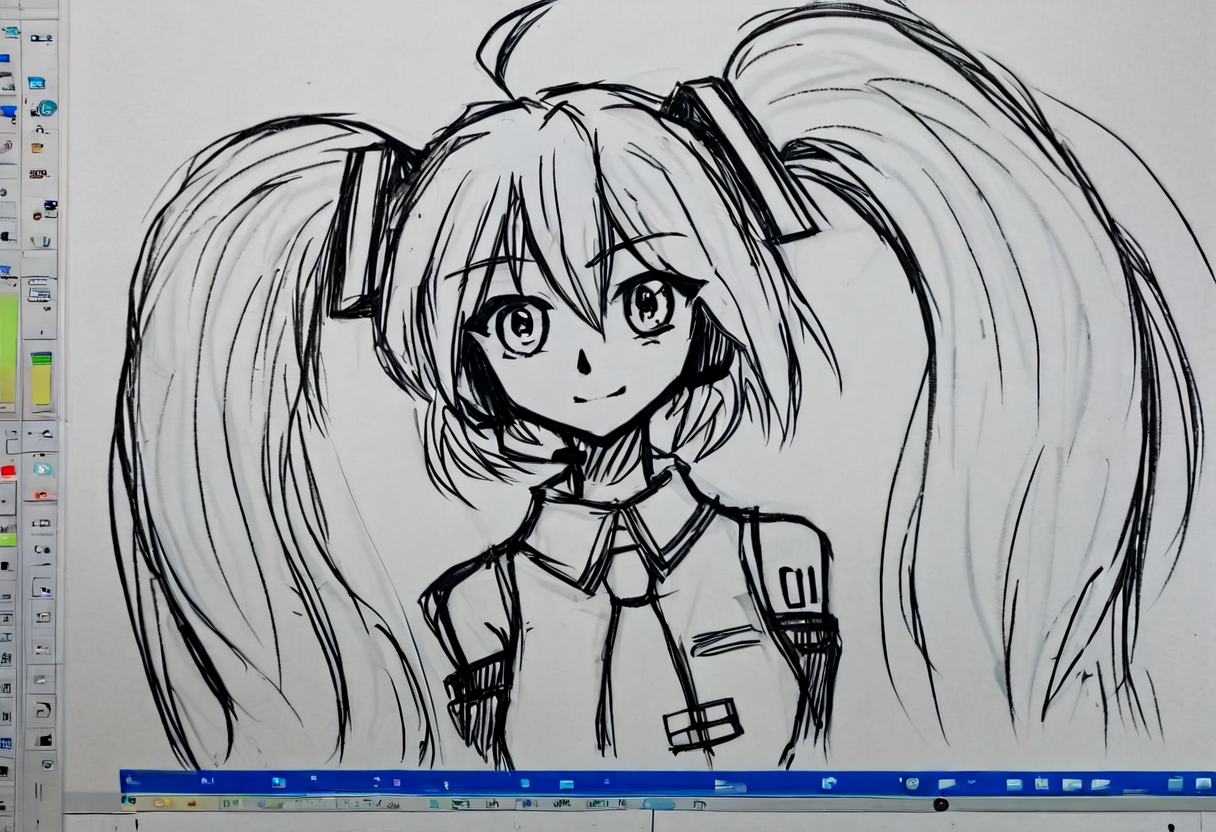}
        \caption{\scriptsize Intentional low-quality generation, with prompt \textbf{1girl, hatsune miku, worst quality, ms paint (medium)}.}
    \end{subfigure}
    \hspace{1em} 
    \begin{subfigure}[b]{0.45\textwidth}
        \centering
        \includegraphics[width=\textwidth]{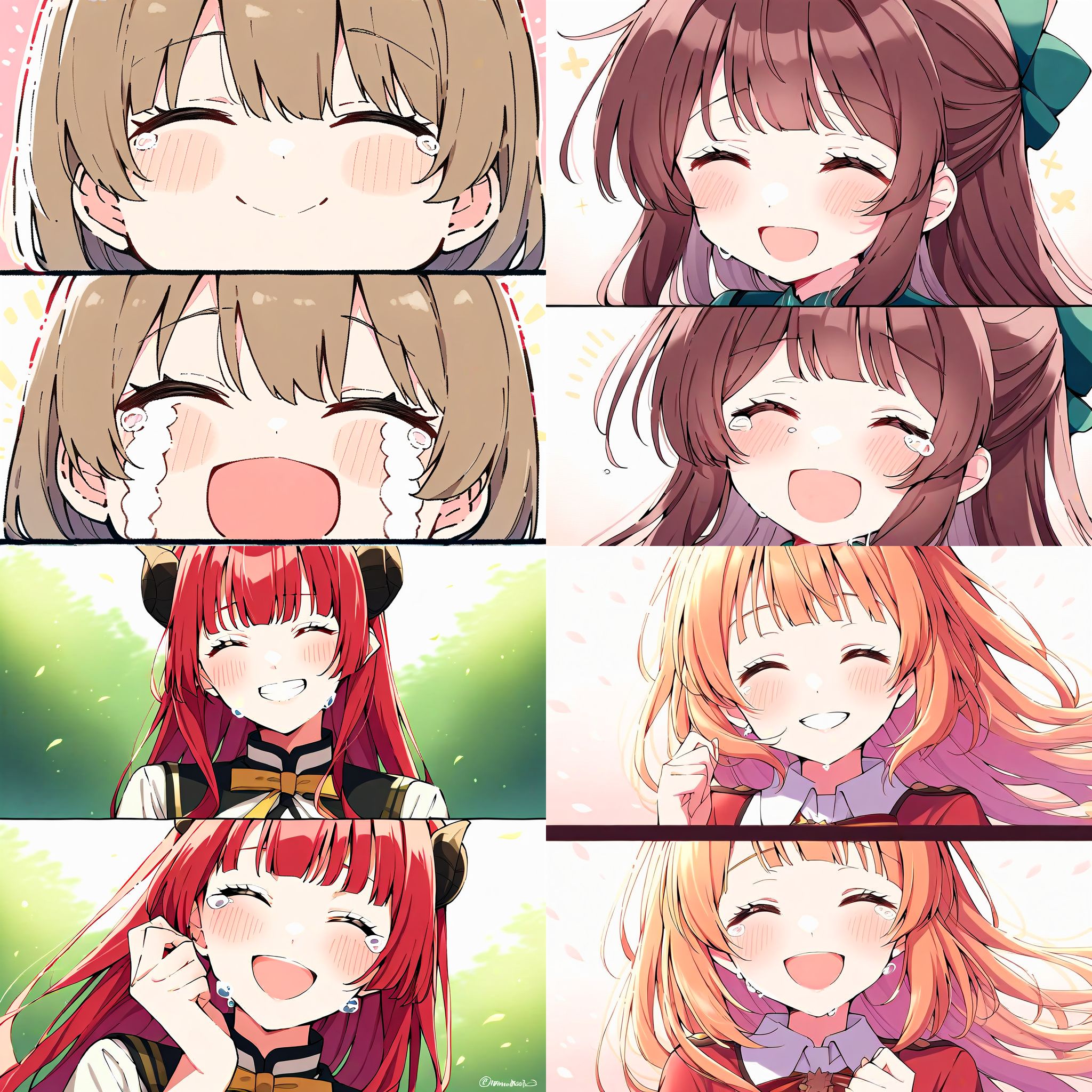}
        \caption{\scriptsize Generation of 2-koma typed illustration, with prompt \textbf{1girl, happy, smile, crying, 2koma}.}
    \end{subfigure}
    \hspace{1em} 
    \caption{\footnotesize Minimal data pruning strategy has allowed various concept genreation, including extremely rare ms-paint like concepts, without harnessing overall generation quality.}
    \label{fig:fig2_combined}
\end{figure}

Unlike the common practice of removing comics or low-quality images, our approach aimed to prune only a minimum of problematic images. This allowed us to expand the overall knowledge base, enhancing the model's understanding of diverse samples, increasing its ability to handle diverse inputs while maintaining overall control. A broader dataset also enables the model to generate low-quality sparse samples, as depicted in Figure \ref{fig:fig2_combined}.

\subsubsection{Resolution}

We trained the v0.1 model within 1MP range, as standard resolution. The v1.0 and v1.1 models were further trained at 2.25MP range, enabling native 2MP generation and up to 20MP generation when combined with proper img2img pipelines with reduced artifacts. The v2.0 model was additionally augmented with 0.15MP images, allowing it to generate outputs at a wide range of resolutions. Generated examples are shown in Figure \ref{fig:highresolution}.

\subsubsection{Limited Corpus}
We identified several critical limitations in the Danbooru tag vocabulary, making it unsuitable for interpolation tasks. For instance, while the model can accurately generate objects like "stained glass" and "sword", it struggles with more complex concepts like "covering wound with left hand" due to insufficient data for such specific actions. Furthermore, the v0.1 model has difficulty processing natural language-based prompts, especially longer ones, as it was not well adapted to such formats.

\subsection{Training Method}
Based on the characteristics of the dataset described earlier, we attempted to overcome such problems by conducting the training using the following methods:

Firstly, we implemented a \textbf{No Dropout Token approach} to ensure that provocative or specific tokens are never excluded. In conventional training methods, random tokens are dropped during image pairing to prevent overfitting and enhance model generalization. However, this approach led to the occasional generation of provocative images. By ensuring provocative tokens were always retained and training the model to recognize these concepts with 100\% accuracy, we found that controlling the sampling the provocative tokens by CFG, or preventing their use entirely effectively prevented the generation of provocative or inappropriate content \cite{waluigi_effect_2023}.

Next, we employed \textbf{Cosine Annealing scheduler} \cite{SGDR} empirically. Such a schedule enables to achieve a lower learning rate and to gain reasonable converged checkpoints with a focus on improving the quality of image and stability of model training. Therefore, we adopted it into v1.0, v1.1, and v2.0 Illustrious models. 

Third, we used \textbf{Quasi-Register Tokens}  \cite{RegisterToken} to embed concepts the model doesn't understand into specific tokens for training. Since the dataset cannot contain all metadata, certain image characteristics may not be reflected. We identified these outlier concepts that the model couldn't comprehend and embedded them into register tokens during training. Conversely, when random tokens are included during training, concepts not represented in the text encoder or metadata can be captured by these random tokens. By attaching random alphanumeric strings, the model is allowed to separate 'bad characteristics' into leftover tokens, by separating known concepts from ambiguous ones. However, we observed that padding tokens used for sequence length matching in batching, are also treated as register tokens, a phenomenon we discuss in detail in the appendix.

Fourth, we trained model in \textbf{Contrastive Learning by Weak-Probability Dropout Tokens}. Similar to the first method, we prevented certain character names or artist names from being dropped with a set probability during training, which improves the model's ability to understand character names and artist styles, while other tokens are dropped as usual. This approach significantly improved character-wise understanding with fewer mixed features. Additionally, we observed that with this method, character learning accelerated even with smaller batch sizes, allowing more contrastive learning between no-character tokens and character token conditions. However, unlike the tag weighting strategy used by NovelAI, the absence of CFG control over character tags sometimes led to the model generating specific characters inductively, leading to weak dataset leakage, as expected.

Fifth, we implemented a simple \textbf{paraphrasing sequence process} to train the model on more diverse texts. Tags like "1girl, 1boy" were paraphrased as "one girl, single women," etc. This process enables the model to understand various inputs, instead of relying strictly on tag-based conditioning.

Finally, we adopted \textbf{Multi Level Dropout} by dividing the dropout into 4 stages, ranging from minimal, critical tokens to full tags. This allows the model to adapt to  varying levels of caption detail. By 30\% chance, we utilize max(30\% * total tokens, 10) tags, 20\% chance, max (40\% * total tokens, 15), 10\% chance, min(6, total tokens), 4\% chance, min(total tokens, 4) tokens. The no-dropout tokens ignores this rule, for strict controllability. 

We applied the eps-prediction loss objective and also utilized Input Perturbation Noise Augmentation with strength \begin{itemize}\item \( 0 < \varepsilon < 0.1 \)\end{itemize}  \cite{ning2023inputperturbationreducesexposure}, and Debiased Estimation Loss  \cite{yu2024unmaskingbiasdiffusionmodel}. We observed noise offset  \cite{lin2024common} to be useful for broader color ranges. However, with lower batch sizes, it was not suitable for the common training procedure. 

\section{Training Setups}
\begin{table*}[htbp]
\caption{Illustrious Training Setups}
\begin{center}
\resizebox{\textwidth}{!}{ 
\begin{tabular}{c|cccccccccc}
\toprule
\multirow{2}{*}{\textbf{Model}}  & \multirow{2}{*}{\textbf{Dataset Size}} & \multirow{2}{*}{\textbf{Batch Size}} & \multirow{2}{*}{\textbf{LR}} & \multirow{2}{*}{\textbf{TE LR}} & \multirow{2}{*}{\textbf{Epoch}} & \multirow{2}{*}{\textbf{Resolution}} & \multirow{2}{*}{\textbf{Prompt Style}} & \multicolumn{2}{c}{\textbf{Tag Manipulation}} & \multirow{2}{*}{\textbf{Multi Caption}}  \\ 
 & & & & & & & & \textbf{Dropout Level} & \textbf{Register Token} \\ 
\midrule
\multicolumn{1}{c|}{Illustrious v0.1} & 7.5M & 192 & 3.5e-5 & 4.5e-6 & 20 & 1024 $\times$ 1024 & Tag & N & N & N \\ 
\multicolumn{1}{c|}{Illustrious v1.0} &  10M & 128 & 1e-5 & 6e-6 & 8 & 1536 $\times$ 1536 & Tag & Y & Y & N \\ 
\multicolumn{1}{c|}{Illustrious v1.1} &  12M  & 512  & 3e-5 & 4e-6 & 4 & 1536 $\times$ 1536 & Tag + Natural Language  & Y & Y & N \\ 
\multicolumn{1}{c|}{Illustrious v2.0} & 20M & 512 & 4e-5 & 3e-6 & 2 & \textbf{1536 $\times$ 1536} & Tag + Natural Language & Y & Y & Y \\ 
\bottomrule
\end{tabular}
} 
\label{tab2}
\end{center}
\end{table*}

We trained models using different strategies sequentially.
\begin{itemize}
    \item Illustrious v0.1 was trained on a 7.5M dataset consisting of 1024 × 1024 images with a batch size of 192. The data were tagged using the original Danbooru tags. The learning rate for the U-Net was set to 3.5e-5, and the text encoder learning rate was 4.5e-6, trained over 20 epochs.
        
    \item Illustrious v1.0 used a 10M dataset of 1536 × 1536 images with a batch size of 128, also tagged with the original Danbooru tags, with duplicate separated higher-resolution images. The U-Net learning rate was 1e-5, and the text encoder learning rate was 6e-6, trained over 8 epochs. For this dataset, we applied tag manipulation strategies, Dropout-Leveling and Register Tokens.
    
    \item Illustrious v1.1 was trained on a 12M dataset of the same 1536 × 1536 resolution images as v1.0. It used a batch size of 512 and was trained for 4 epochs with a U-Net learning rate of 3e-5 and a text encoder learning rate of 4e-6. The dataset for v1.1 was tagged using a combination of natural language descriptions and tags.
    
    \item Illustrious v2.0 was trained on a 20M dataset with the same 1536 × 1536 image resolution as v1.1. The model was trained with a batch size of 512 for 2 epochs, using a U-Net learning rate of 4e-5 and a text encoder learning rate of 3e-6. Illustrious v2.0 mainly incorporated the multi-caption method for enhanced text-image correspondence.
\end{itemize}

\section{Evaluation}
We conducted evaluations of our models with the well known rating method, Elo Rating and TrueSkill 2, and Character wise similarity, CCIP.
\subsection{User Preference with Elo Rating}
The ELO Rating system, developed by Arpad Elo, is being widely used to evaluate user's skill levels in competitive survey by adjusting user's ratings based on match outcomes. The rating changes reflect the difference between expected and actual results, providing a dynamic measure of a user's relative strength. The standard Elo rating update formula is given as following:
\[
R' = R + K \times (\text{actual} - \text{expected})
\]

Where:

\begin{itemize}
    \item \( R' \) is the new rating after the match.
    \item \( R \) is the current rating before the match.
    \item \( K \) is the K-factor, a constant that determines the sensitivity of rating changes.
    \item \( \text{actual} \) is the actual result of the match (1 for a win, 0.5 for a draw, 0 for a loss).
    \item \( \text{expected} \) is the expected score, calculated using the formula:
\end{itemize}

\[
\text{expected} = \frac{1}{1 + 10^{(R_{\text{opponent}} - R)/400}}
\]

Here, \( R_{\text{opponent}} \) is the rating of the opponent. 

Recently, various research studies have been evaluating models based on ELO rating using win rates. \cite{chatbotarena} \cite{LLMasaJudge} In the case of images, in particular, traditional metrics \cite{SSIM} \cite{LPIPS} tend to focus on the similarity of the image itself, such as pixel-level similarity, rather than the meaning of the image. Therefore, human evaluation becomes even more essential in such cases.

\begin{figure*}[ht!]
    \centering
    \begin{minipage}{0.5\textwidth}
        \includegraphics[width=\textwidth,height=8cm,keepaspectratio]{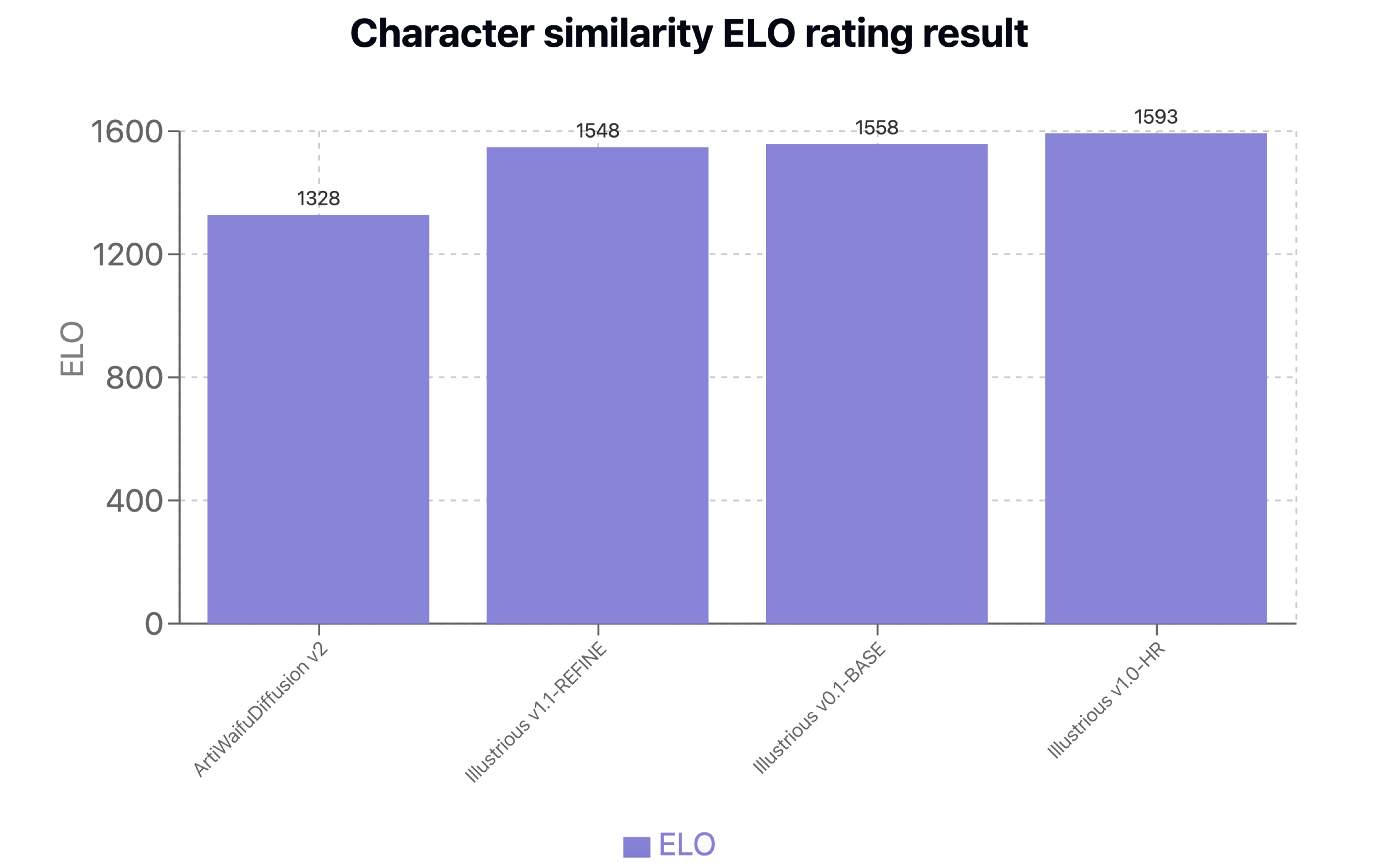}
    \end{minipage}\hfill
    \begin{minipage}{0.5\textwidth}
    \includegraphics[width=\textwidth,height=8cm,keepaspectratio]{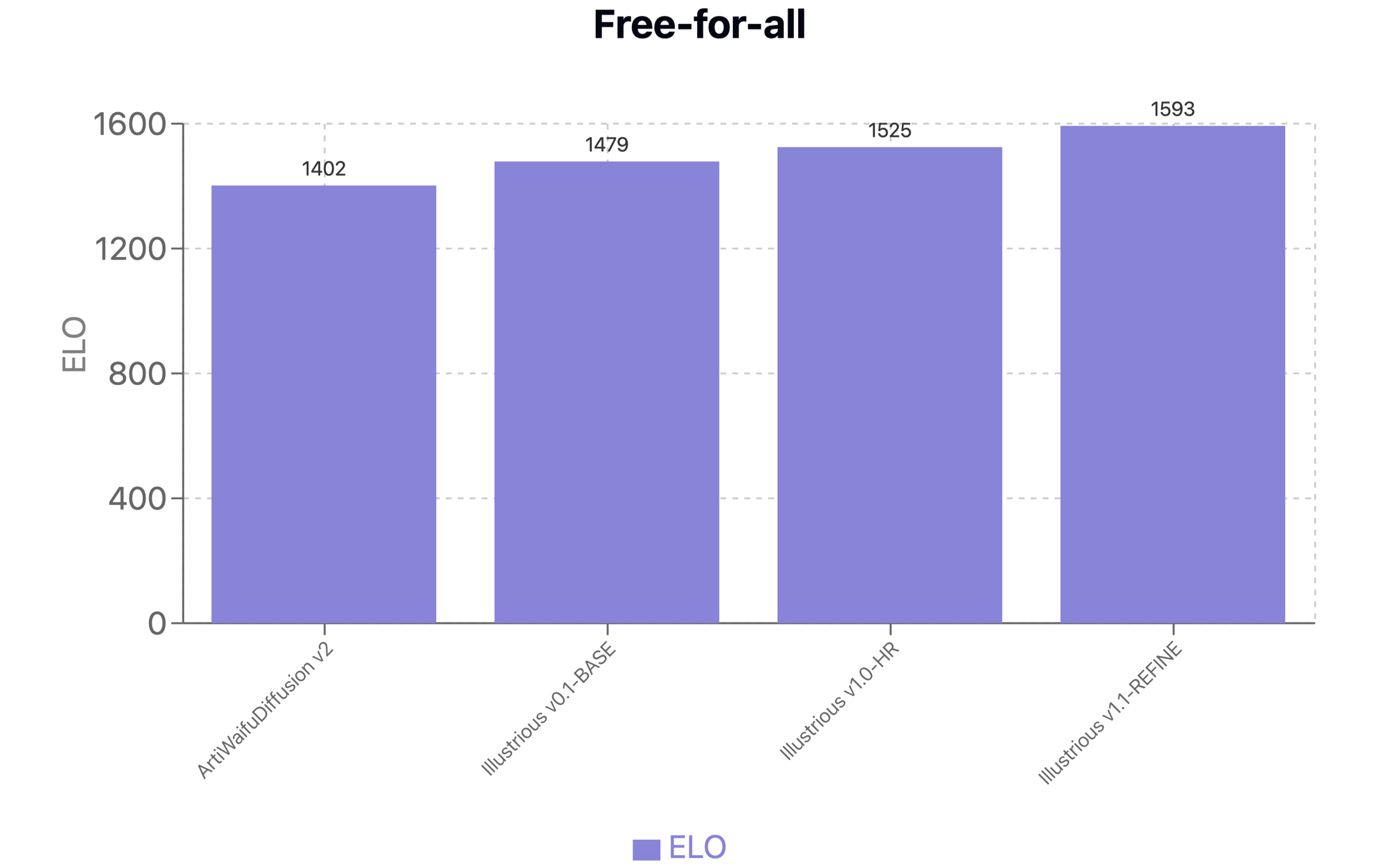}
    \end{minipage}\hfill
\caption{\textbf{Character Similarity ELO Ratings Result}, time-weighted average is applied. and \textbf{Free-for-all ELO}}
\label{fig: ELO_1}
\end{figure*}

Fixed-Characteristics means 2 random images are shown on poll and users select one with fixed prompt generations. This match accepts draw. Free-prompt-duel means 2 random images from free prompt and one is selected. Free-for-All is 1 vs 1 vs 1 vs 1 match.

\begin{figure*}
\centering
{\includegraphics[width=\textwidth]{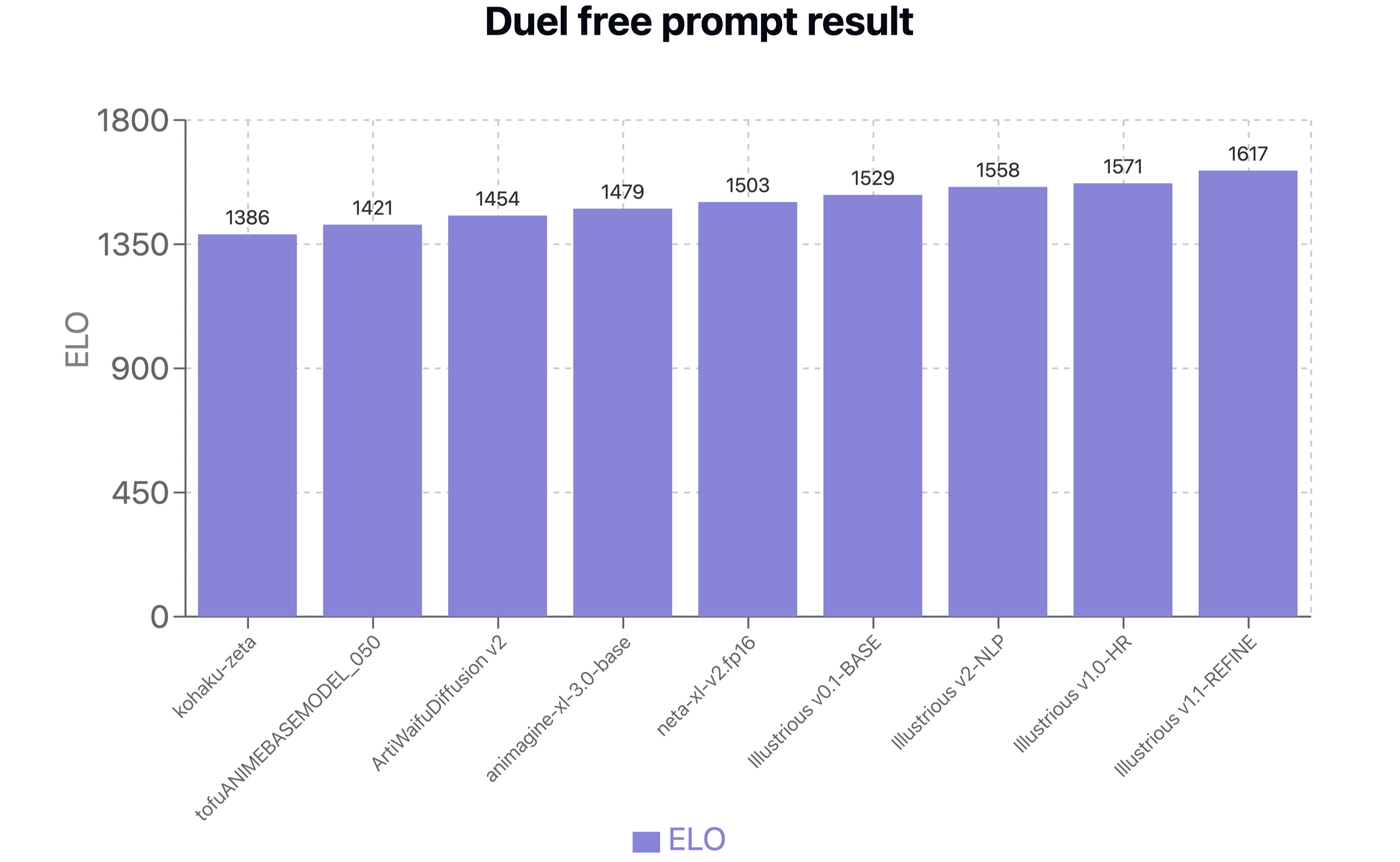}}
\caption{\textbf{Duel Free Prompt ELO Result}.}
\label{fig:ELO_2}
\end{figure*}

\subsection{CCIP}
CCIP  \cite{CCIP} is a metric designed to estimate visual differences between given grouped set and given image for character basis, focusing on feature extraction metric based on CLIP. The difference value in CCIP is calculated as average of given formula:
\[
D(I_1, I_2) = M(I_1,I_2)
\]
Where:

\begin{itemize}
    \item \( D(I_1, I_2) \) represents the difference value between images \( I_1 \) and \( I_2 \).
    \item \( I_1 \) and \( I_2 \) are the two images being compared.
    \item \( M\) is a CCIP model.
\end{itemize}

\begin{figure*}
\centering
{\includegraphics[width=\textwidth]{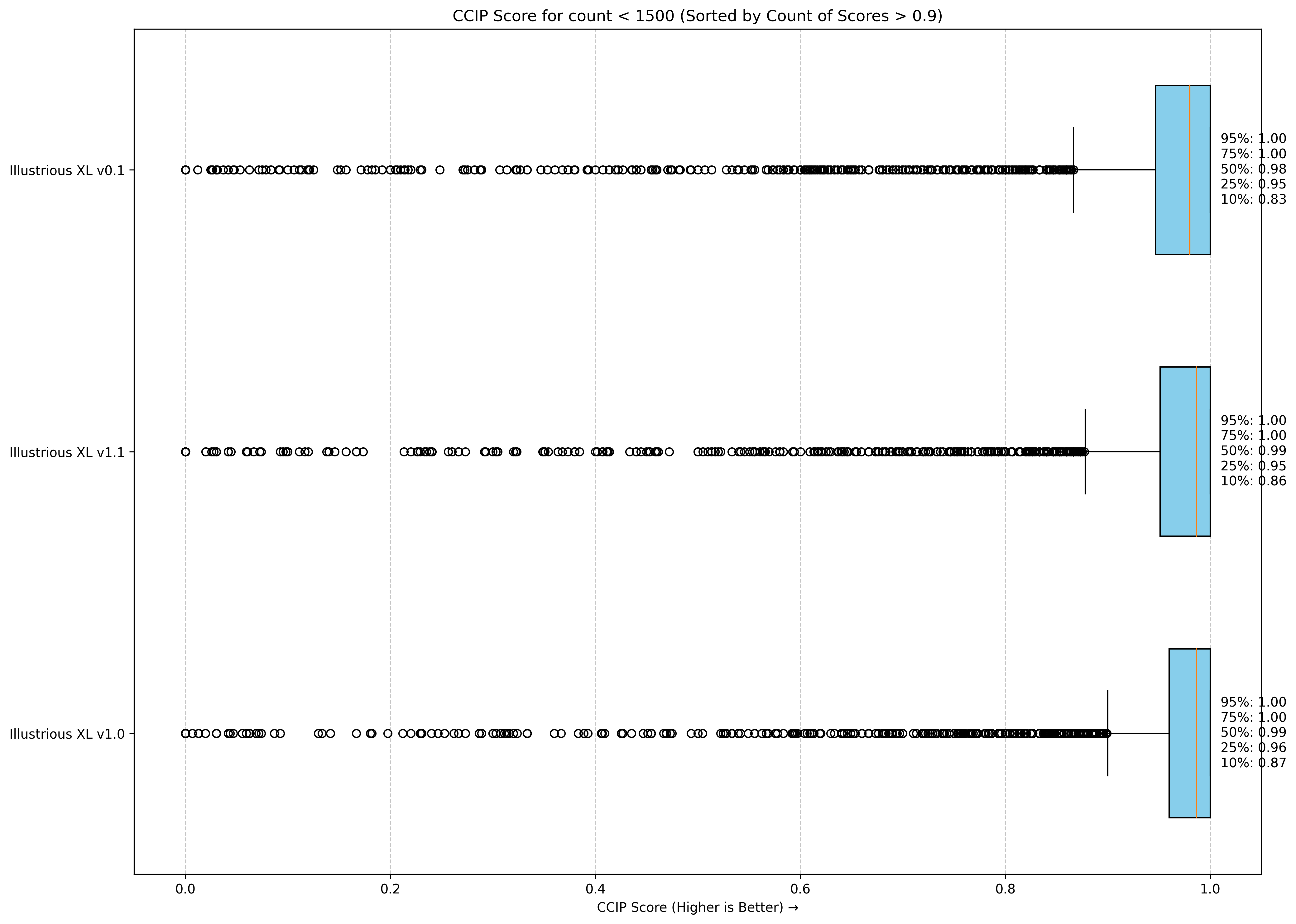}}
\caption{\textbf{CCIP Score}.}
\label{fig:CCIP}
\end{figure*}

CCIP extracts visual features of characters from images and quantifies the differences to assess character similarity. CCIP effectively identifies whether two images contain the same character, focusing on features like facial attributes, clothing, and color schemes.

\subsection{TrueSkill Algorithm}
TrueSkill is a skill-based ranking system proposed by Microsoft. Unlike the Elo rating system, which was originally developed for chess, TrueSkill requires less trials to estimate users' expected numerical skill scoring, which is more stable for sparse model duels conditions. \cite{TrueSkill} \cite{TrueSkill2} As documented, the update equations are given as following:
\[
\mu' = \mu + \frac{\sigma^2}{\sigma^2 + \beta^2} \times (s - \mu)
\]

Where:

\begin{itemize}
    \item \( \mu' \) is the updated mean skill level of the player.
    \item \( \mu \) is the current mean skill level before the update.
    \item \( \sigma^2 \) is the variance representing the uncertainty in the player's skill estimate.
    \item \( \beta^2 \) is the variance of the game outcome, reflecting the randomness inherent in game results.
    \item \( s \) is the performance score derived from the game outcome.
\end{itemize}

The variance \( \sigma^2 \) is also updated every match, to reflect the change in uncertainty after each game. 

\bigskip

By integrating these algorithms and metrics into our evaluation framework, we aim to provide a comprehensive assessment that balances quantitative measures with human judgment, which is particularly important in domains like image evaluation where subjective interpretation plays a significant role.

\begin{figure*}[ht!]
    \centering
    \begin{minipage}{0.5\textwidth}
        \includegraphics[width=\textwidth,height=8cm,keepaspectratio]{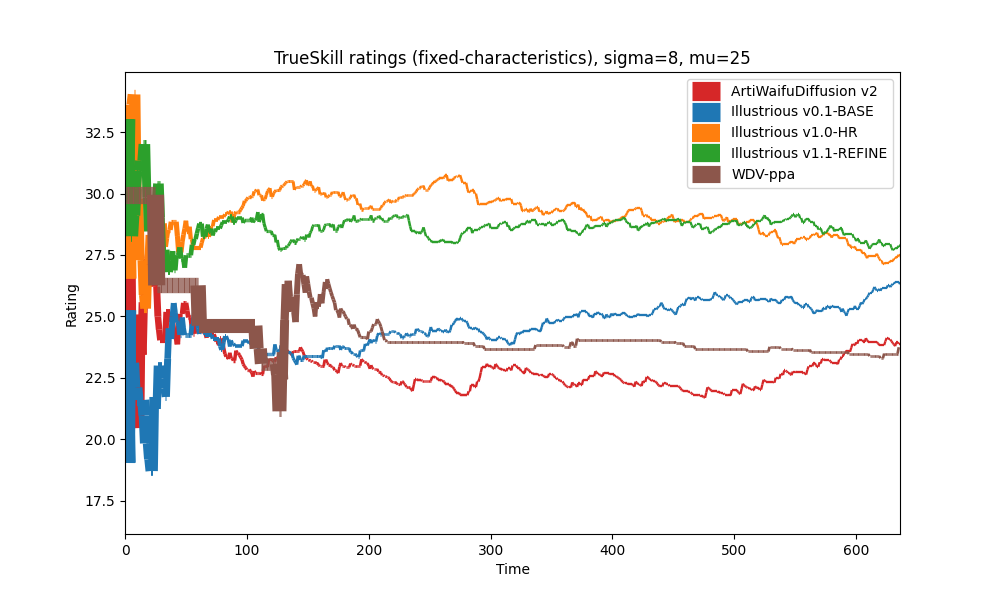}
    \end{minipage}\hfill
    \begin{minipage}{0.5\textwidth}
    \includegraphics[width=\textwidth,height=8cm,keepaspectratio]{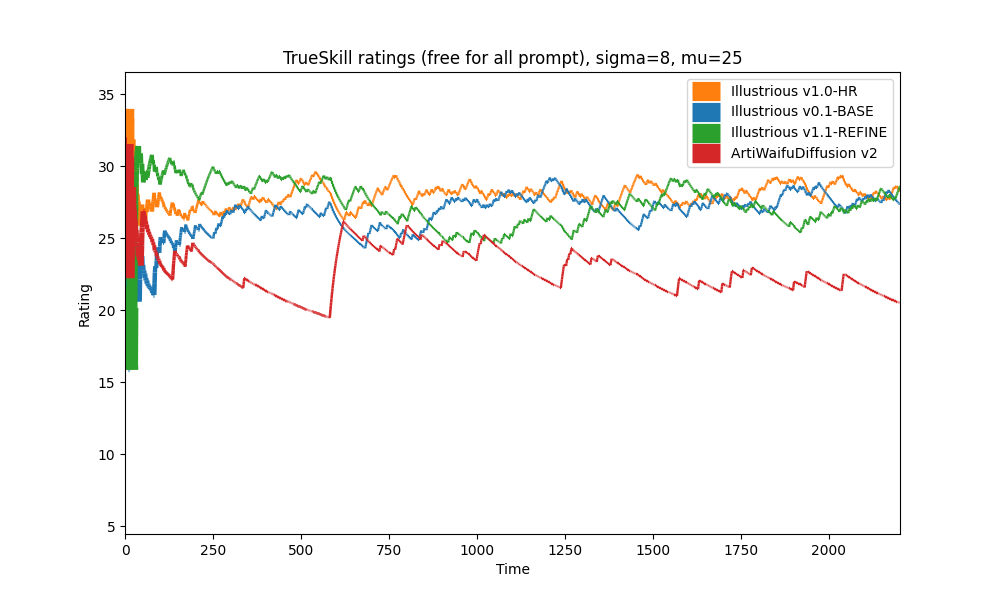}
    \end{minipage}\hfill
\caption{\textbf{TrueSkill Ratings (fixed-characteristics)} and \textbf{TrueSkill Ratings (Free-for-All Prompt)}}
\label{fig: Trueskill2}
\end{figure*}
\begin{figure*}
\centering
{\includegraphics[width=\textwidth]{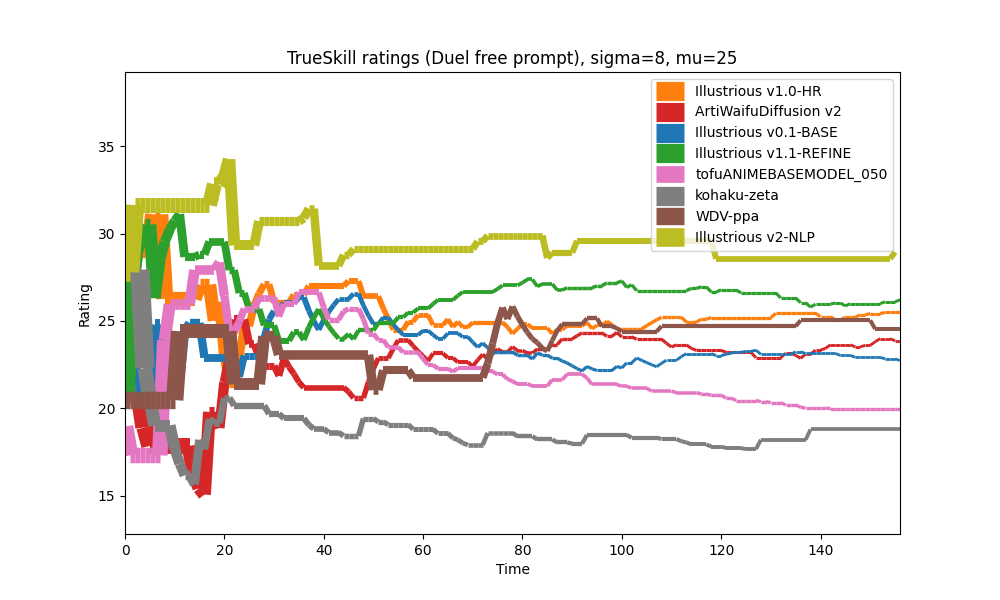}}
\caption{\textbf{TrueSkill ratings (Duel Free Prompt)}.}\label{fig:TrueSkill3}
\end{figure*}

\newpage
\section{Limitations and Future Works}
\textbf{Limitations}

Illustrious is a generalized anime image generation model that can create a variety of images through detailed prompts. However, it has the following limitations.

First, the CLIP text encoder's instability in handling character details can lead to less effective performance in embedding similarity calculations. Recently, models such as Flux or Kolors have addressed this issue by using alternatives like T5 and GLM instead of the CLIP text encoder.

Second, the Danbooru dataset predominantly relies on tag-based metadata, which makes it difficult to describe images across multiple dimensions. This limitation creates challenges in controlling the specific composition and positioning of multiple characters or actions. To fully address this issue, detailed descriptions of each character, their positions, backgrounds, and relationships are necessary—elements often missing in tag-based and other common large-scale datasets.

With the enhanced natural language capabilities introduced in v2.0 and a custom-built, sophisticated dataset (to be released in future work), we propose the development of large-scale, refined natural language datasets to overcome these limitations.

\textbf{Future Works}

Below are some possible directions of Illustrious in future work.

One key challenge identified is the task of rendering text within images for anime image generation. While many real-image generation models can partially support the embedding of text in images, open-source anime image generative models struggle with this task. Phrases like "Merry Christmas" or "Happy New Year" can sometimes be rendered correctly due to their frequent appearance in datasets, but generating full sentences or meaningful words within anime images remain a significant challenge. 

The Illustrious v2.0 shows notable improvements in generating glyphs, albeit with limited capability, through synthetic captions. Future models could be significantly enhanced by incorporating OCR-based datasets and conditioning as part of the training process.

\newpage
\appendix

\section{Appendix / Supplemental Material}
\subsection{Resolution}
Illustrious v1.0+ is capable of generating images in 1536x1536 natively, which can be expanded to 2048x2048 at farmost without any modifcation. In higher resolutions, it allows over 20MP+ generation as depicted in Figure \ref{fig:highresolution}, while other models fails to follow.

\subsection{Analysis}
\label{analysis}
There are experimental results obtained through various efforts to make Illustrious model. We will describe this in detail as follows.

\subsubsection{Limitations regard to aesthetic/biased models}
As part of stabilization, careful considerations must be given during the aesthetic tuning stage. Fitting a baseline model into human preferences \cite{DPO} \cite{D3PO} \cite{HFT2I} can degrade its performance on the true data distribution. This also reduces the diversity of image generation, limiting the model’s applicability. Such overfitting makes future fine-tuning significantly more difficult compared to using an unbiased model, as it necessitates re-aligning the model's knowledge with the true data distribution.

\begin{figure}[htb]
    \centering
    \begin{subfigure}[b]{0.45\textwidth}
        \centering
        \includegraphics[width=\textwidth]{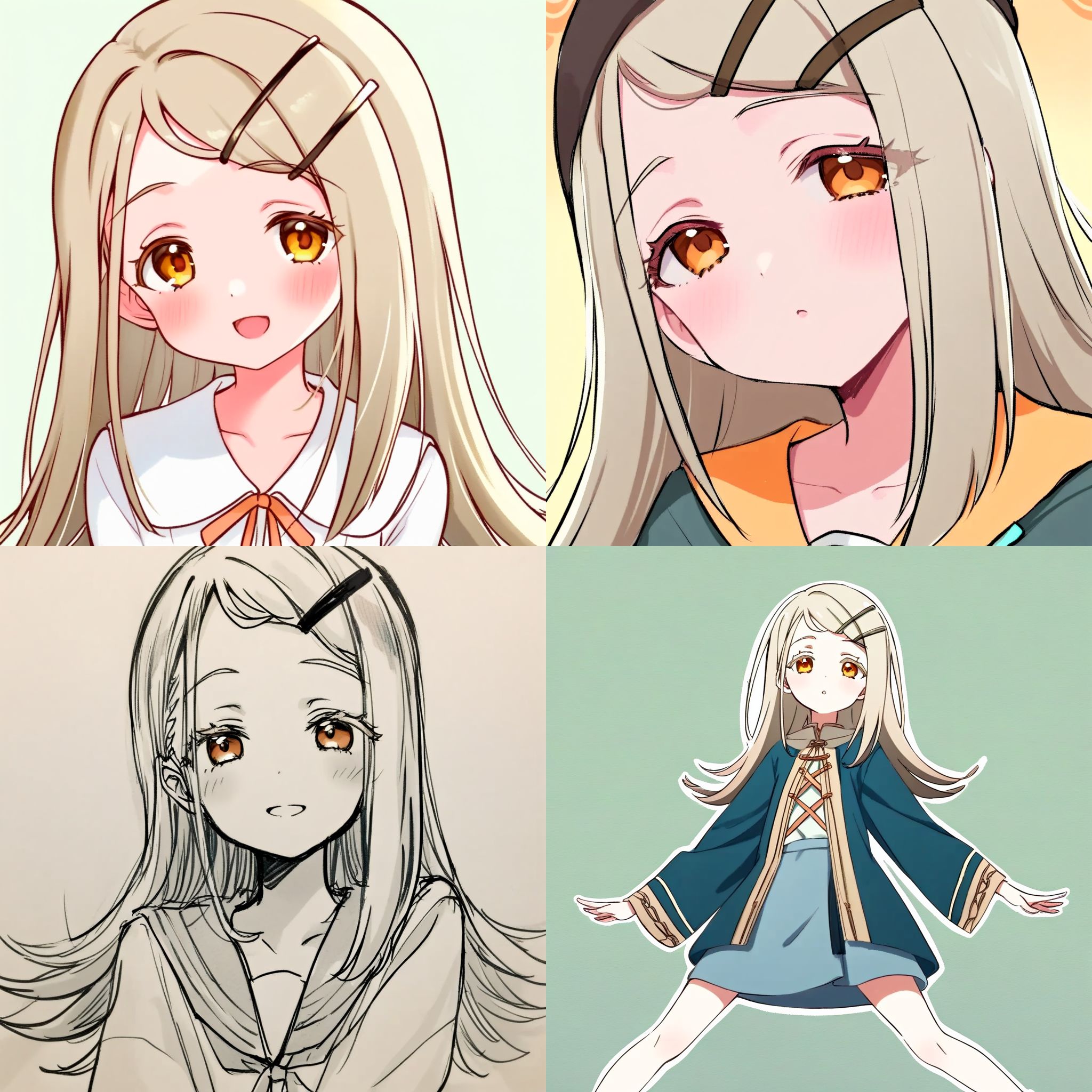}
        \caption{\scriptsize LoRA-applied result in Illustrious v0.1. The prompt was \textbf{1girl,  shinosawa hiro, general, masterpiece}.}
    \end{subfigure}
    \hspace{1em} 
    \begin{subfigure}[b]{0.45\textwidth}
        \centering
        \includegraphics[width=\textwidth]{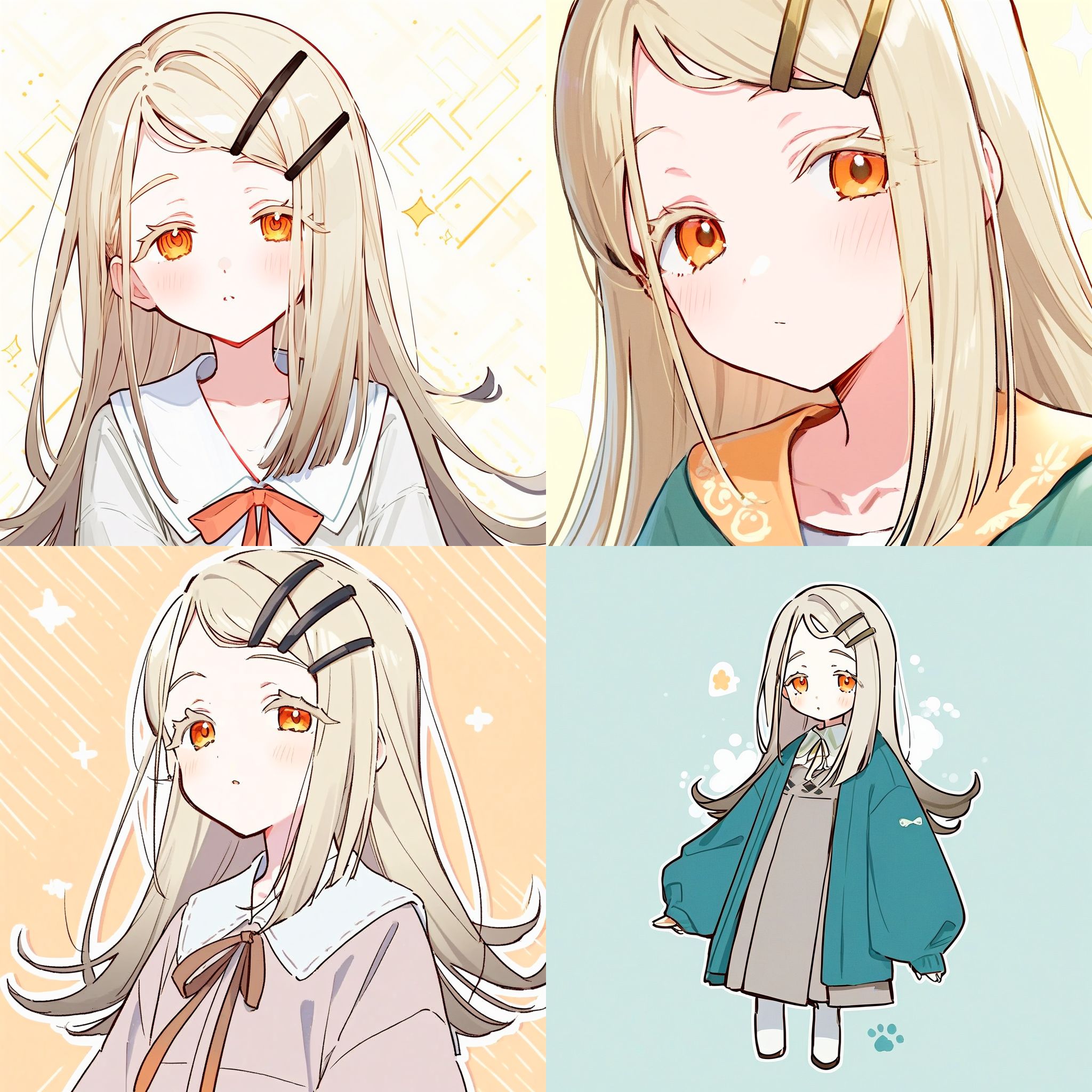}
        \caption{\scriptsize LoRA-applied result in Illustrious v2.0.}
    \end{subfigure}
    \hspace{1em} 
    \caption{\footnotesize The LoRA trained on Illustrious v0.1, is widely usable across checkpoints.}
    \label{fig:loracapability}
\end{figure}

For this reason, to ensure broader public usability, we have decided to release non-fine-tuned base models. These models can be adapted for various tasks and concepts. We also demonstrate that model-derived add-ons, such as LoRAs \cite{LoRA}, remain compatible with future models and allow for effective model derivation as depicted in Figure \ref{fig:loracapability}\footnote{https://civitai.com/models/794775/llustrious-xl-shinosawa}. 

\subsubsection{Multiple-character generation}
\begin{figure}[htb]
    \centering
    \begin{subfigure}[b]{0.20\textwidth}
        \centering
        \includegraphics[width=\textwidth]{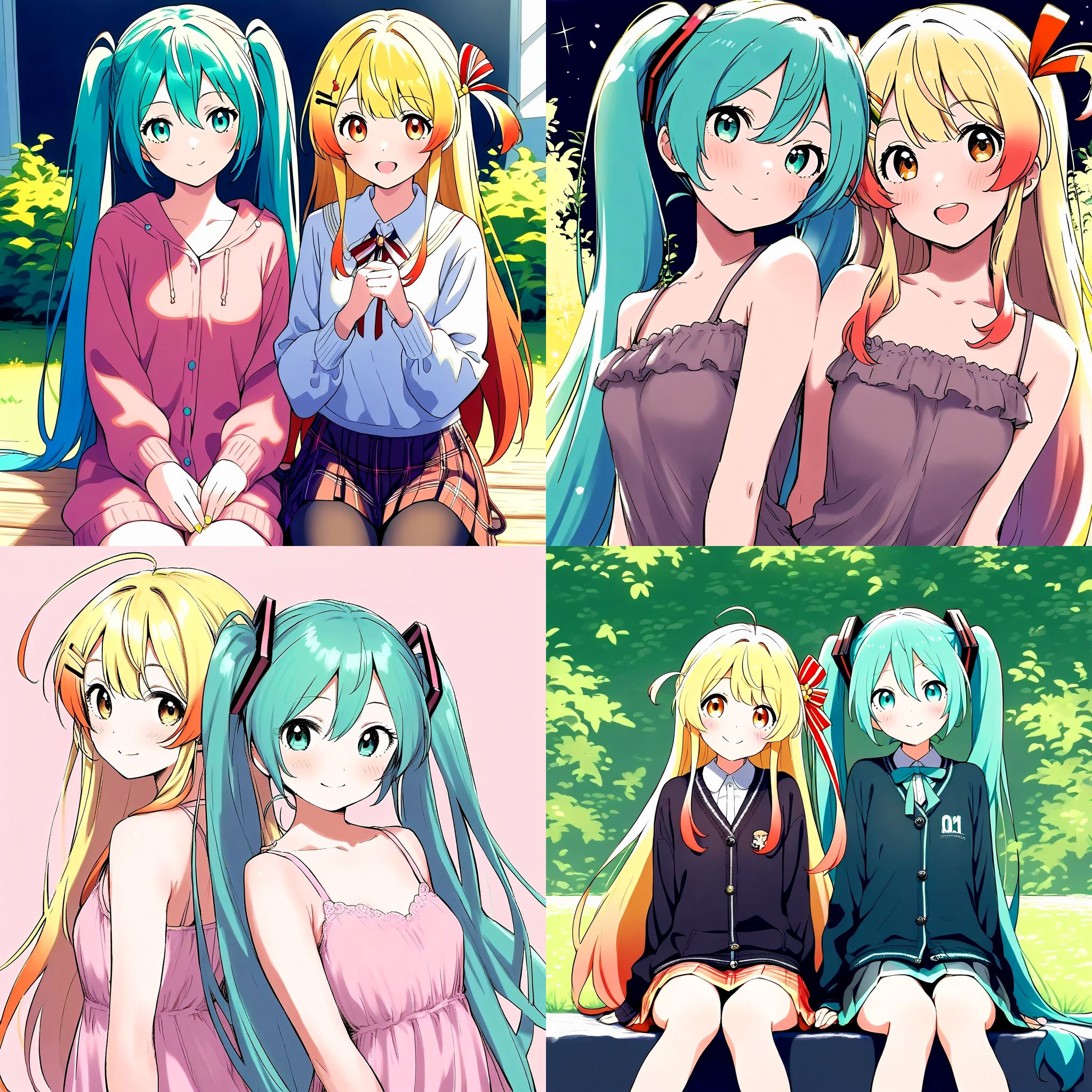}
        \caption{\scriptsize Multi-character separation result in Illustrious v0.1. The prompt was \textbf{2girls, otonose kanade, hatsune miku, side-by-side, masterpiece}.}
    \end{subfigure}
    \hspace{1em} 
    \begin{subfigure}[b]{0.20\textwidth}
        \centering
        \includegraphics[width=\textwidth]{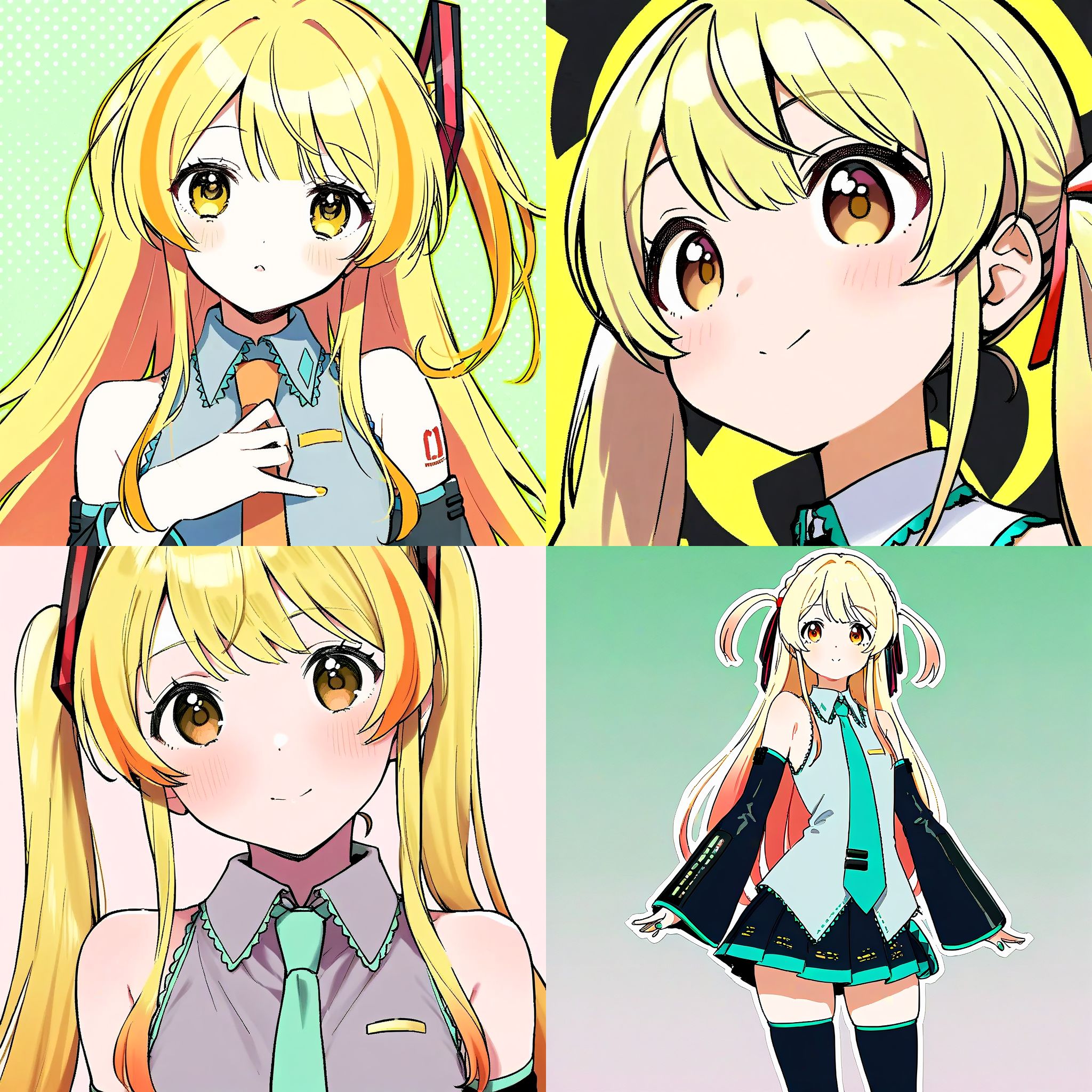}
        \caption{\scriptsize Character combine result in Illustrious v0.1. The prompt was \textbf{1girl, otonose kanade, hatsune miku (cosplay),general, masterpiece, blonde hair}.}
    \end{subfigure}
    \hspace{1em} 
    \begin{subfigure}[b]{0.38\textwidth}
        \centering
        \includegraphics[width=\textwidth]{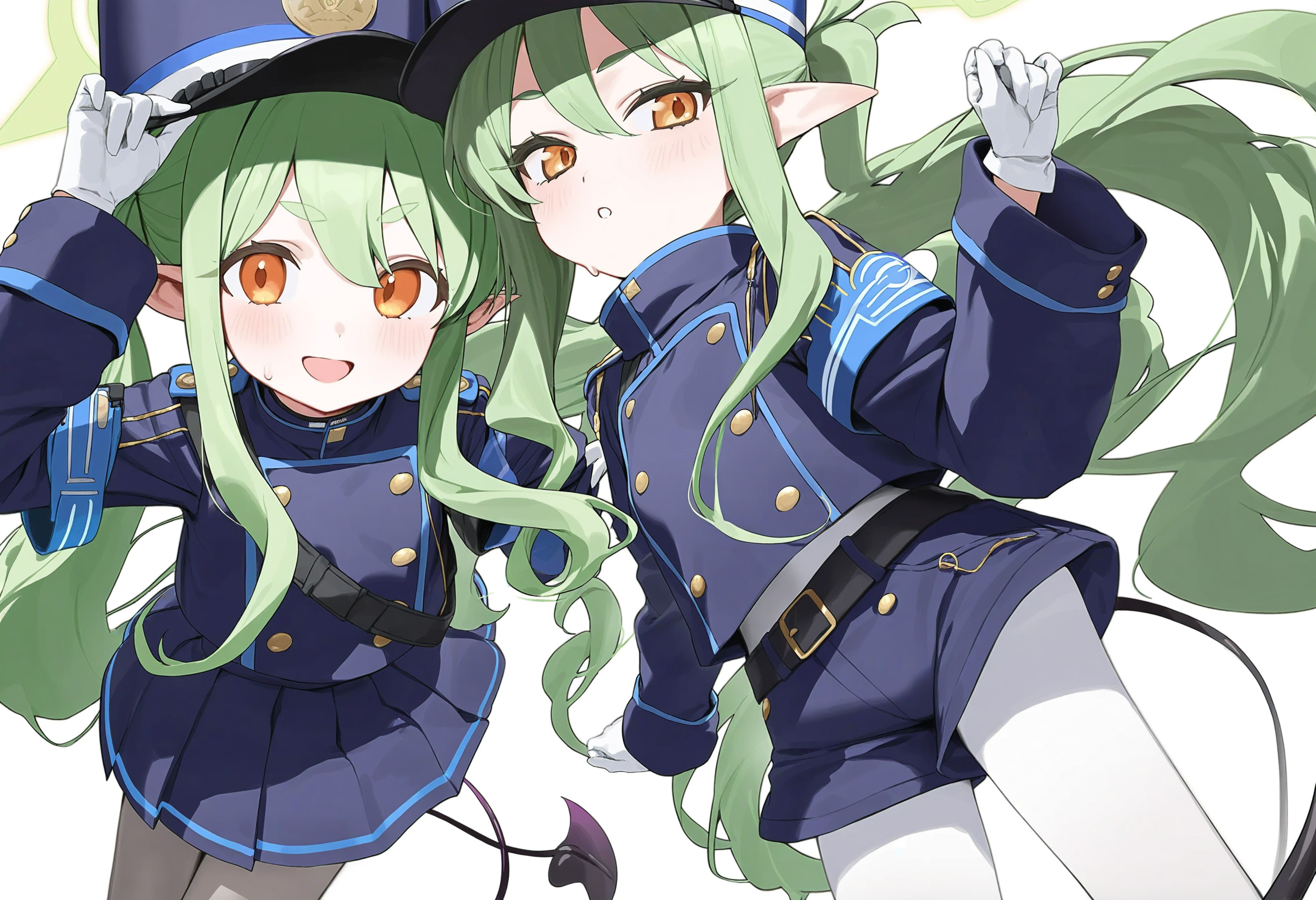}
        \caption{\scriptsize Multi-character separation result in Illustrious v2.0. The prompt was \textbf{multiple girls, 2girls, nozomi (blue archive), hikari (blue archive),year 2023,dynamic angle,shiny, hat, pointy ears,long hair, shorts, green hair, siblings, pantyhose, thick eyebrows, demon tail, gloves, open mouth, tail, twintails, smile, blush, looking at viewer, sisters, orange eyes, white gloves, skirt parted lips, simple background,white background, masterpiece, absurdres}}
    \end{subfigure}
    \caption{\footnotesize The character separation behavior of Illustrious.}
    \label{fig:charseparation}
\end{figure}

We observe that the strict token control approach results in excelling character feature separation in limited budget. The phenomenon is sustained from Illustrious v0.1, toward the cutting edge model, Illustrious v2.0, as depicted in Figure \ref{fig:charseparation}. 

\newpage
\subsubsection{the Effect of "Long Prompts"}
It is commonly known that long prompts or detailed tags are capable of generating sophisticated images. Illustrious also benefits from upsampled / detailed captions, especially when controlled by sophisticated models. While simple prompts directly exposes model's creativity, we recommend sophisticated captions to further utilize the prompts and models' capability.

\begin{figure}[htbp]
    \centering
    \begin{subfigure}[b]{0.40\textwidth}
        \centering
        \includegraphics[width=\textwidth]{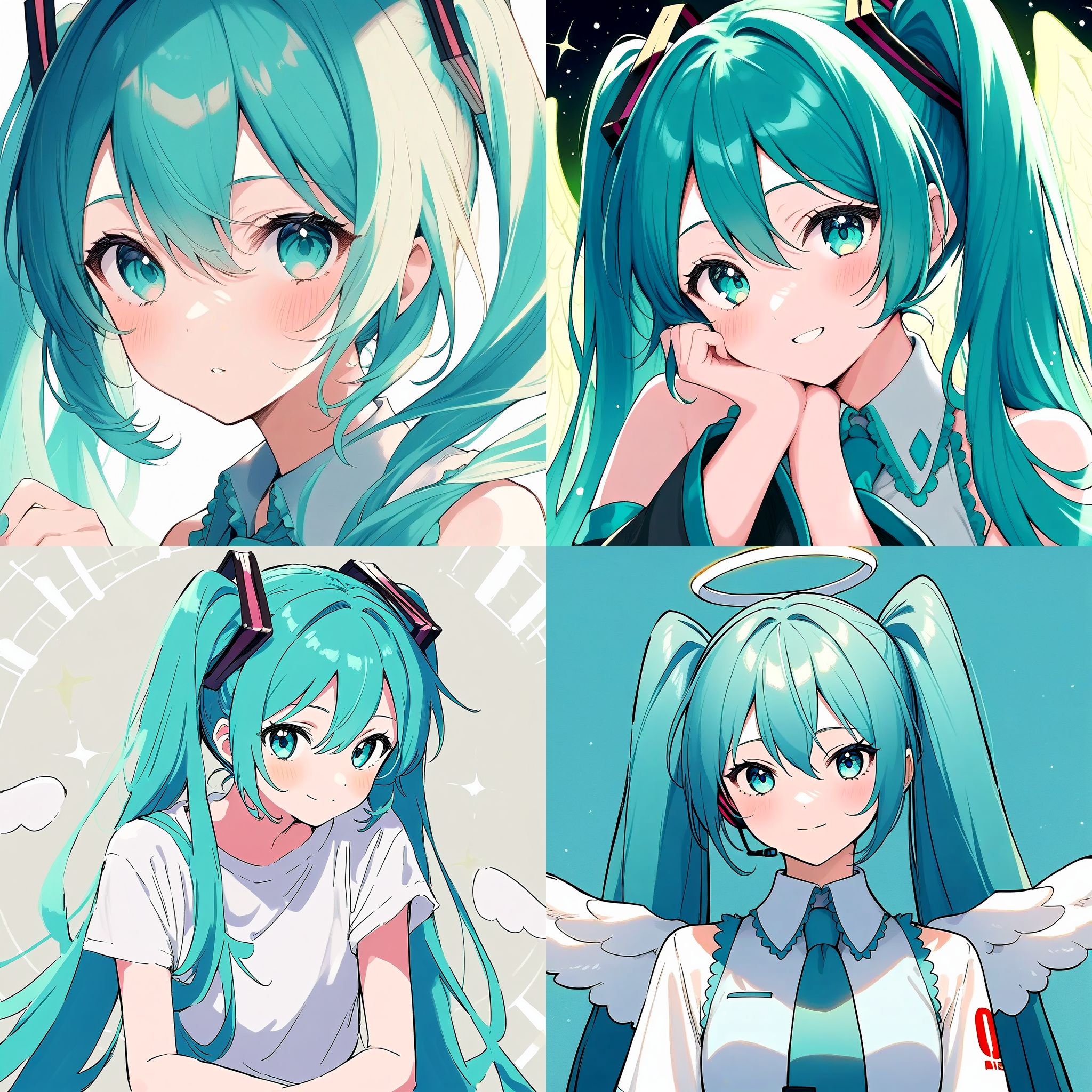}
        \caption{\scriptsize Simple prompt generation with Illustrious v0.1, prompt \textbf{1girl, hatsune miku, angel, masterpiece, general}.}
    \end{subfigure}
    \hspace{1em} 
    \begin{subfigure}[b]{0.40\textwidth}
        \centering
        \includegraphics[width=\textwidth]{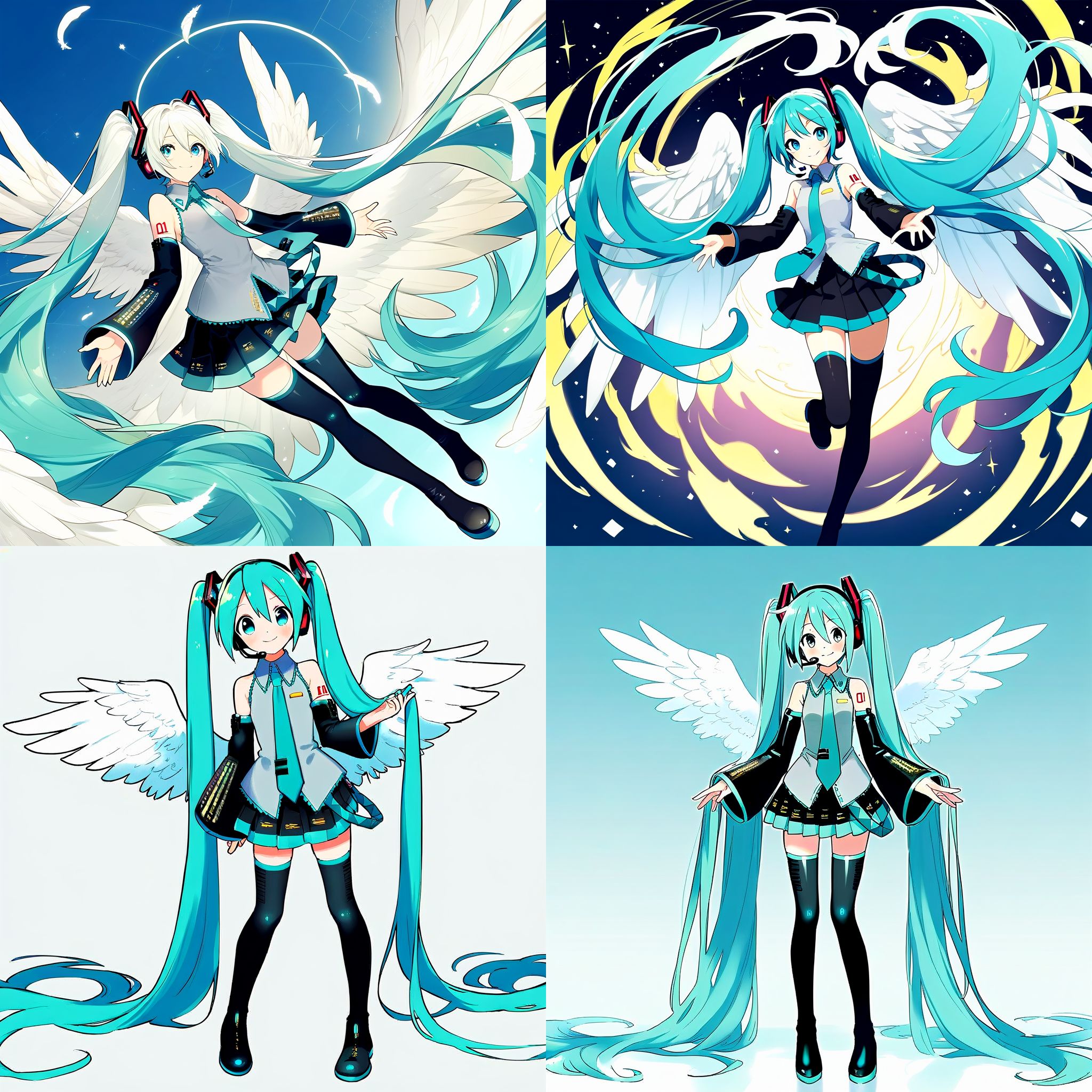}
        \caption{\scriptsize Complex prompt generation upsampled by TIPO with Illustrious v0.1, prompt \textbf{1girl, hatsune miku. An illustration of a girl with long white hair and wings. she is wearing a school uniform with a red bow on her head and a pair of headphones on her ears. the wings are spread out behind her, creating a sense of movement and energy. the overall style of the illustration is anime-inspired. solo, skirt, feathered wings, necktie, smile, very long hair, collared shirt, long hair, headset, blue eyes, aqua necktie, looking at viewer, black footwear, black skirt, twintails, grey shirt, bare shoulders, detached sleeves, full body, zettai ryouiki, closed mouth, miniskirt, sleeveless, boots, thighhighs, shirt, standing, wing collar, aqua hair, sleeveless shirt, pleated skirt, angel wings, absurdly long hair, wings, black thighhighs,masterpiece, general}.}
    \end{subfigure}
    \caption{\footnotesize The upsampling prompt can escape trivial solutions by providing details.}
    \label{fig:upsample}
\end{figure}

For this, we utilize TIPO library \cite{yeh2024tipo}, to show the drastic sample differences across the models, in Figure \ref{fig:upsample}.

\subsubsection{Batch Size and Learning Rates}
We found that large batch sizes can effectively help sparse tags to learn, making the model more stable against parameter updates. \cite{batchsize1} \cite{batchsize2} In contrast, small batch sizes lead to more frequent attention binding, which benefits general / broader concept handling. This suggests that when training on large datasets which are focused on few new concepts, using small batch sizes can accelerate the learning process. However, for sparse concepts, larger batch sizes promote more stable training.

Additionally, if the learning rate falls below a specific threshold, the model may struggle to learn new concepts, favoring convergence toward stable attention splits rather than forming new attention bindings. Based on these observations, we propose that using adaptive batch sizes \cite{AdaBatch}, combined with learning rate scheduling, could offer a more effective alternative for model training.

\subsection{Inpainting}
\begin{figure*}[htb]
    \begin{minipage}{0.5\textwidth}
    \centering
    \includegraphics[width=\textwidth,height=8cm,keepaspectratio]{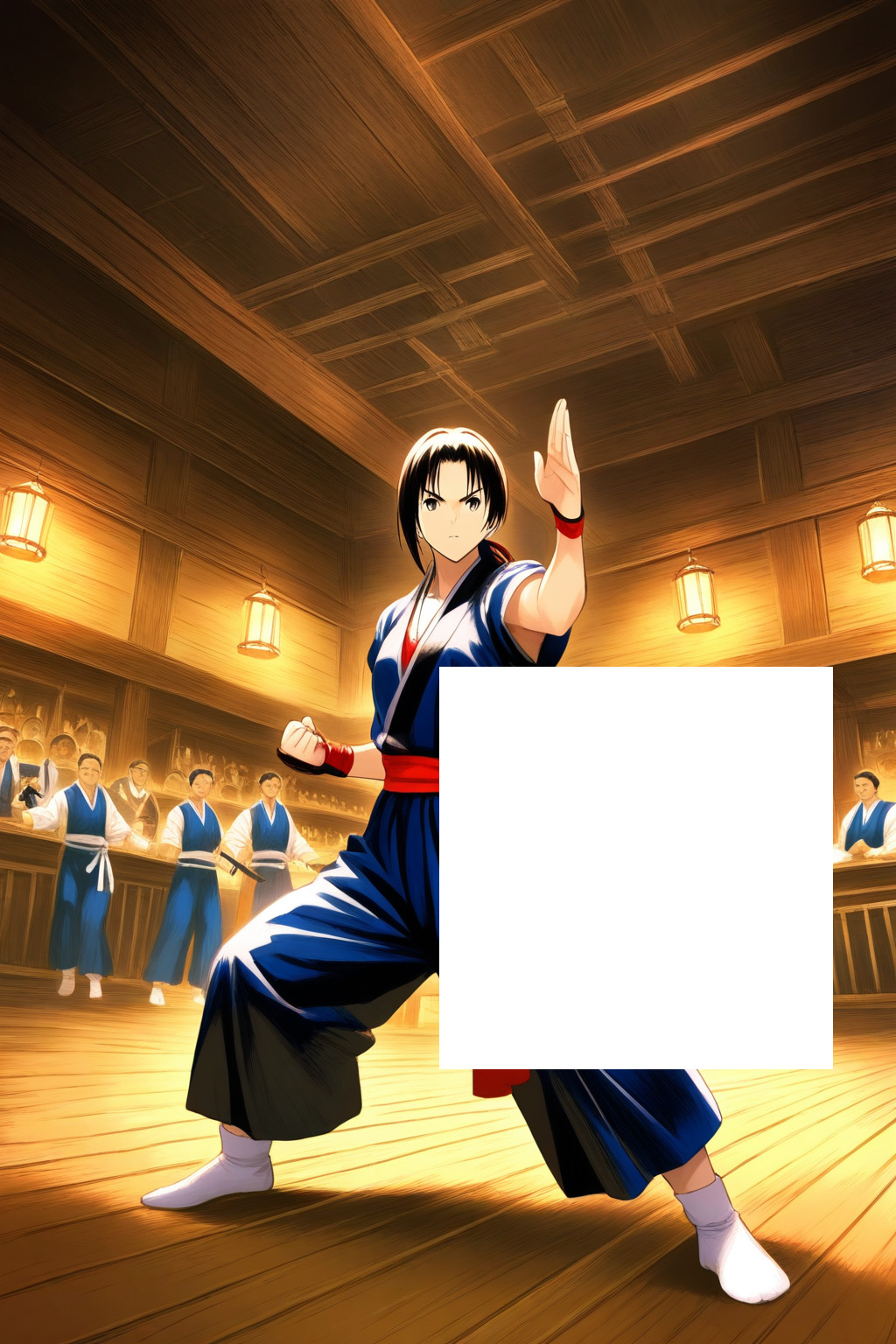}
    \end{minipage}\hfill
    \begin{minipage}{0.5\textwidth}
    \centering
    \includegraphics[width=\textwidth,height=8cm,keepaspectratio]{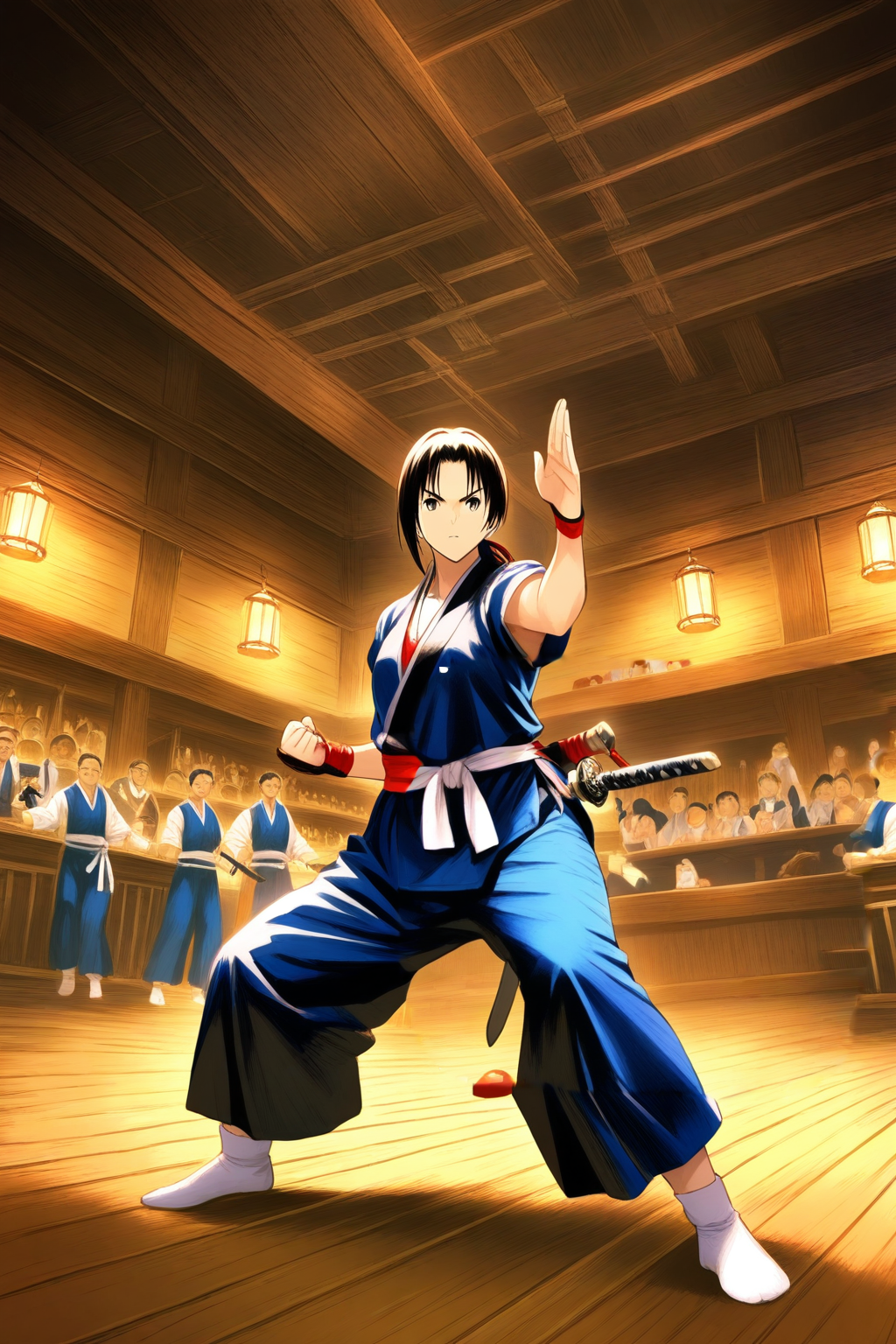}
    \end{minipage}\hfill
    \caption{\textbf{Enhanced Inpainting} As the model's generation capability and prompt control improve, we can also observe significant advancements in its inpainting functionality.}
    \label{fig: inpainting}
\end{figure*}

As Illustrious's prompt control capabilities have improved, it has become capable of supporting powerful image generation. Based on this, we conducted various experiments not only on text-to-image generation but also on image-to-image generation. One of the most interesting findings was that as the model's image generation abilities improved, so did its inpainting capabilities. To demonstrate the improvements in inpainting, we partially cropped and corrupted images, masked the damaged areas, and then applied inpainting using Illustrious. Unlike previous models, which struggled with color or saturation mismatches in inpainting, Illustrious successfully generates images that harmonized seamlessly with the original content. The example image is shown in figure \ref{fig: inpainting}.

\subsection{Dynamic Color Range}
\begin{figure*}[ht!]
    \centering
    \begin{minipage}{0.49\textwidth}
        \includegraphics[width=\textwidth,height=8cm,keepaspectratio]{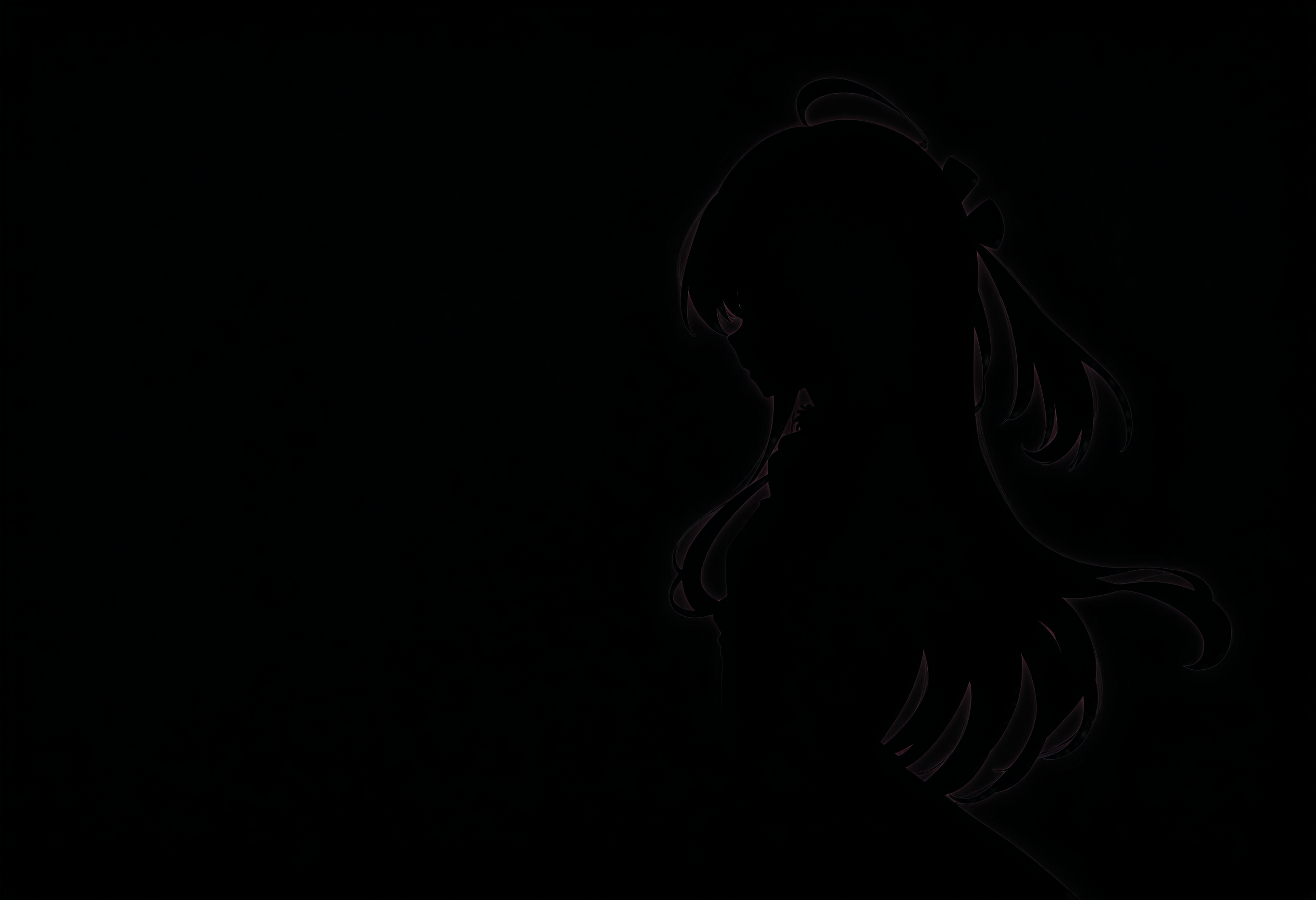}
    \end{minipage}\hfill
    \begin{minipage}{0.49\textwidth}
        \includegraphics[width=\textwidth,height=8cm,keepaspectratio]{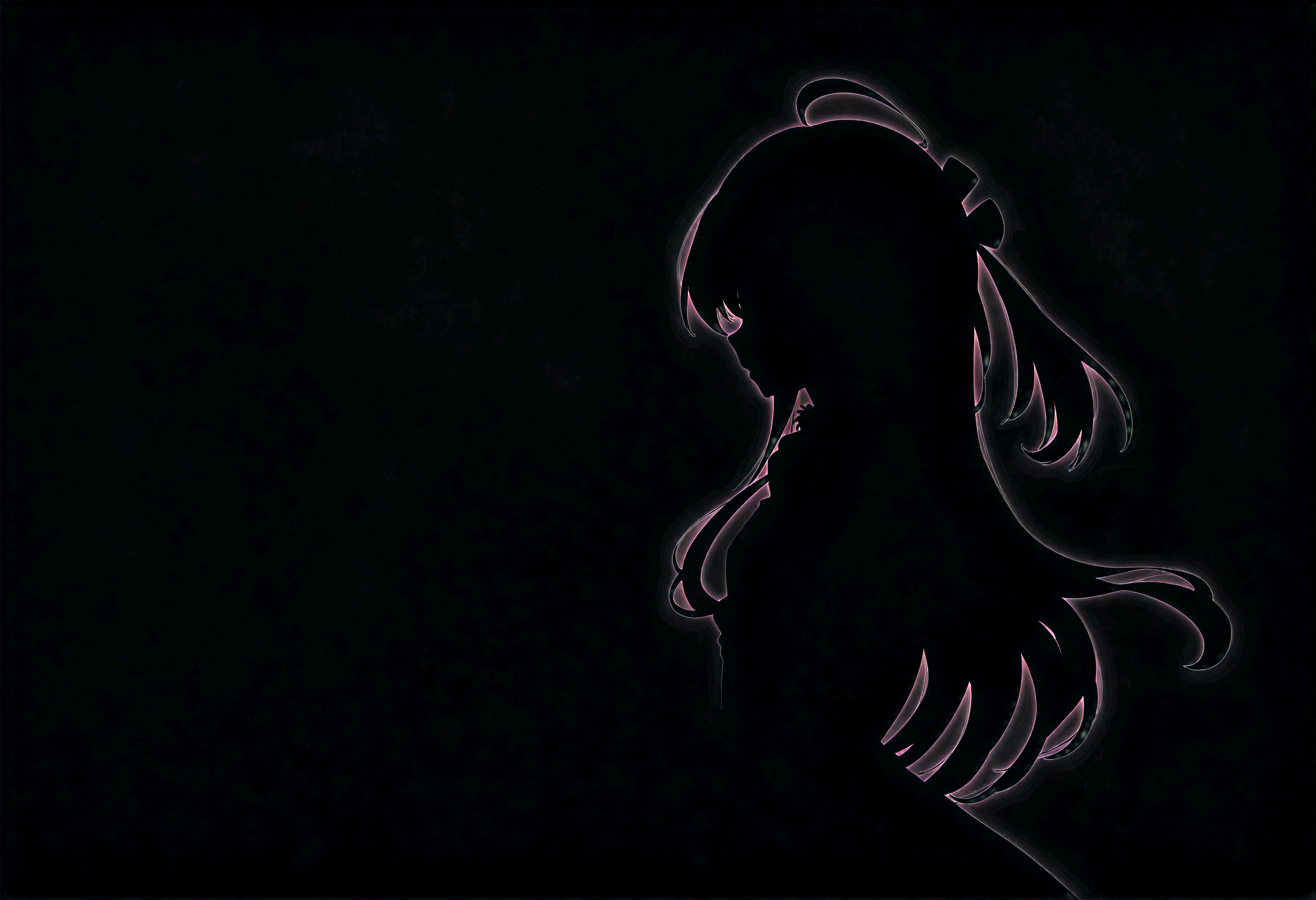}
    \end{minipage}\hfill
\caption{\textbf{Dynamic Color Range} Our model can adjust brightness through silhouette generation and similar techniques. Left is the original image generated from our model. Right is the same Image but upper the brightness 0 to 230.}
\label{fig:dynamic-color-range}
\end{figure*}
Illustrious has significantly improved its understanding of color, allowing control over color and brightness through prompts. In particular, its understanding of brightness has significantly improved. It successfully generates images with colors that are present even at very low brightness levels. We generated images with low brightness and then increased the brightness to demonstrate that the subject could clearly form a silhouette. The example images is shown in figure \ref{fig:dynamic-color-range}.

\subsection{Multi Level Captions}
\begin{figure*}
\centering
{\includegraphics[width=\textwidth]{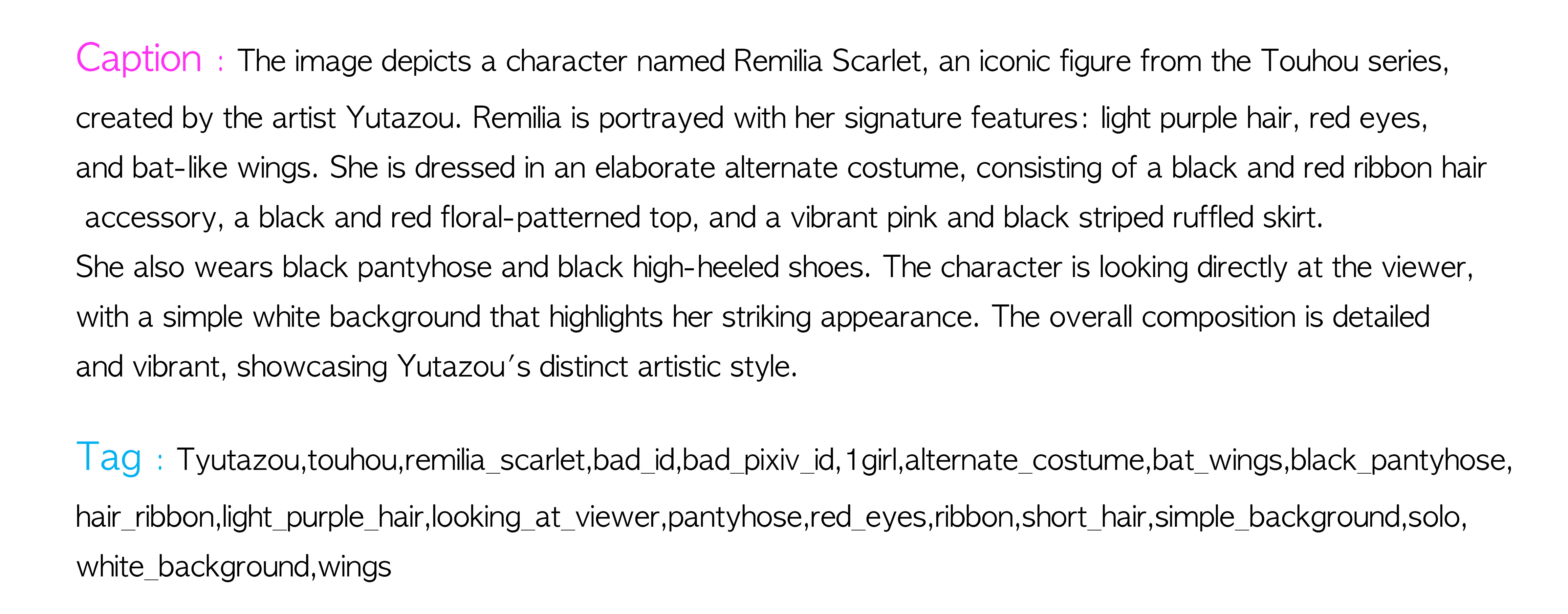}}
\caption{\textbf{Example of Multi Caption}.}
\label{fig:multicaption}
\end{figure*}

Starting from Illustrious v2.0, Multi Level Captions has been introduced. We realized that it is difficult to control multiple objects simultaneously through prompts using tagging alone. \cite{Multicaptionsynthetic} Even when grouping the sequence of tags or the subcomponents of objects, expressing context solely through tagging proved to be quite challenging. Therefore, it is crucial to tag in a way that makes the context easily understandable through natural language. However, having humans manually caption large amounts of data is labor-intensive and has its limitations. At the same time, we could not abandon the advantages of tagging, so we implemented Multi-Captioning for images. Multi-Captioning involves assigning multiple captions to a single image likes natural language and tags. In the future, we plan to increase the number of captions to not only provide detailed descriptions of the image but also include context and narrative elements. The example of multi caption is shown in figure \ref{fig:multicaption}.

\subsection{Padding token wise analysis}
\begin{figure*}
\centering
{\includegraphics[width=\textwidth]{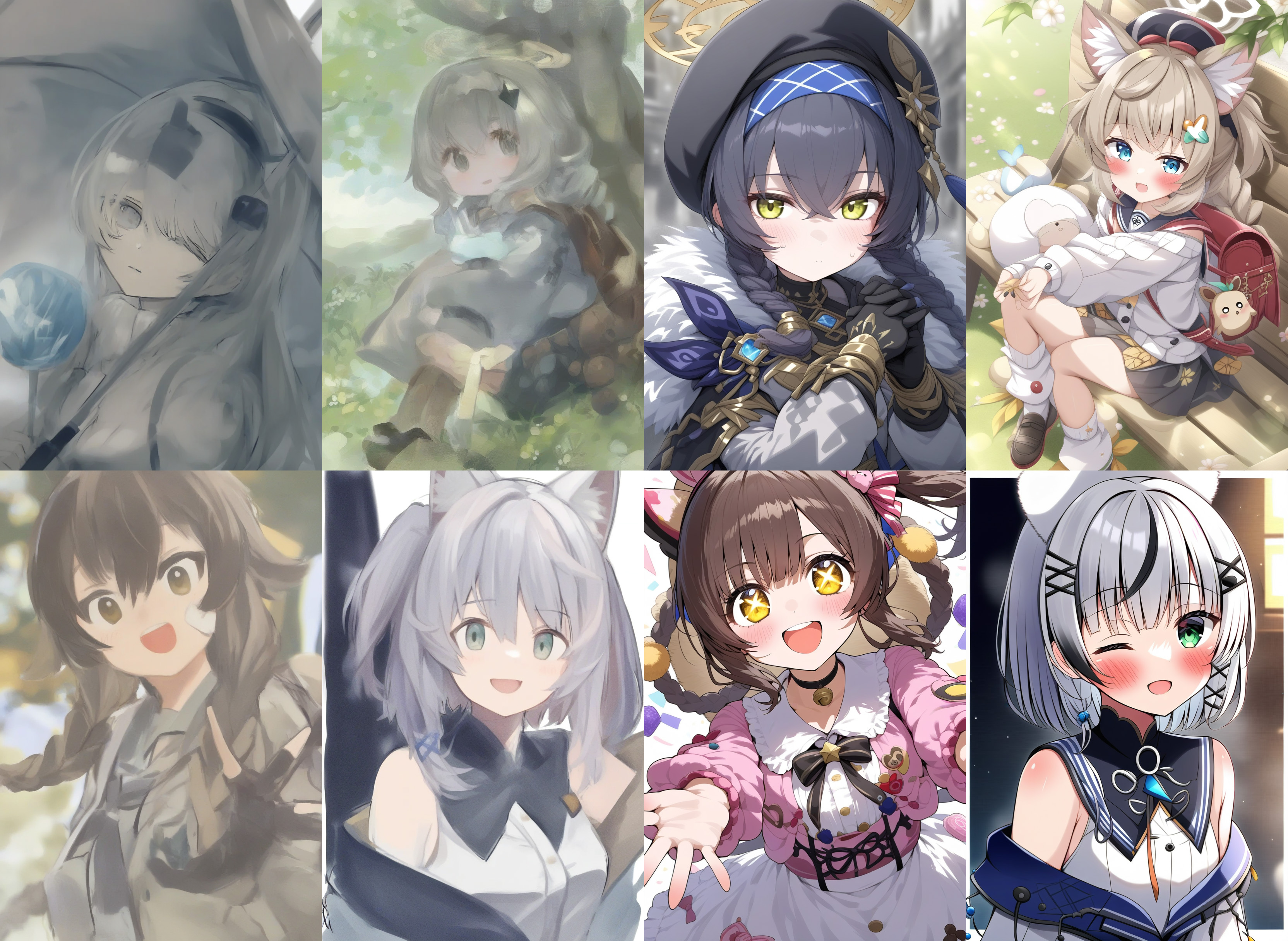}}
\caption{The intensive padding token being used in CFG, causes problem since padding token was not utilized via masked loss. Left, with 2 tokens + 75 tokens padding, right, no CFG. The phenomenon is reduced when minimal padding token is used.}
\label{fig:paddingproblem}
\end{figure*}

We find that allowing padding tokens to be trained can cause multiple problems.
During training, text encoder outputs must be padded to be packed in batch. This makes padding token usage in CFG setups problematic with imbalanced token lengths, as it retains significant composition knowledge unlike different models. We recommend masked loss to overcome this problem in future training. We show the example in Figure \ref{fig:paddingproblem}.

\newpage
\subsection{Further finetuning recommendations}
We found that the Illustrious XL Text Encoders are stably converged - the text encoders are interchangable without major issues, despite of current tradition of not tuning text encoders for knowledge conservation and memory requirements.
Despite of our method's empirical success, we do \textbf{not} recommend to finetune text encoder, unless datasets are sufficiently large enough to counter possible catastrophic forgetting issues.

As noted previously, we found that character learning trend fluctuates with lower batch sizes, whilst higher batch size stabilizes its forgetting phenomenon. Even larger batch size may be required for sparse concepts.

\subsection{Safety control and Red-Teaming}
Image dataset domain is abstract and not well researched, publicly available solutions and systems, and its detail lacks, which makes user uncontrollable from unwanted content generation.
Following waluigi dillema, we instead finetune with strict control condition to make model understand the concepts separately, then utilize LECO \cite{gandikota2023conceptslidersloraadaptors} method-based approach, allowing safety control over provocative generations, as released in GUIDED variants, suggested as reference  \cite{safetycontrol}. However, we also note here that simple control can be achieved by rating tokens, inputting "general" in prompt conditioning.

\section{Model Compare}
\begin{figure*}
    \centering
    \includegraphics[width=0.9\linewidth]{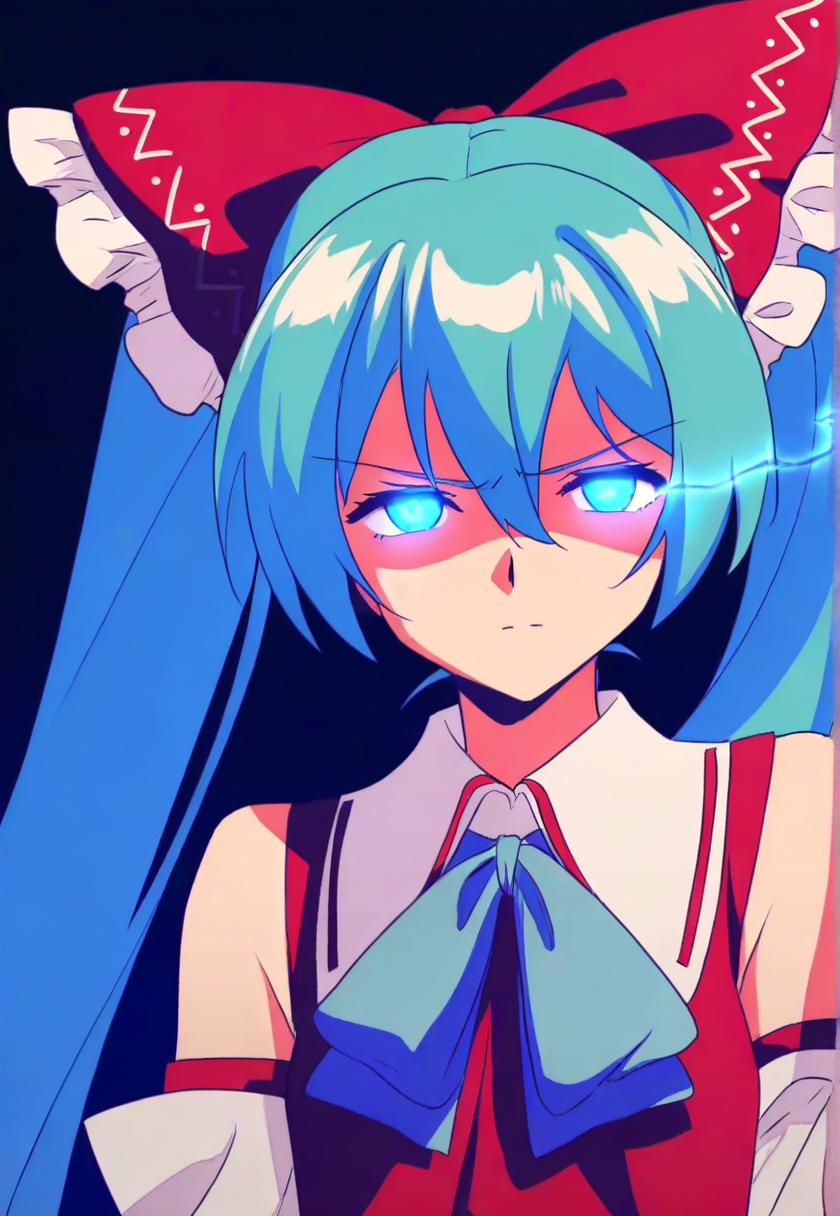}
    \caption{\textbf{Hatsune miku, cosplaying hakurei reimu, in 90s animation style, with glowing eyes}, generated in Illustrious v2.0 with 840$\times$1216 resolution.}
    \label{fig:multiconcept}
\end{figure*}
\subsection{Illustrious Qualitative Images}
\subsubsection{Illustrious v0.1}
\begin{figure*}[htb]
    \centering
    \begin{minipage}{0.33\textwidth}
        \includegraphics[width=\textwidth,height=8cm,keepaspectratio]{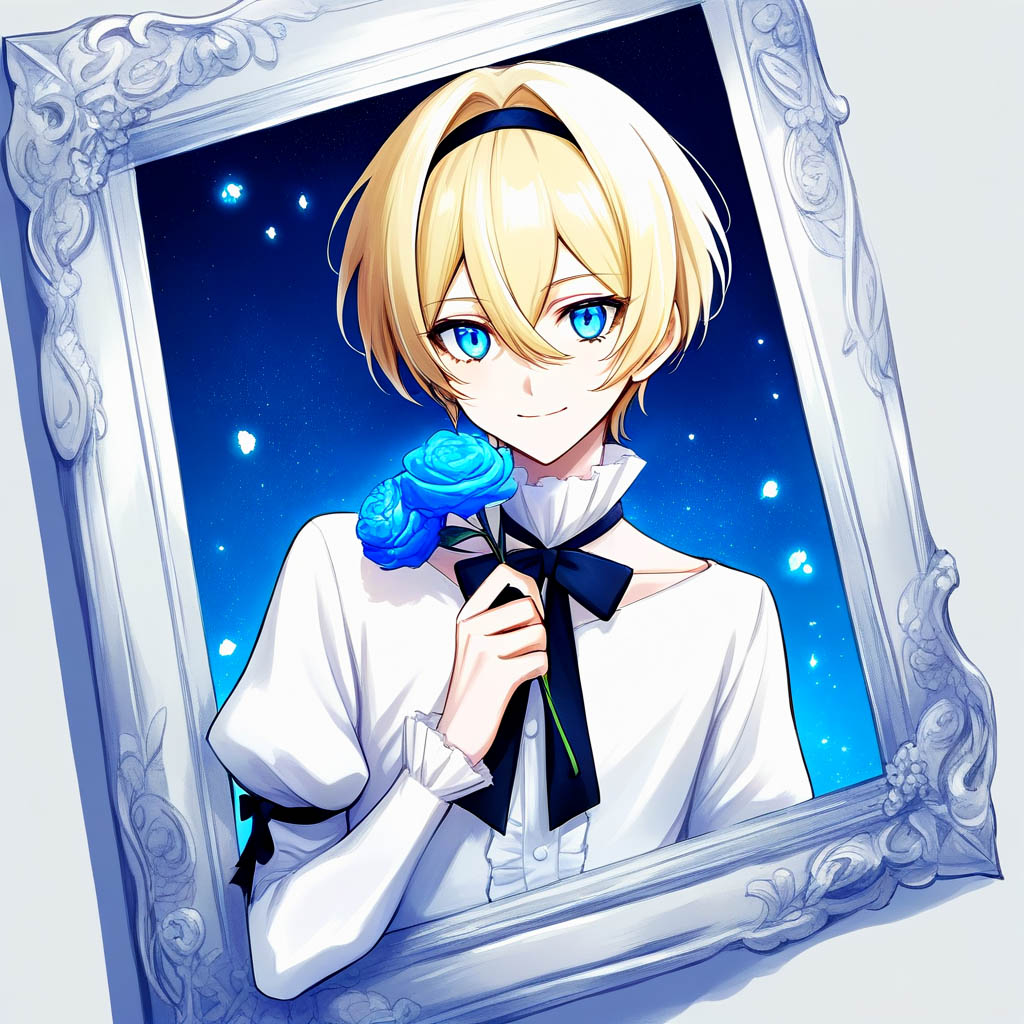}
    \end{minipage}\hfill
    \begin{minipage}{0.33\textwidth}
        \includegraphics[width=\textwidth,height=8cm,keepaspectratio]{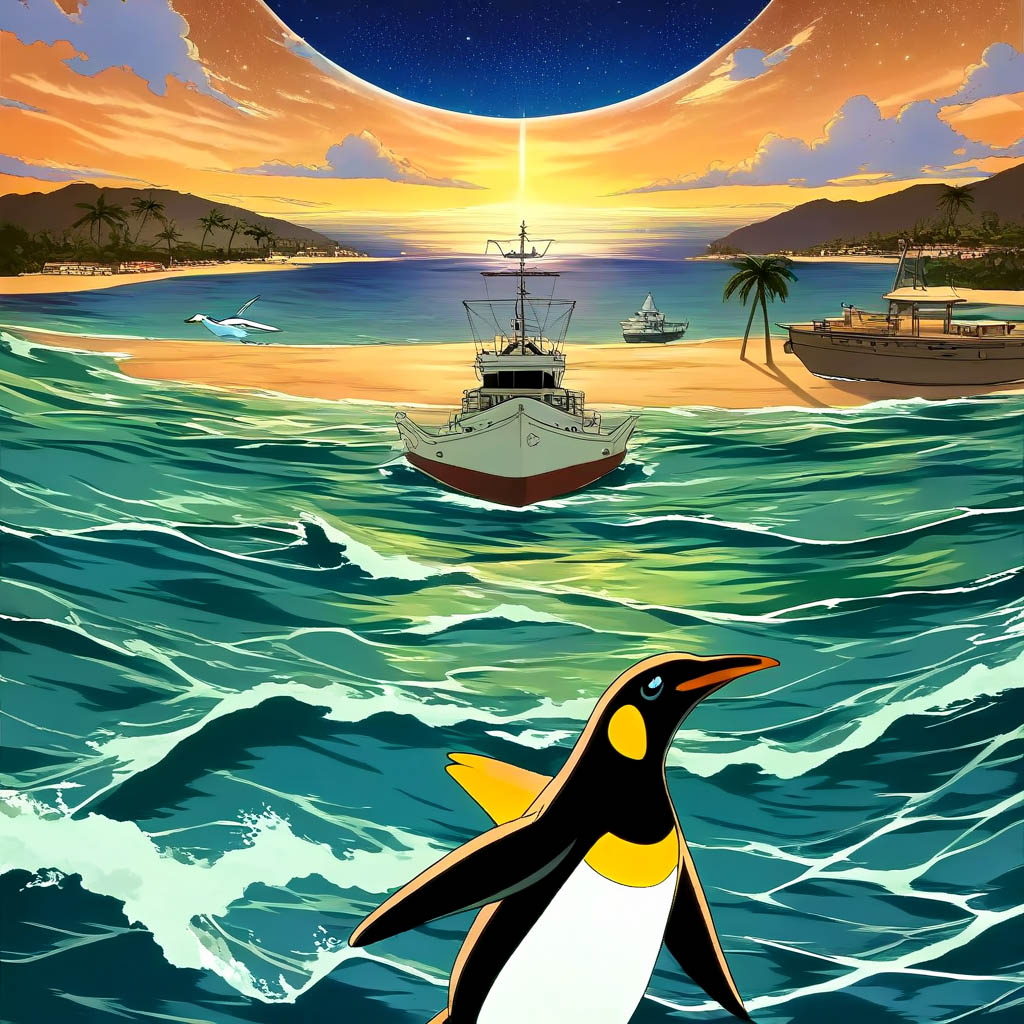}
    \end{minipage}\hfill
    \begin{minipage}{0.33\textwidth}
        \includegraphics[width=\textwidth,height=8cm,keepaspectratio]{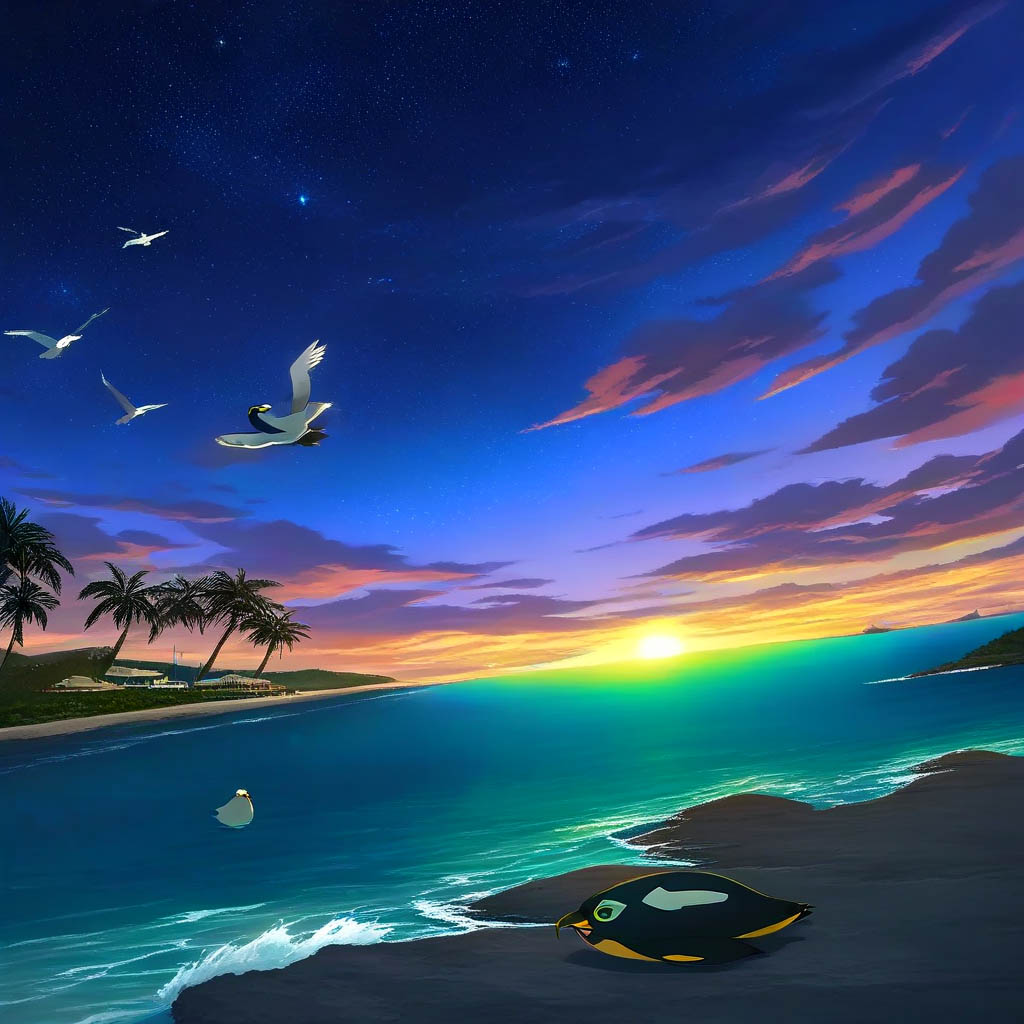}
    \end{minipage}
    
    
    \begin{minipage}{0.33\textwidth}
        \includegraphics[width=\textwidth,height=8cm,keepaspectratio]{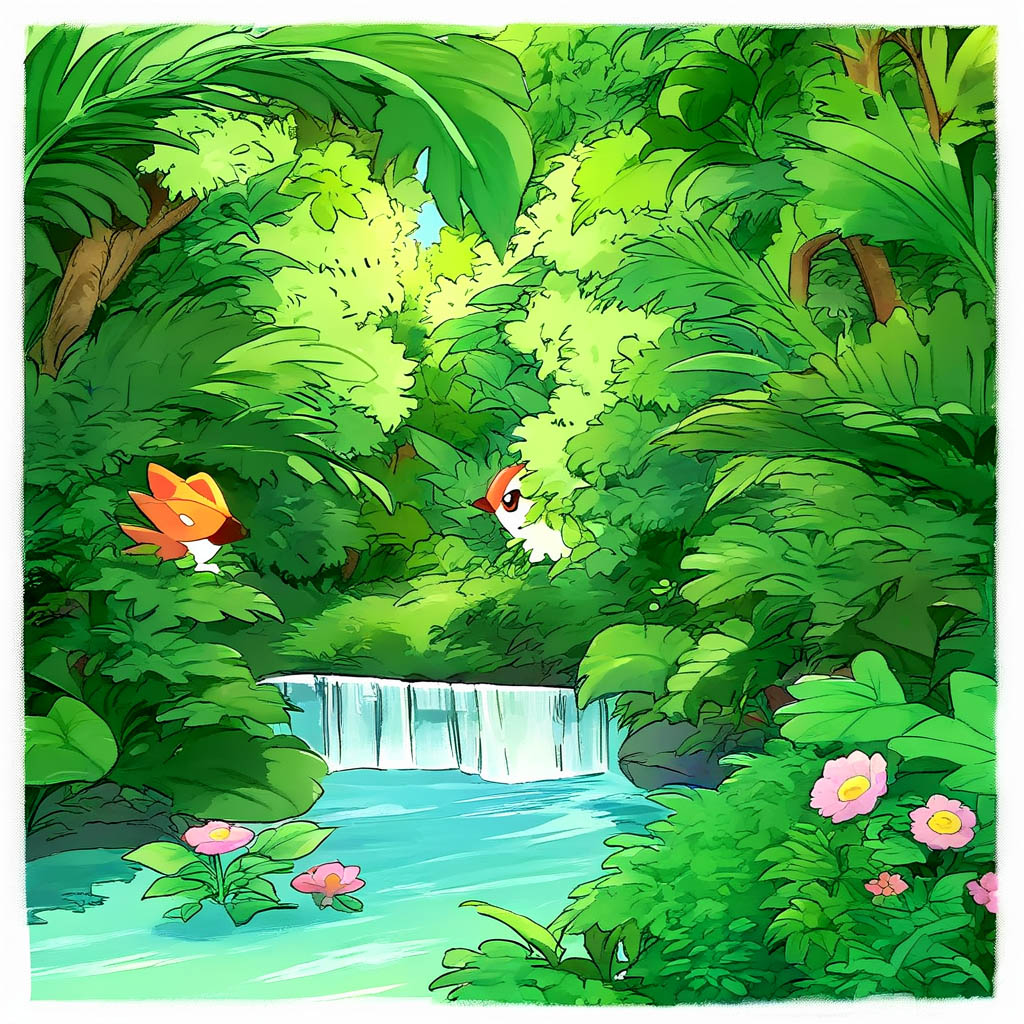}
    \end{minipage}\hfill
    \begin{minipage}{0.33\textwidth}
        \includegraphics[width=\textwidth,height=8cm,keepaspectratio]{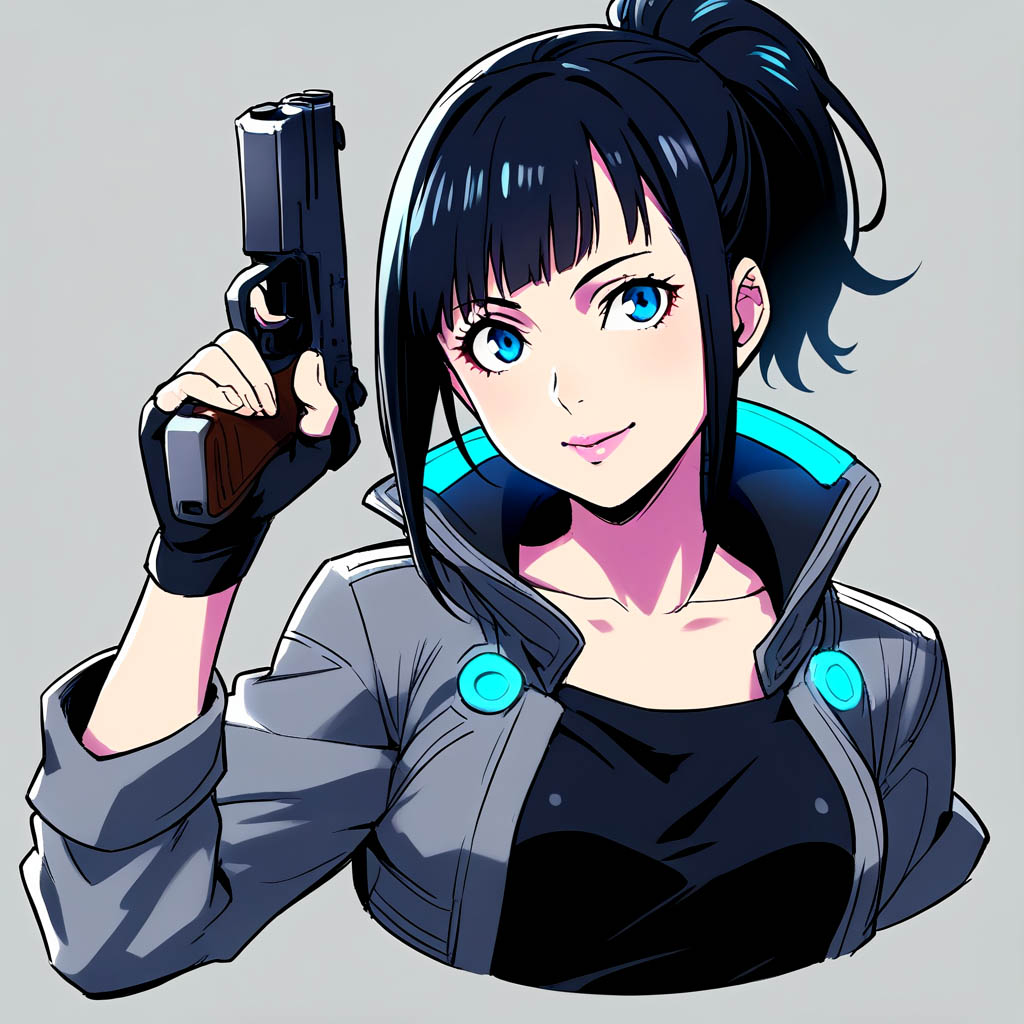}
    \end{minipage}\hfill
    \begin{minipage}{0.33\textwidth}
        \includegraphics[width=\textwidth,height=8cm,keepaspectratio]{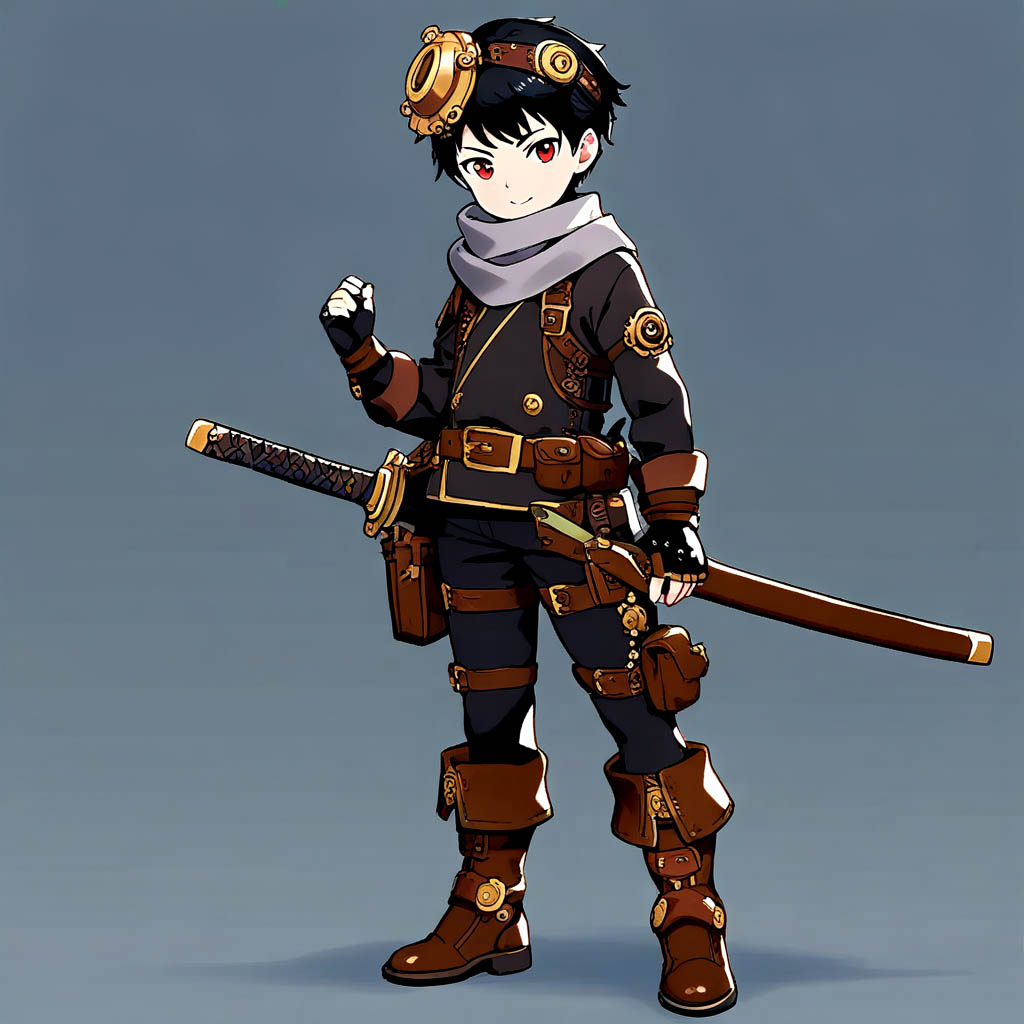}
    \end{minipage}


    \begin{minipage}{0.33\textwidth}
        \includegraphics[width=\textwidth,height=8cm,keepaspectratio]{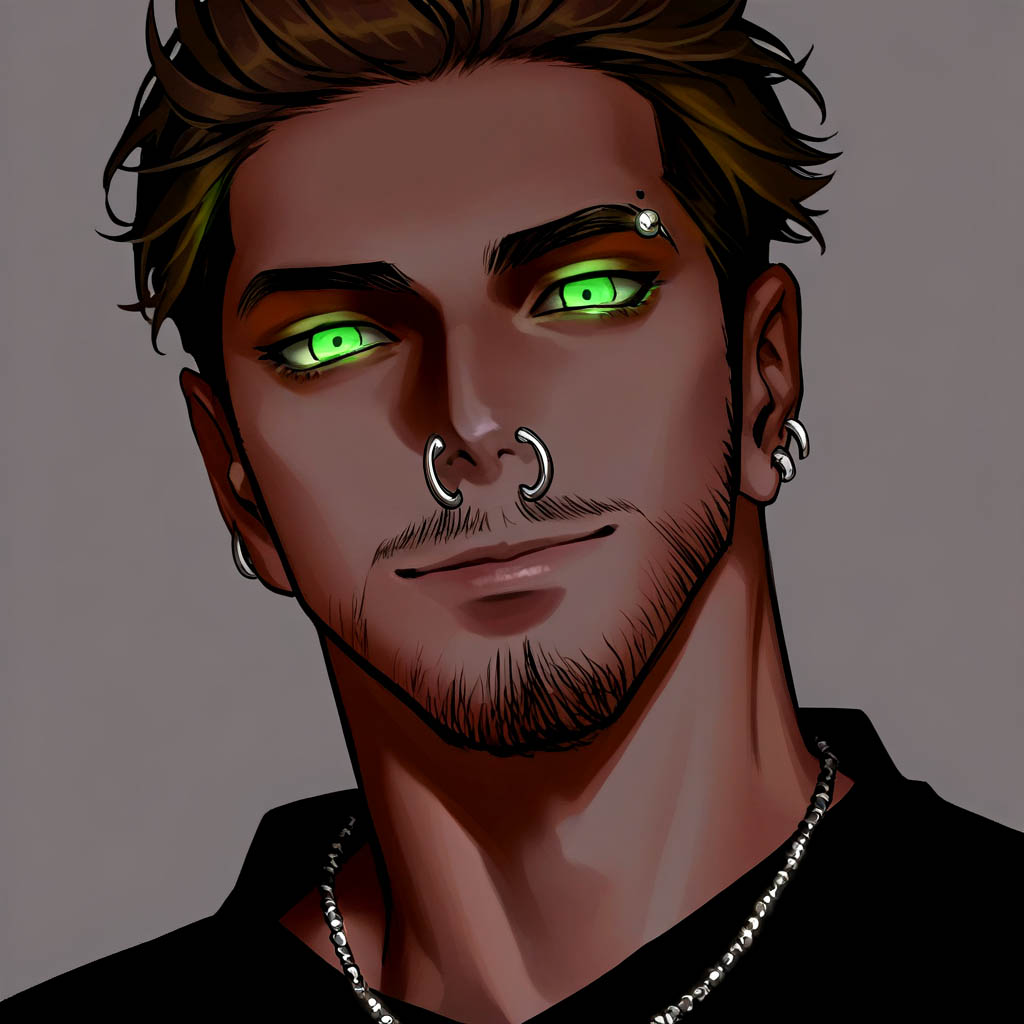}
    \end{minipage}\hfill
    \begin{minipage}{0.33\textwidth}
        \includegraphics[width=\textwidth,height=8cm,keepaspectratio]{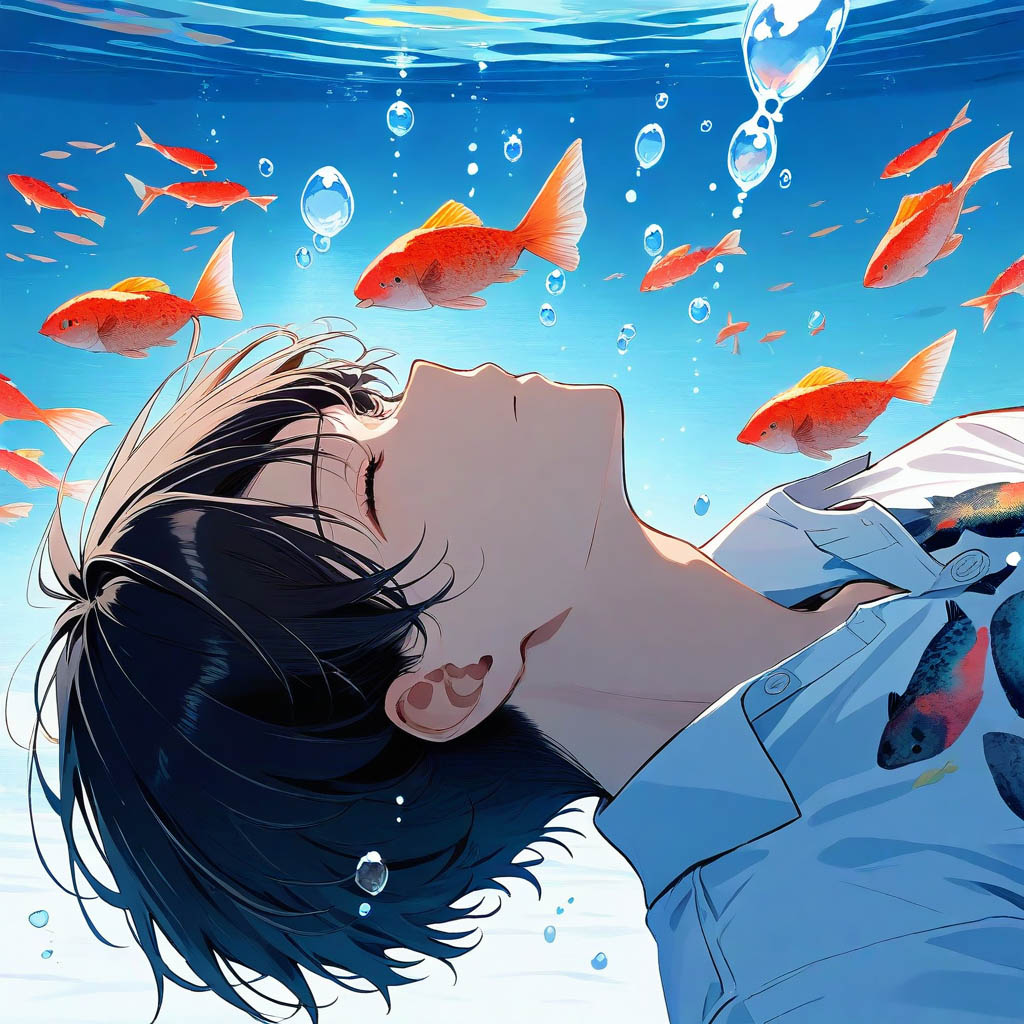}
    \end{minipage}\hfill
    \begin{minipage}{0.33\textwidth}
        \includegraphics[width=\textwidth,height=8cm,keepaspectratio]{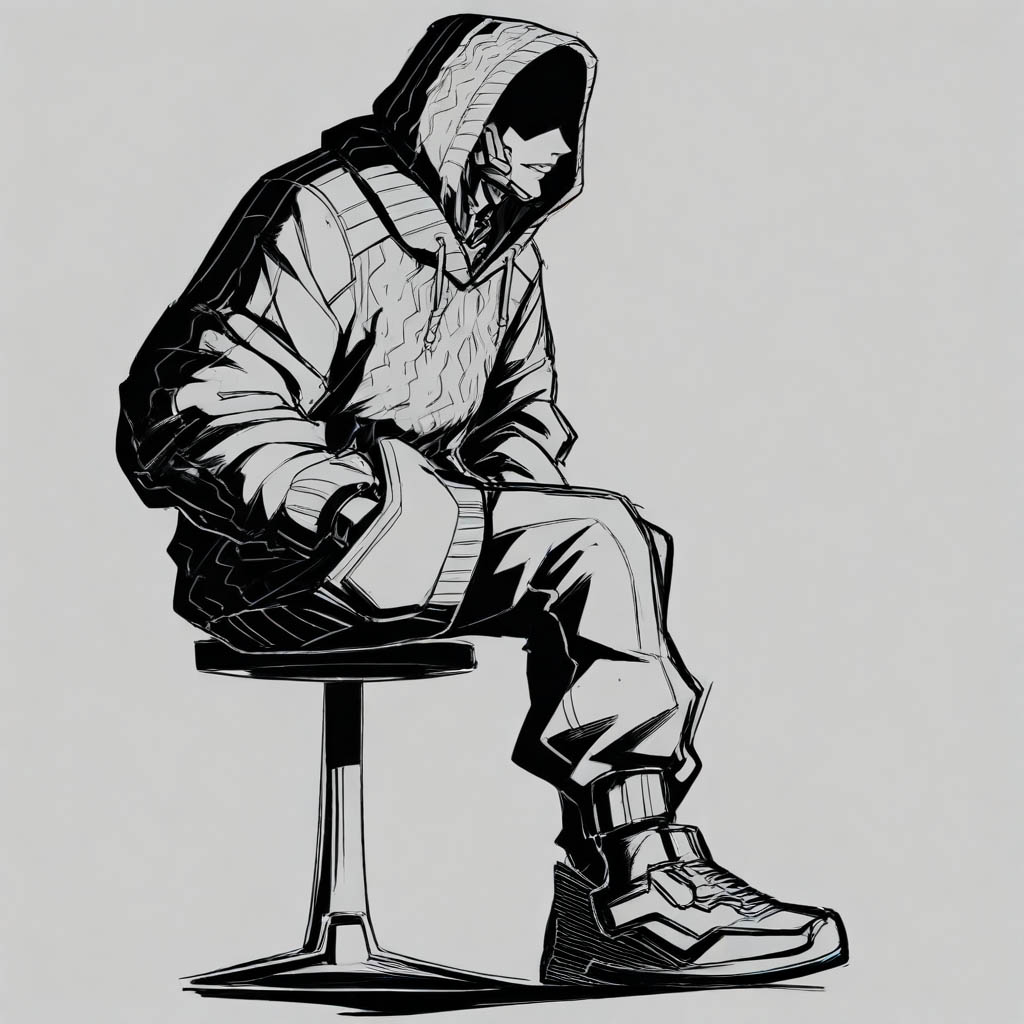}
    \end{minipage}\hfill
    
    \begin{minipage}{0.33\textwidth}
        \includegraphics[width=\textwidth,height=8cm,keepaspectratio]{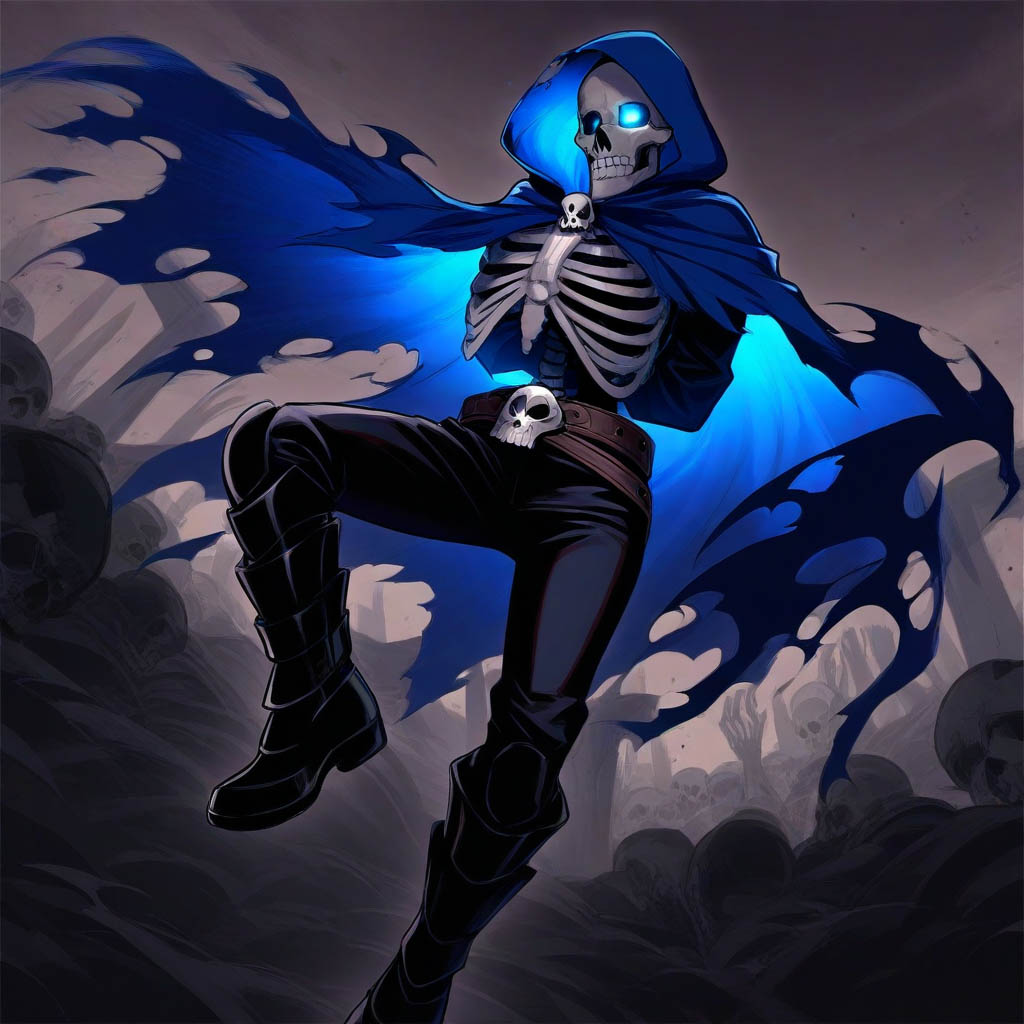}
    \end{minipage}\hfill
    \begin{minipage}{0.33\textwidth}
        \includegraphics[width=\textwidth,height=8cm,keepaspectratio]{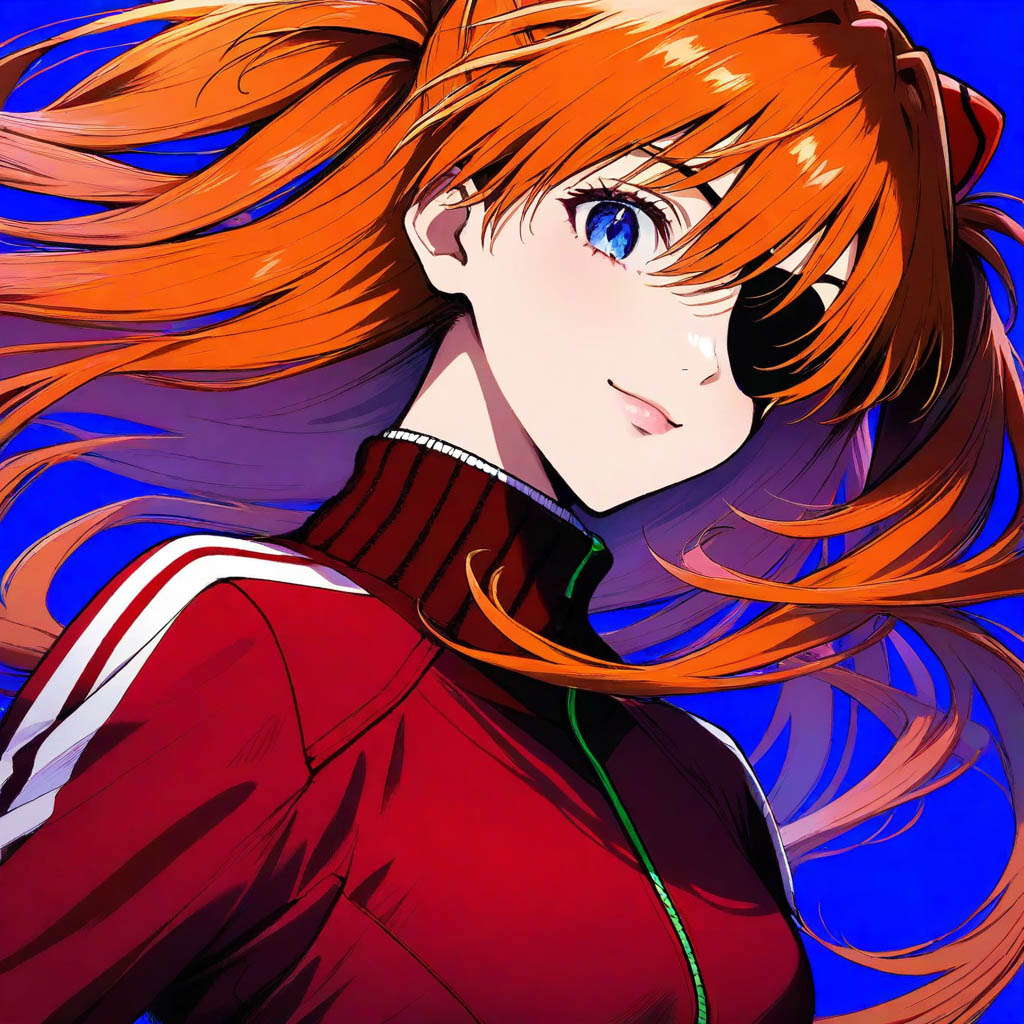}
    \end{minipage}\hfill
    \begin{minipage}{0.33\textwidth}
        \includegraphics[width=\textwidth,height=8cm,keepaspectratio]{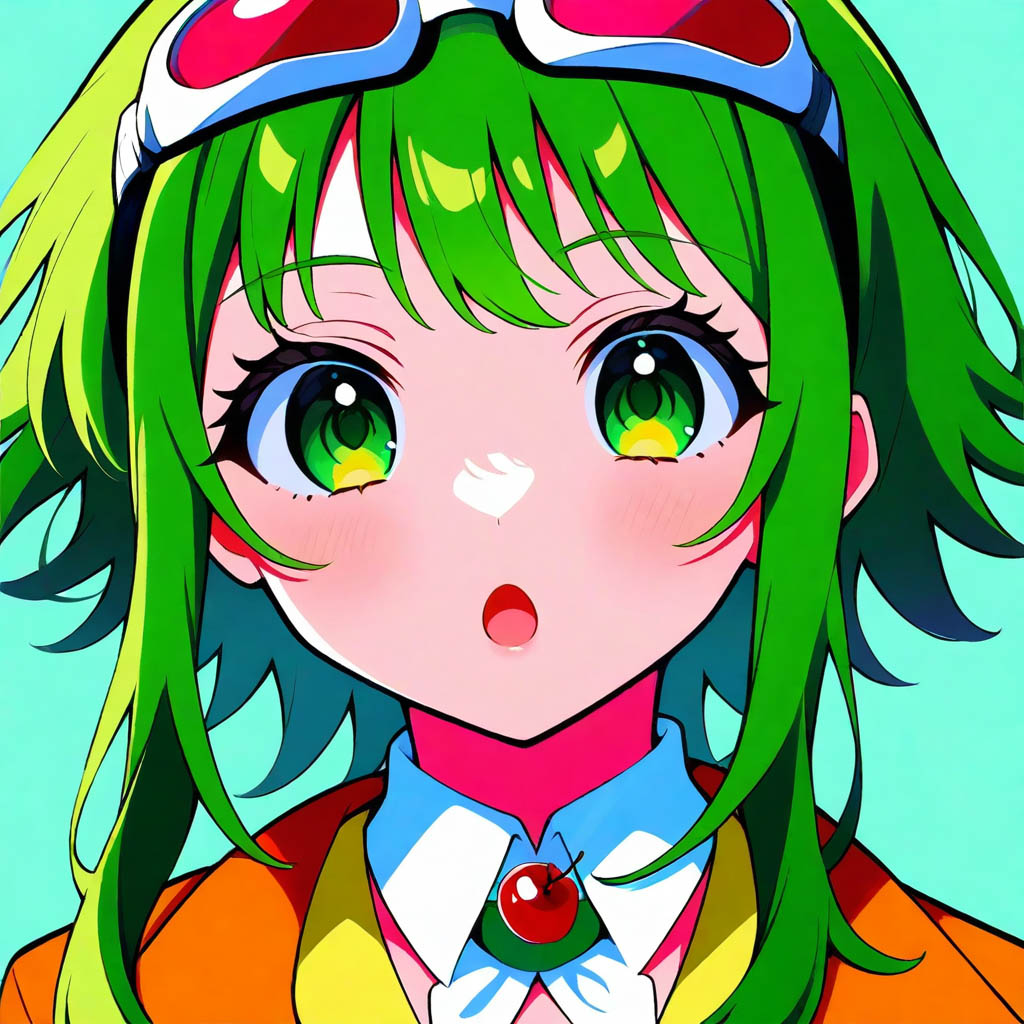}
    \end{minipage}\hfill
    \caption{\textbf{High-quality samples from Illustrious v0.1.} Illustrious v0.1 can generate creative pictures.}
    \label{fig:Samples_v0.1}
\end{figure*}

\newpage
Illustrious v0.1's sample image is depicted as Figure \ref{fig:Samples_v0.1}.

\newpage
\subsubsection{Illustrious v1.0}
\begin{figure*}
\centering
{\includegraphics[width=\textwidth]{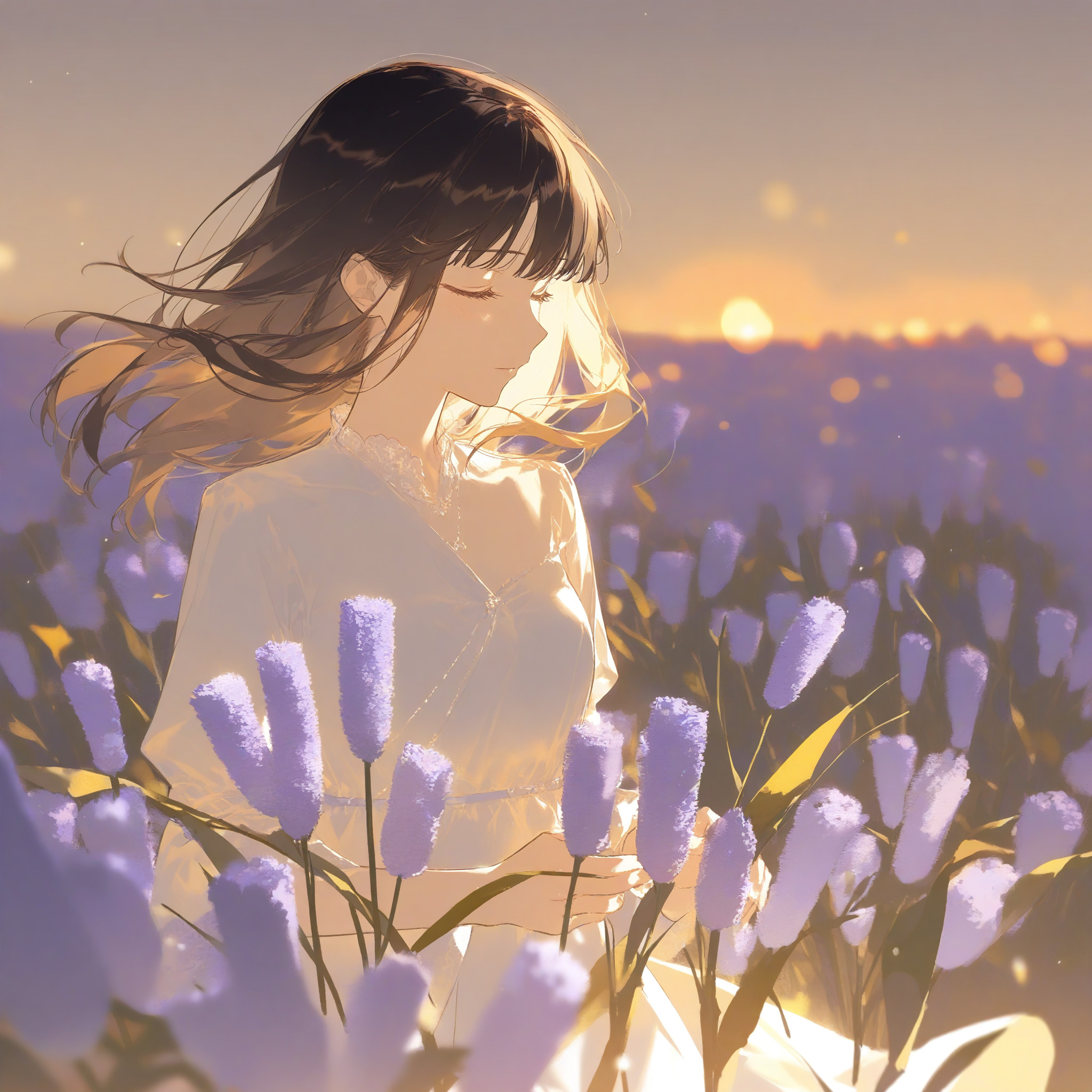}}
\caption{\textbf{High resolution samples from Illustrious v1.0.} Illustrious v1.0 can generate the high resolution images. This image is 2048 $\times$ 2048 pixels with no upscale.}
\label{fig:v1.0_sample}
\end{figure*}

\begin{figure*}[htb]
    \centering
    \begin{minipage}{0.33\textwidth}
        \includegraphics[width=\textwidth,height=8cm,keepaspectratio]{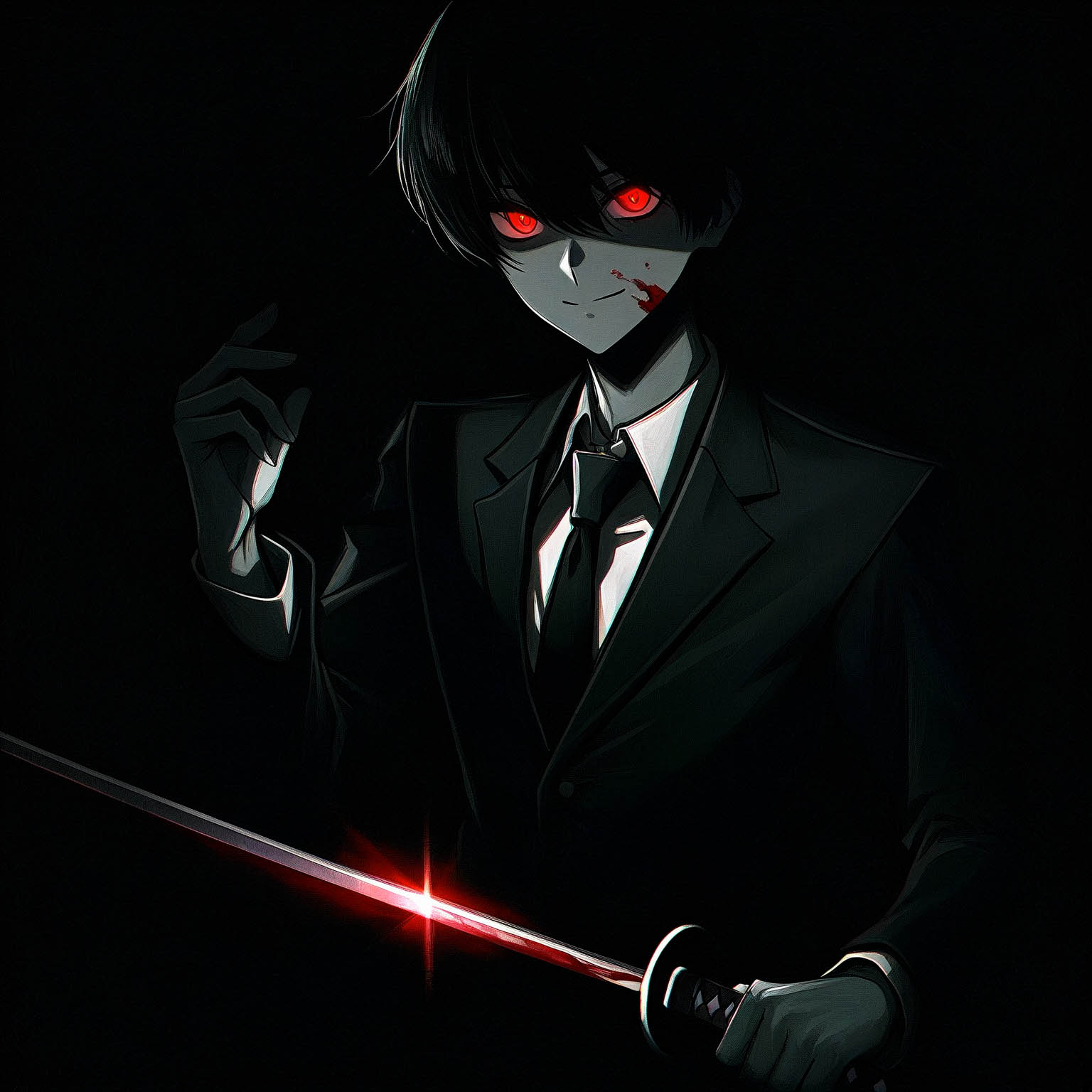}
    \end{minipage}\hfill
    \begin{minipage}{0.33\textwidth}
        \includegraphics[width=\textwidth,height=8cm,keepaspectratio]{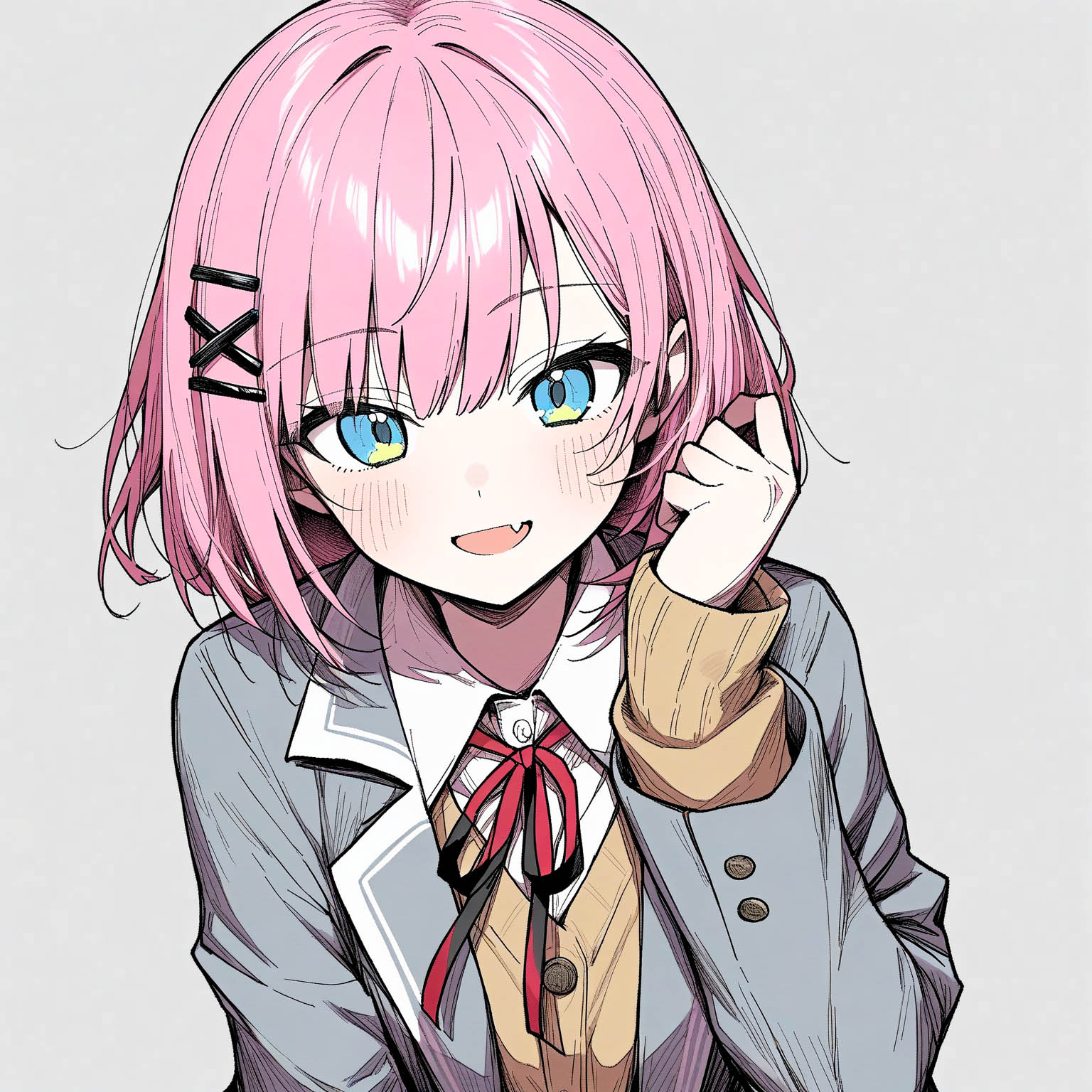}
    \end{minipage}\hfill
    \begin{minipage}{0.33\textwidth}
        \includegraphics[width=\textwidth,height=8cm,keepaspectratio]{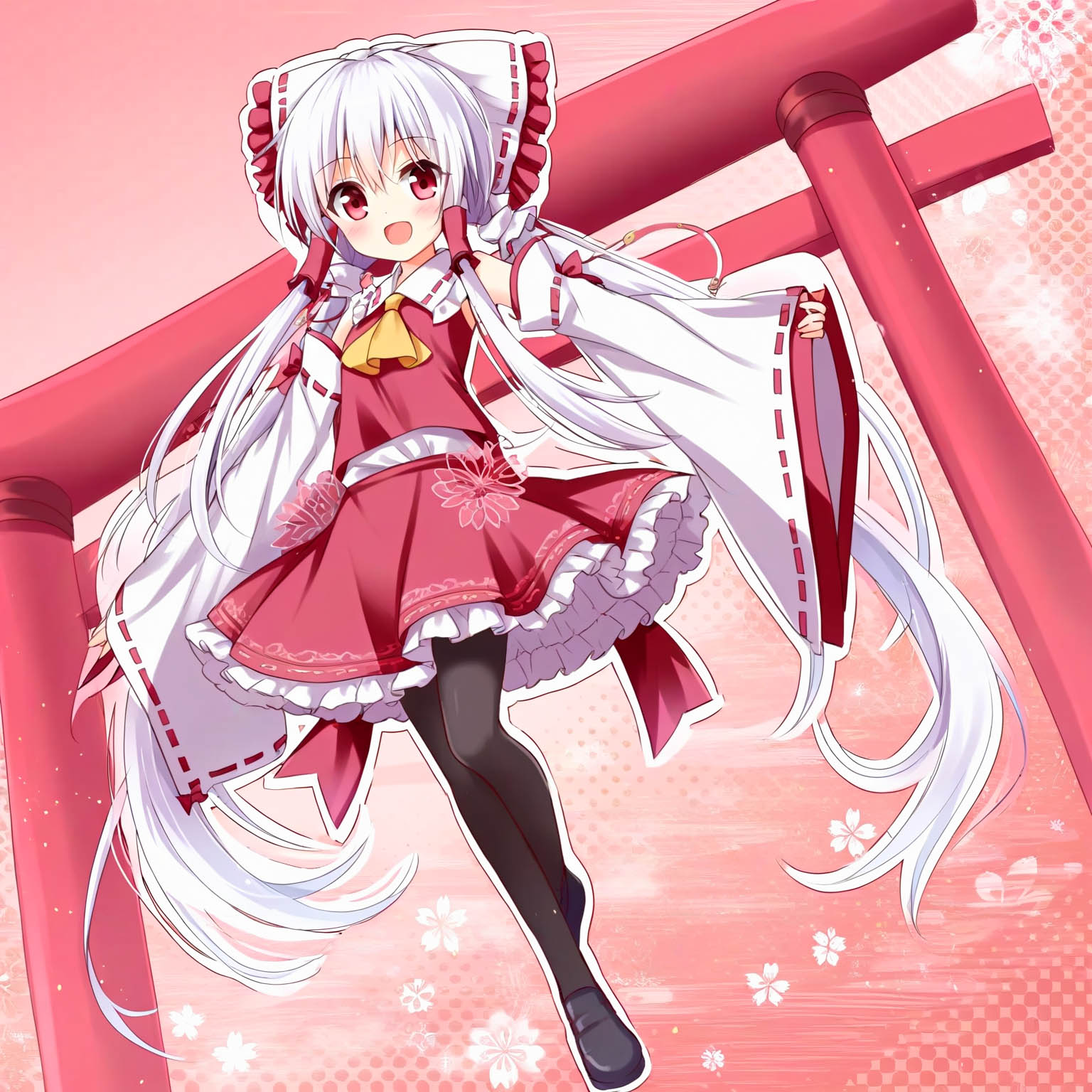}
    \end{minipage}
    
    
    \begin{minipage}{0.33\textwidth}
        \includegraphics[width=\textwidth,height=8cm,keepaspectratio]{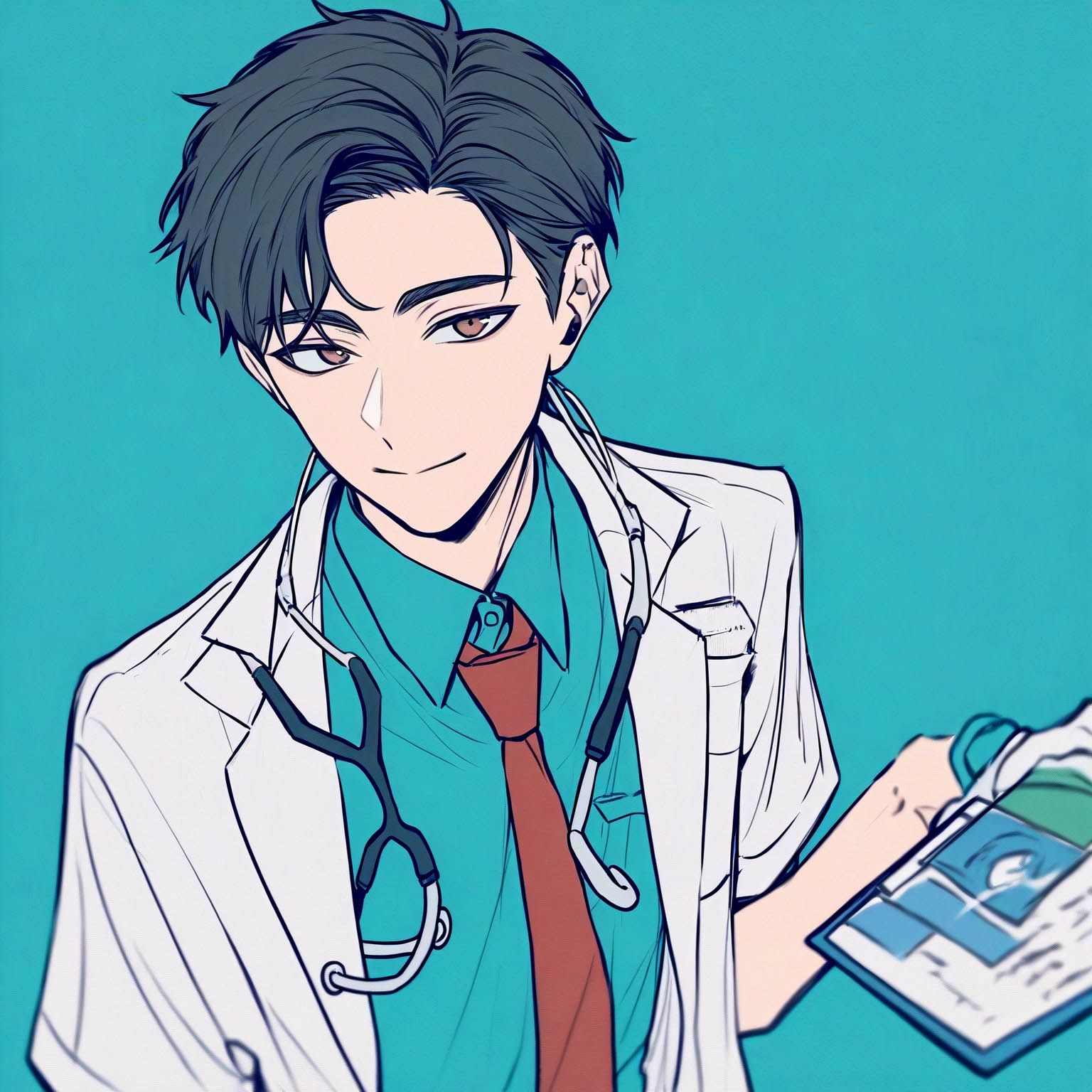}
    \end{minipage}\hfill
    \begin{minipage}{0.33\textwidth}
        \includegraphics[width=\textwidth,height=8cm,keepaspectratio]{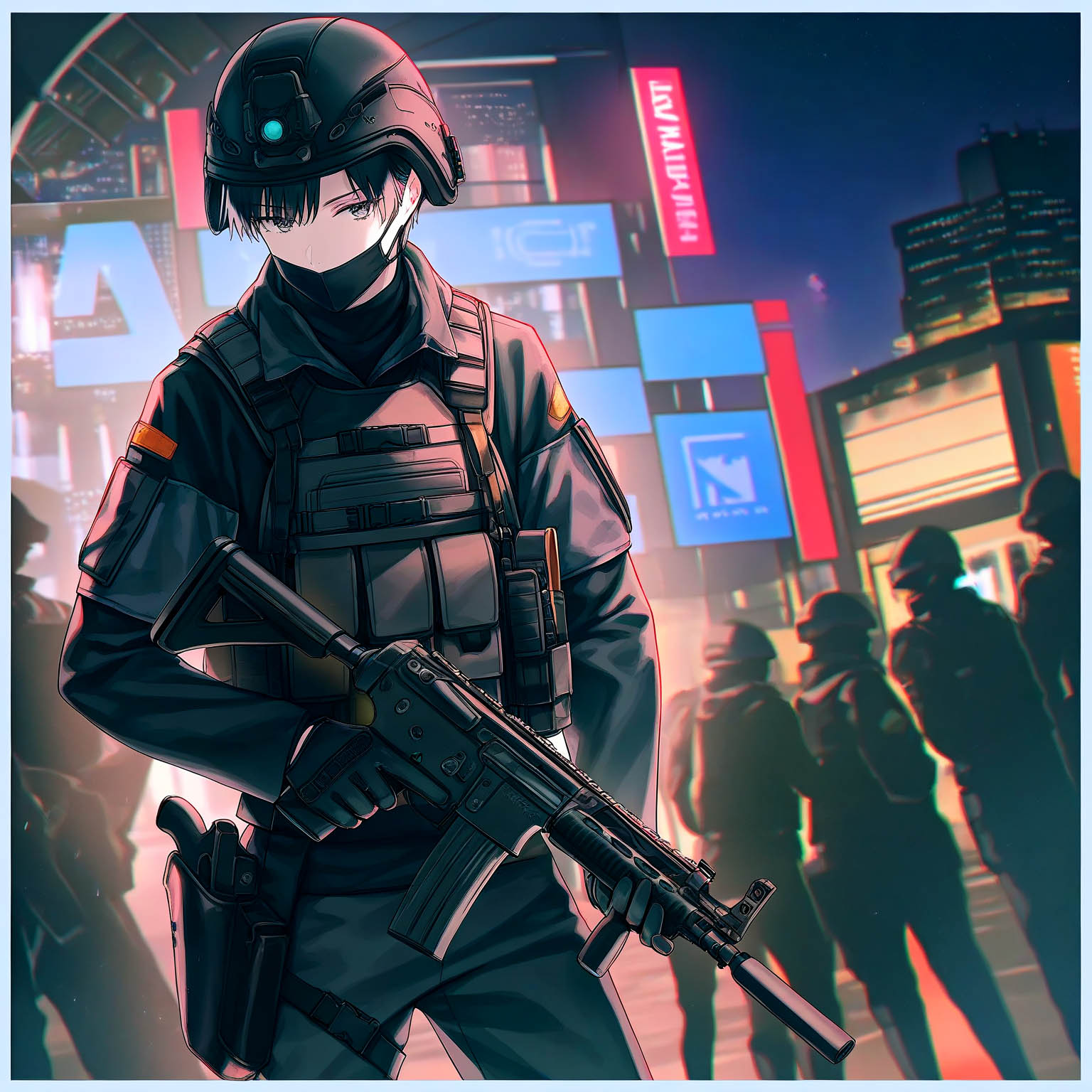}
    \end{minipage}\hfill
    \begin{minipage}{0.33\textwidth}
        \includegraphics[width=\textwidth,height=8cm,keepaspectratio]{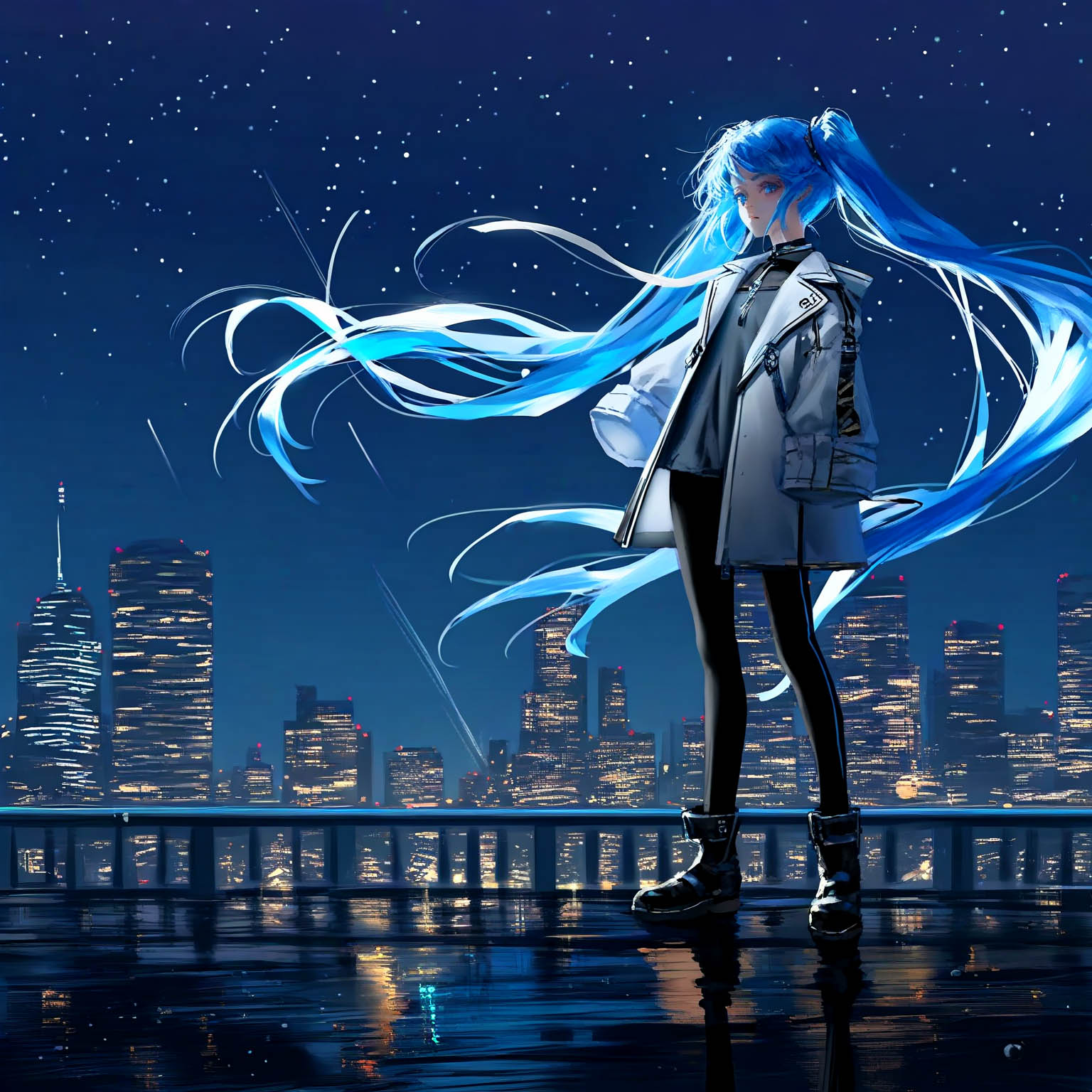}
    \end{minipage}
    \caption{\textbf{High-quality samples from Illustrious v1.0.} Illustrious v1.0 can generate various styles. These images are all 1536 $\times$ 1536 pixels.}
    \label{fig:Samples_v1.0}
\end{figure*}

Illustrious v1.0's sample image is depicted as Figure \ref{fig:v1.0_sample} and \ref{fig:Samples_v1.0}.

\newpage
\subsubsection{Illustrious v1.1}
\begin{figure*}[htb]
    \centering
    \begin{minipage}{0.33\textwidth}
        \includegraphics[width=\textwidth,height=8cm,keepaspectratio]{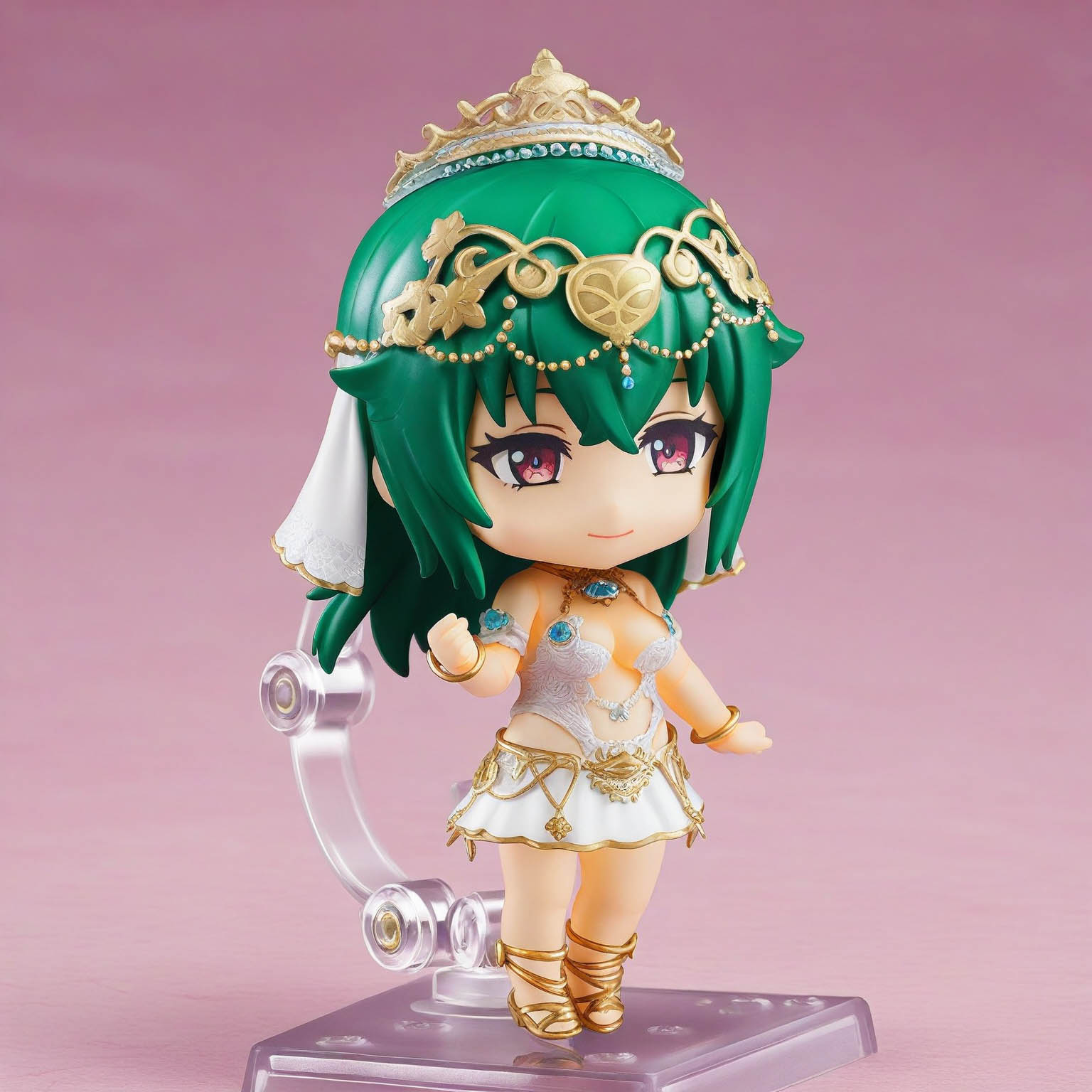}
    \end{minipage}\hfill
    \begin{minipage}{0.33\textwidth}
        \includegraphics[width=\textwidth,height=8cm,keepaspectratio]{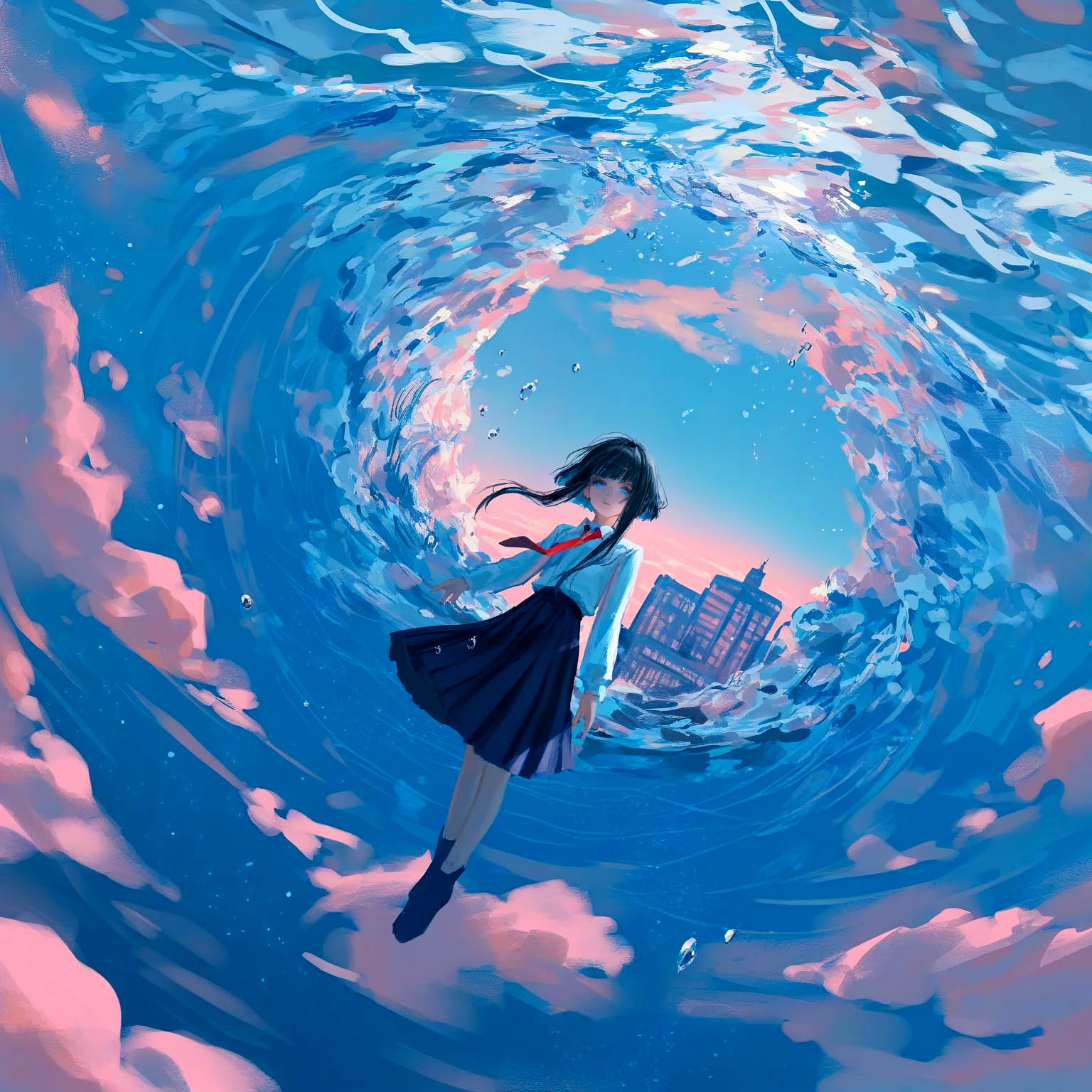}
    \end{minipage}\hfill
    \begin{minipage}{0.33\textwidth}
        \includegraphics[width=\textwidth,height=8cm,keepaspectratio]{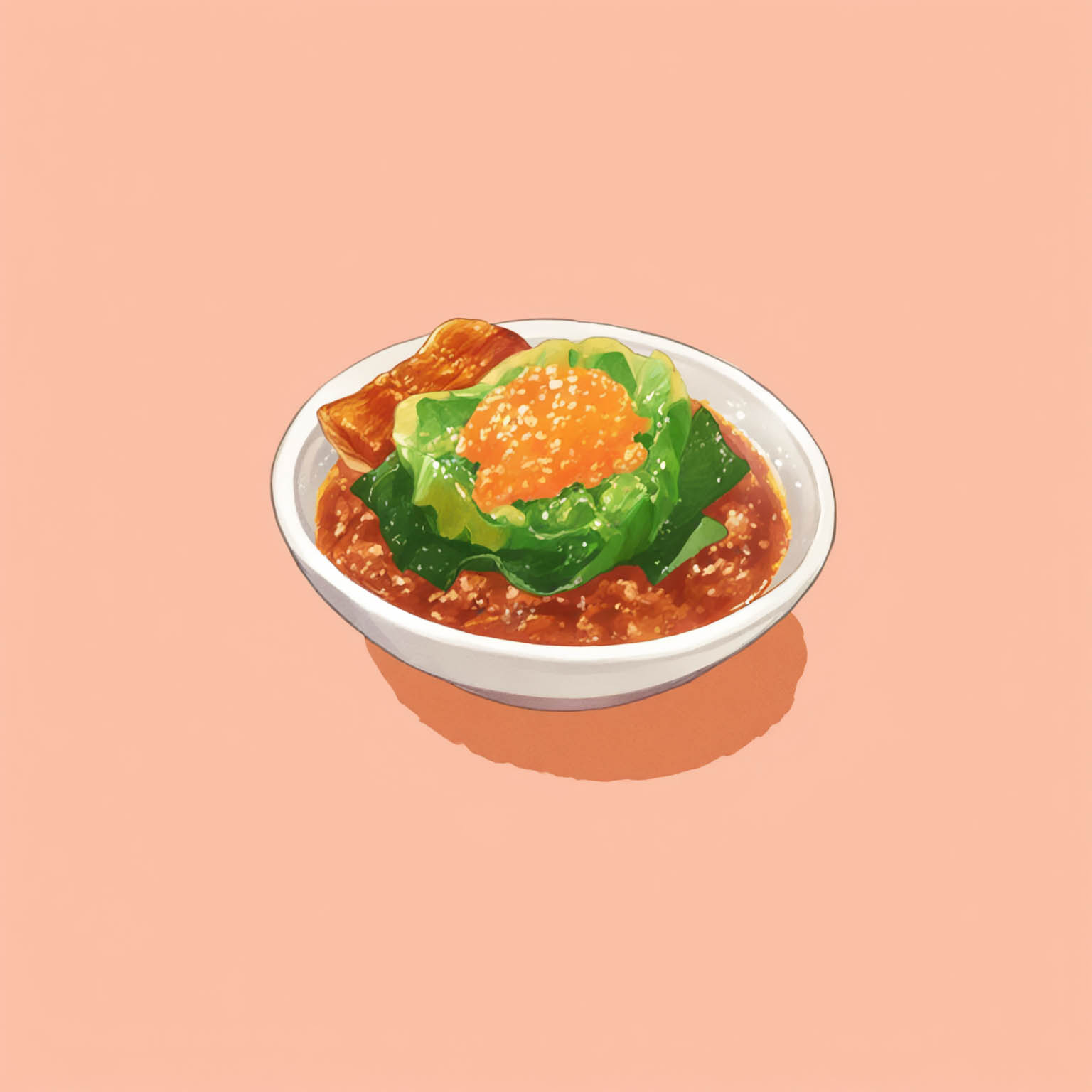}
    \end{minipage}
    
    
    \begin{minipage}{0.33\textwidth}
        \includegraphics[width=\textwidth,height=8cm,keepaspectratio]{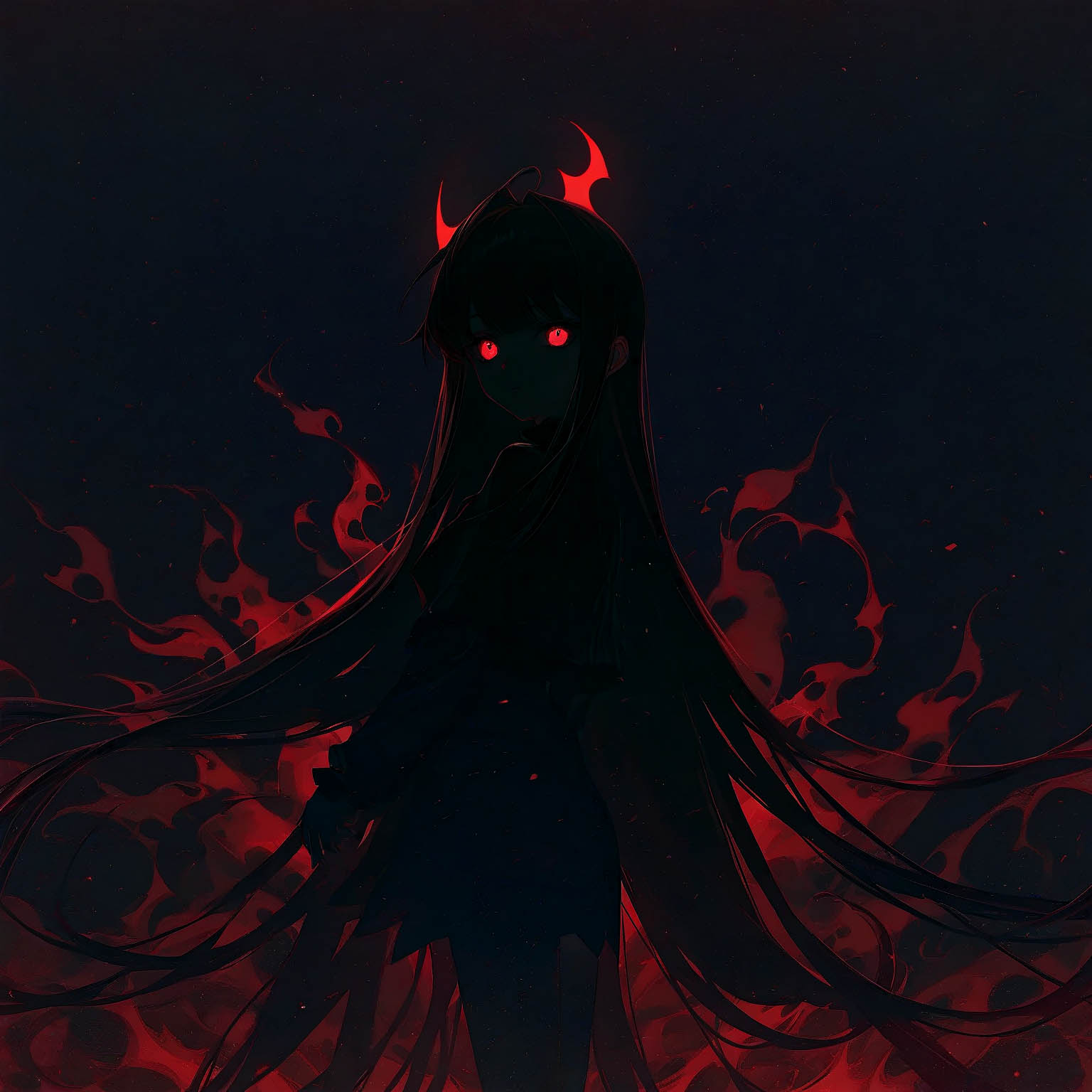}
    \end{minipage}\hfill
    \begin{minipage}{0.33\textwidth}
        \includegraphics[width=\textwidth,height=8cm,keepaspectratio]{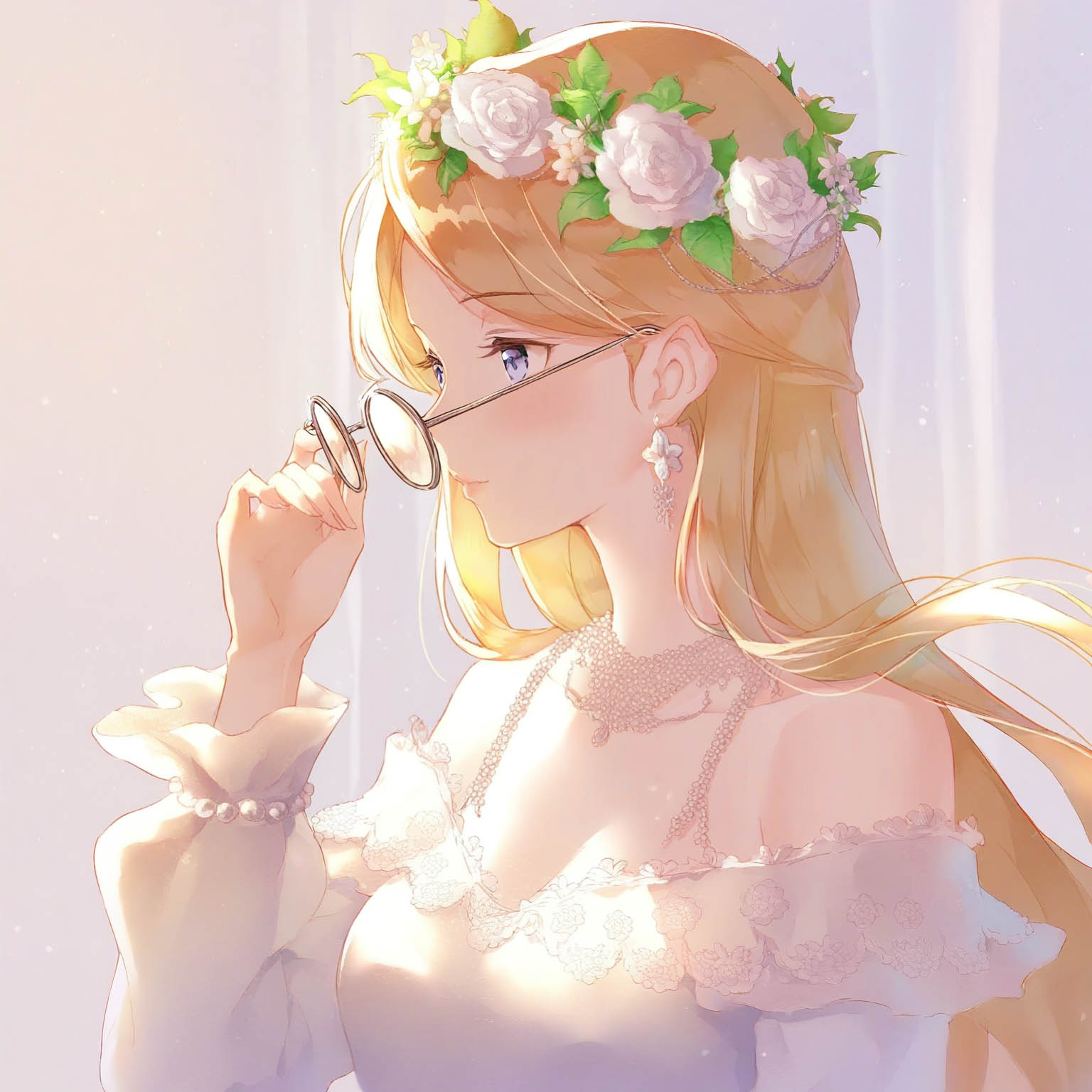}
    \end{minipage}\hfill
    \begin{minipage}{0.33\textwidth}
        \includegraphics[width=\textwidth,height=8cm,keepaspectratio]{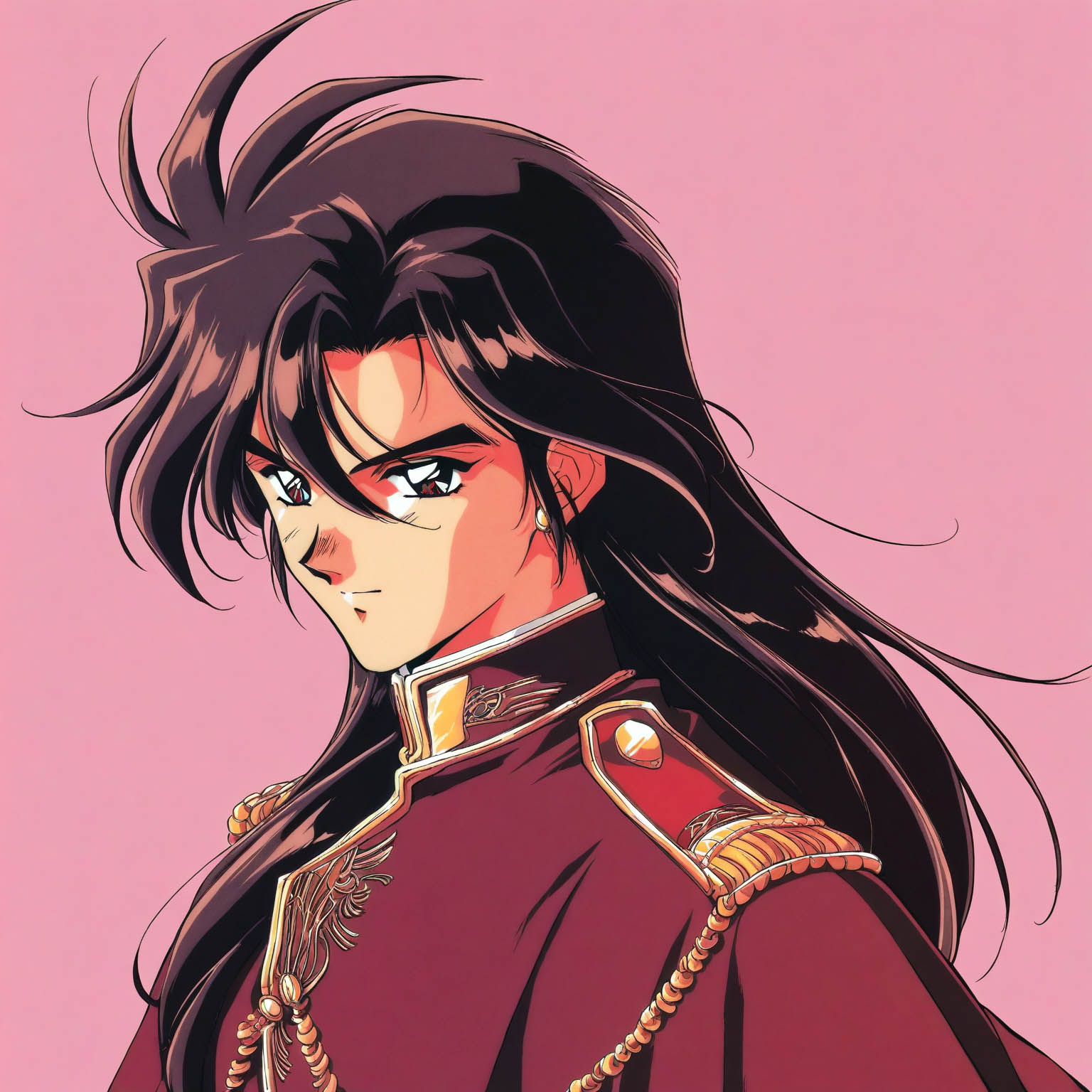}
    \end{minipage}


    \begin{minipage}{0.33\textwidth}
        \includegraphics[width=\textwidth,height=8cm,keepaspectratio]{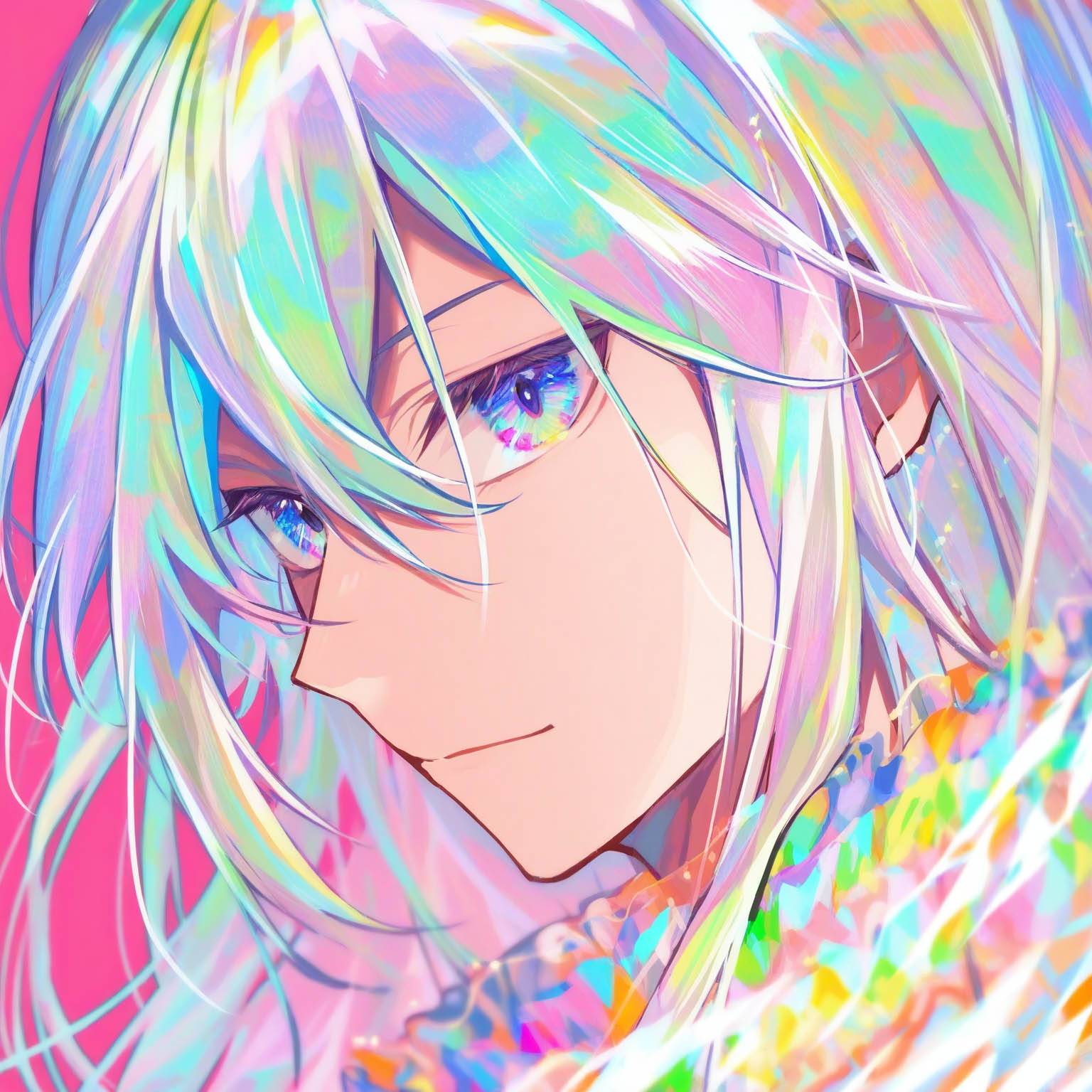}
    \end{minipage}\hfill
    \begin{minipage}{0.33\textwidth}
        \includegraphics[width=\textwidth,height=8cm,keepaspectratio]{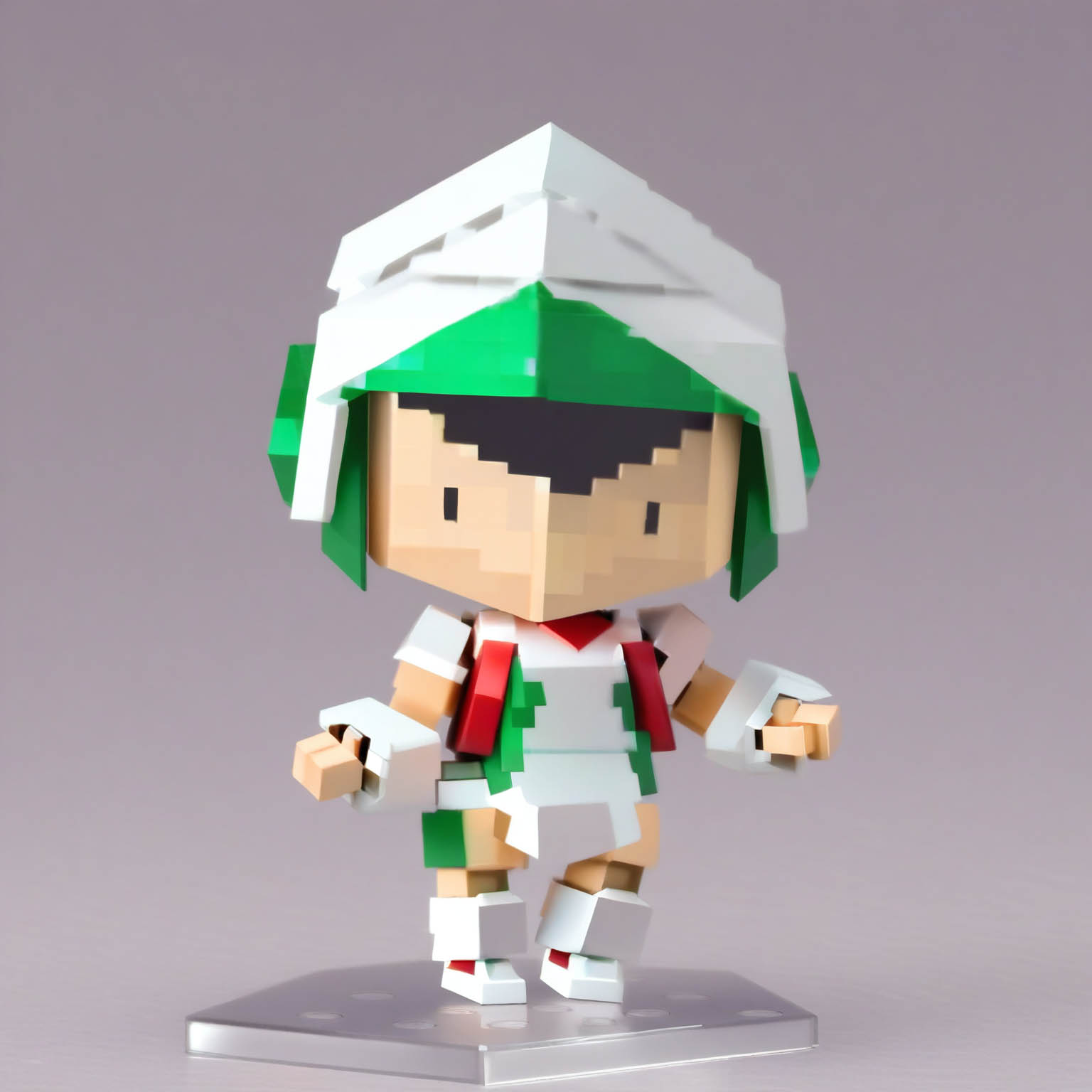}
    \end{minipage}\hfill
    \begin{minipage}{0.33\textwidth}
        \includegraphics[width=\textwidth,height=8cm,keepaspectratio]{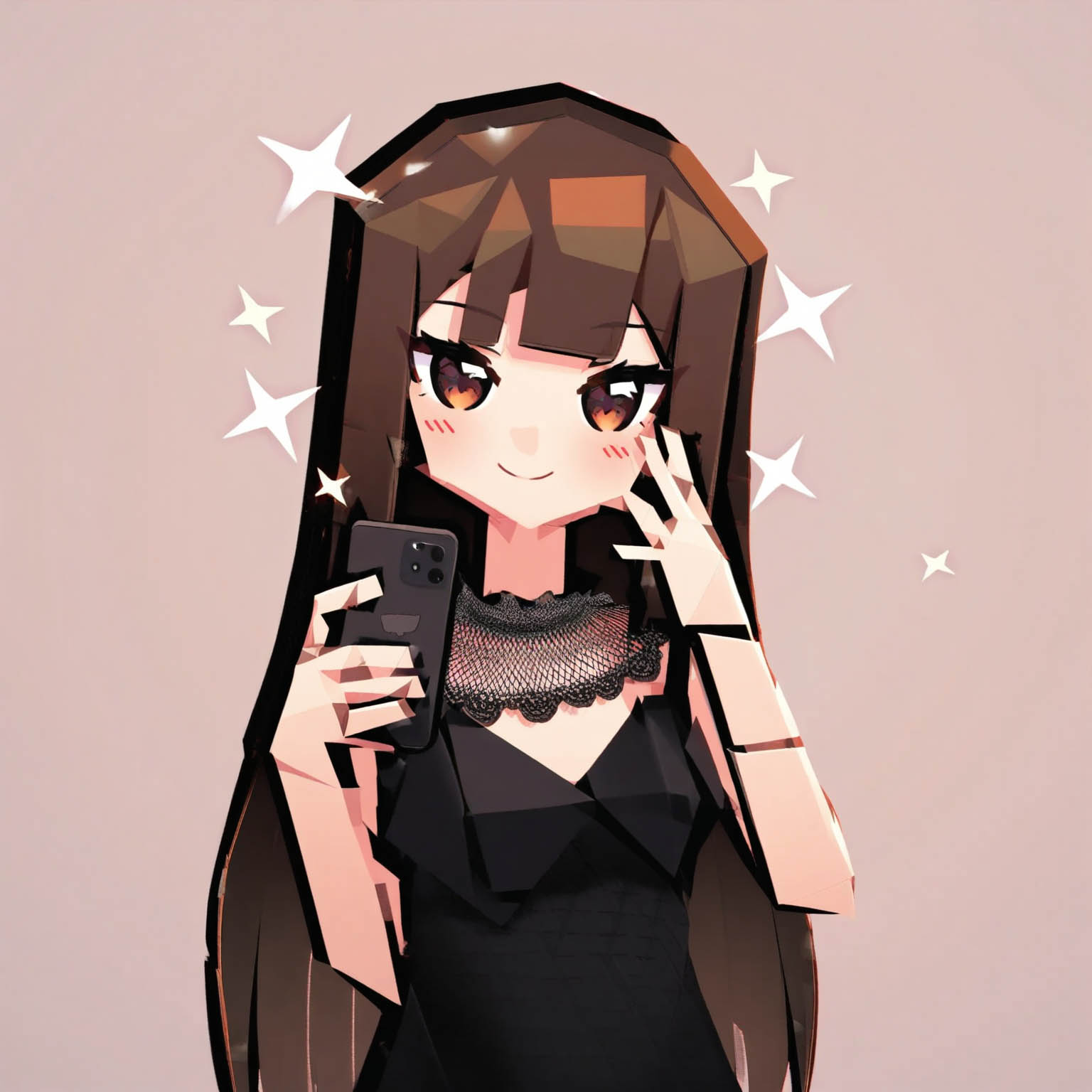}
    \end{minipage}
    \caption{\textbf{High-quality samples from Illustrious v1.1.} Illustrious v1.1 can generate various styles. These images are all 1536 $\times$ 1536 pixels.}
    \label{fig:Samples_v1.1}
\end{figure*}

Illustrious v1.1's sample image is depicted as Figure \ref{fig:Samples_v1.1}.

\newpage
\subsubsection{Illustrious v2.0}
\begin{figure*}[htb]
    \centering
    \begin{minipage}{0.33\textwidth}
    \includegraphics[width=\textwidth,height=8cm,keepaspectratio]{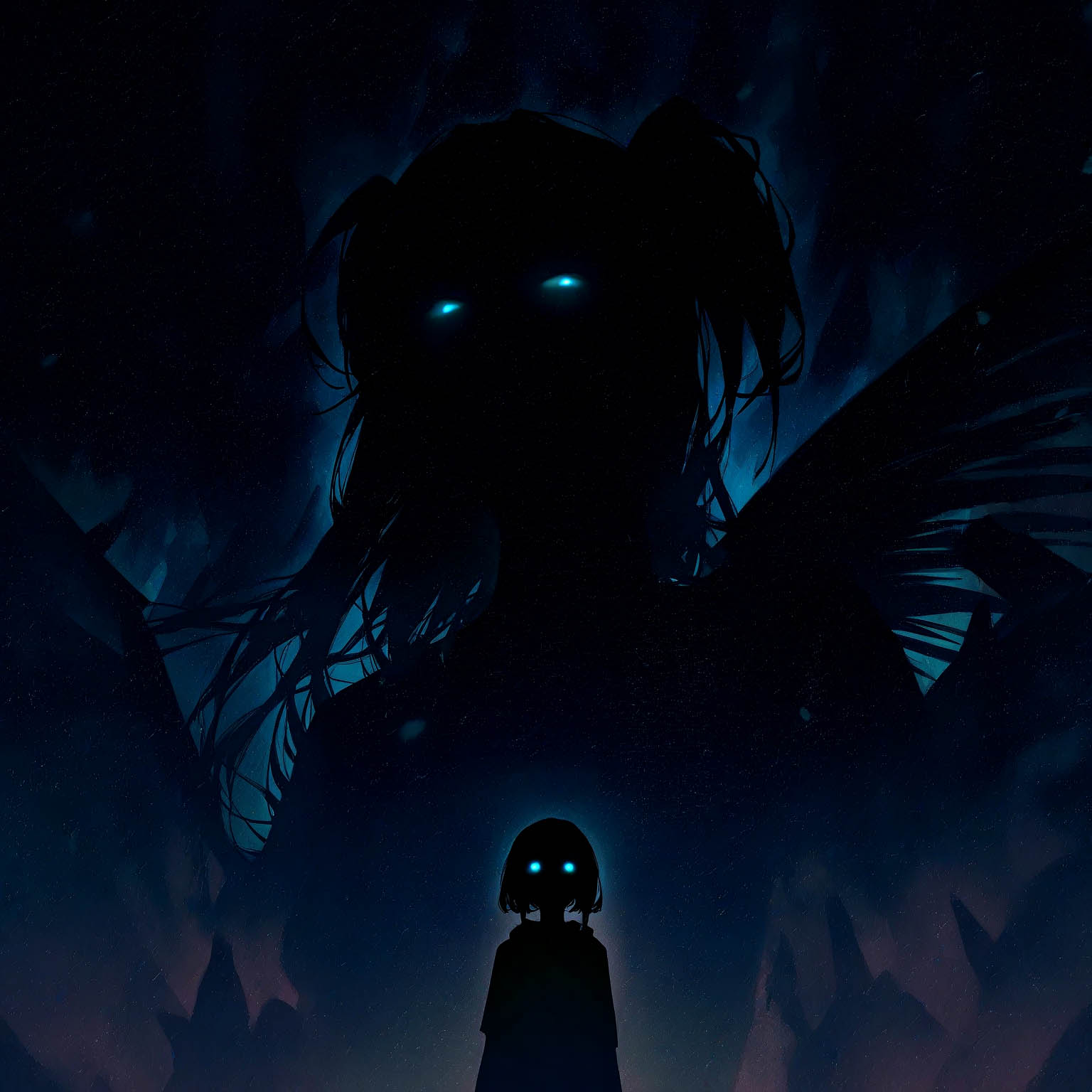}
    \end{minipage}\hfill
    \begin{minipage}{0.33\textwidth}
    \includegraphics[width=\textwidth,height=8cm,keepaspectratio]{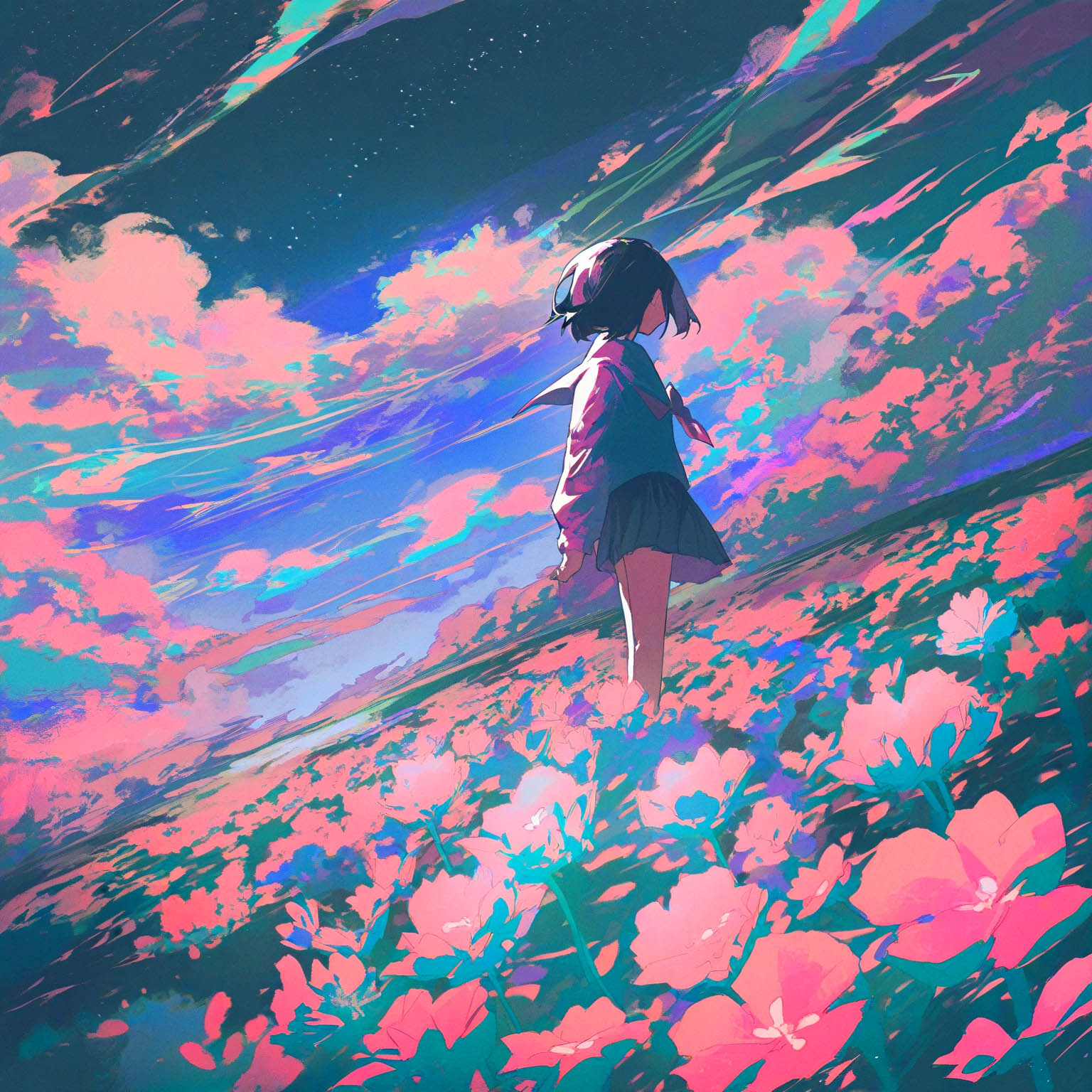}
    \end{minipage}\hfill
    \begin{minipage}{0.33\textwidth}
        \includegraphics[width=\textwidth,height=8cm,keepaspectratio]{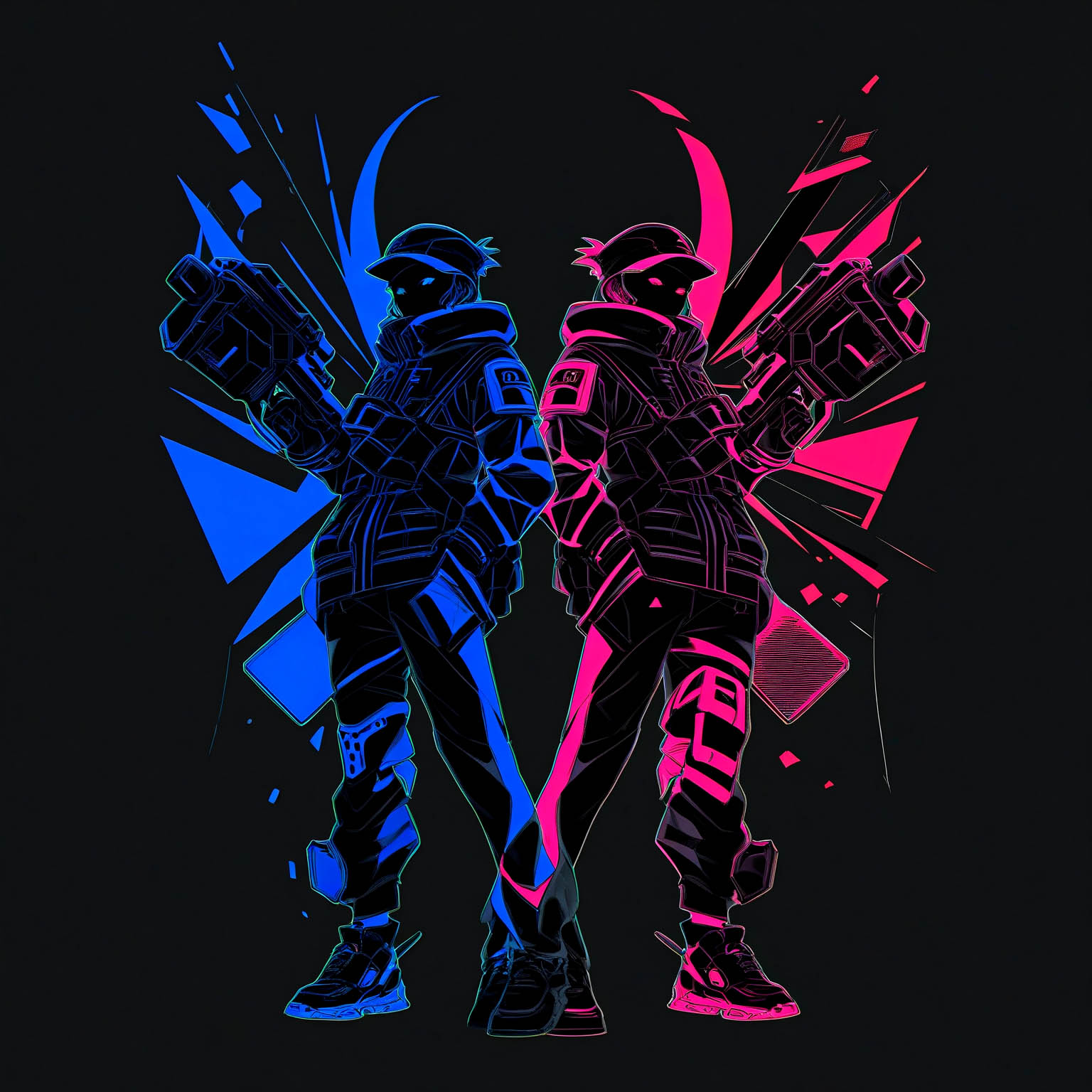}
    \end{minipage}
    
    
    \begin{minipage}{0.33\textwidth}
        \includegraphics[width=\textwidth,height=8cm,keepaspectratio]{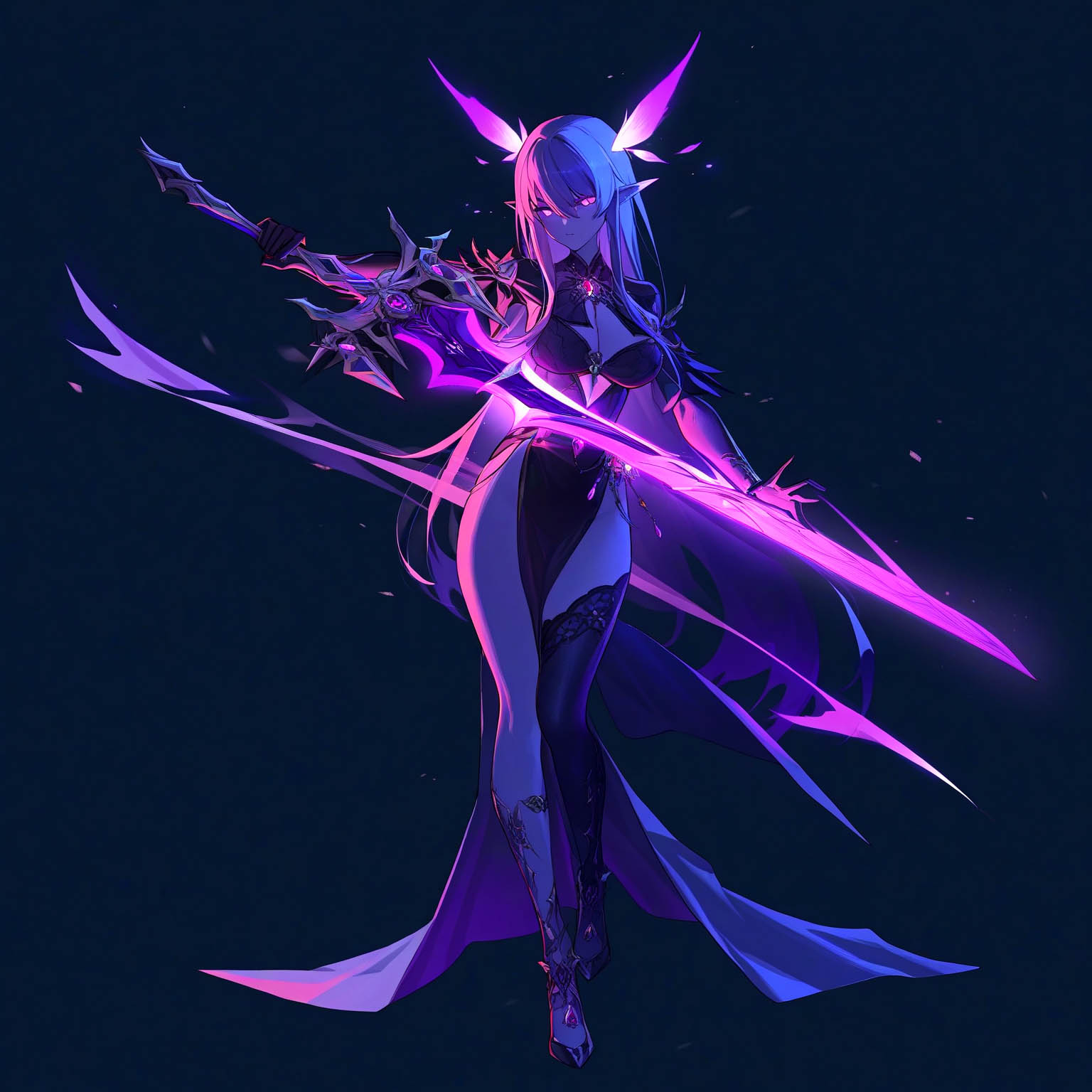}
    \end{minipage}\hfill
    \begin{minipage}{0.33\textwidth}
        \includegraphics[width=\textwidth,height=8cm,keepaspectratio]{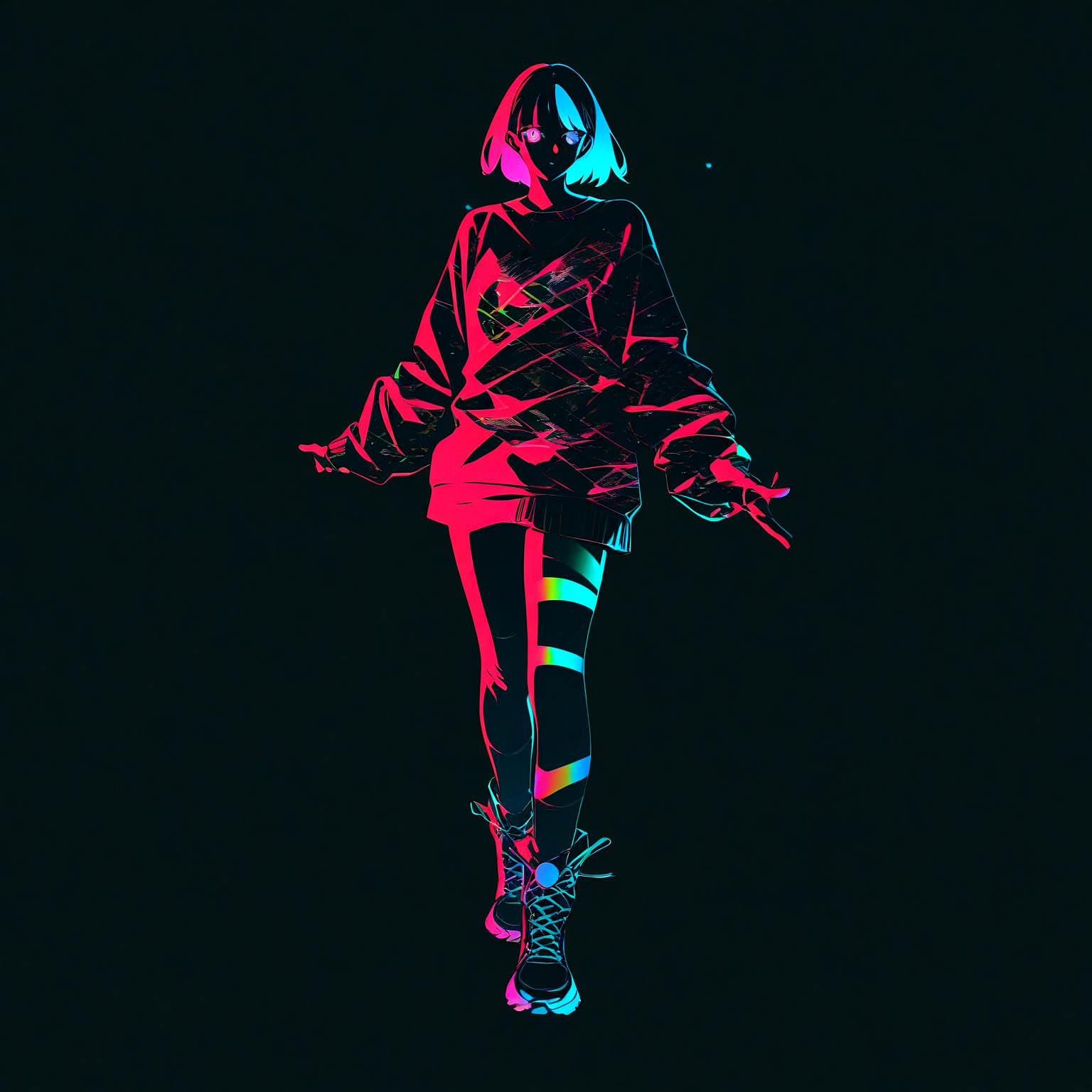}
    \end{minipage}\hfill
    \begin{minipage}{0.33\textwidth}
        \includegraphics[width=\textwidth,height=8cm,keepaspectratio]{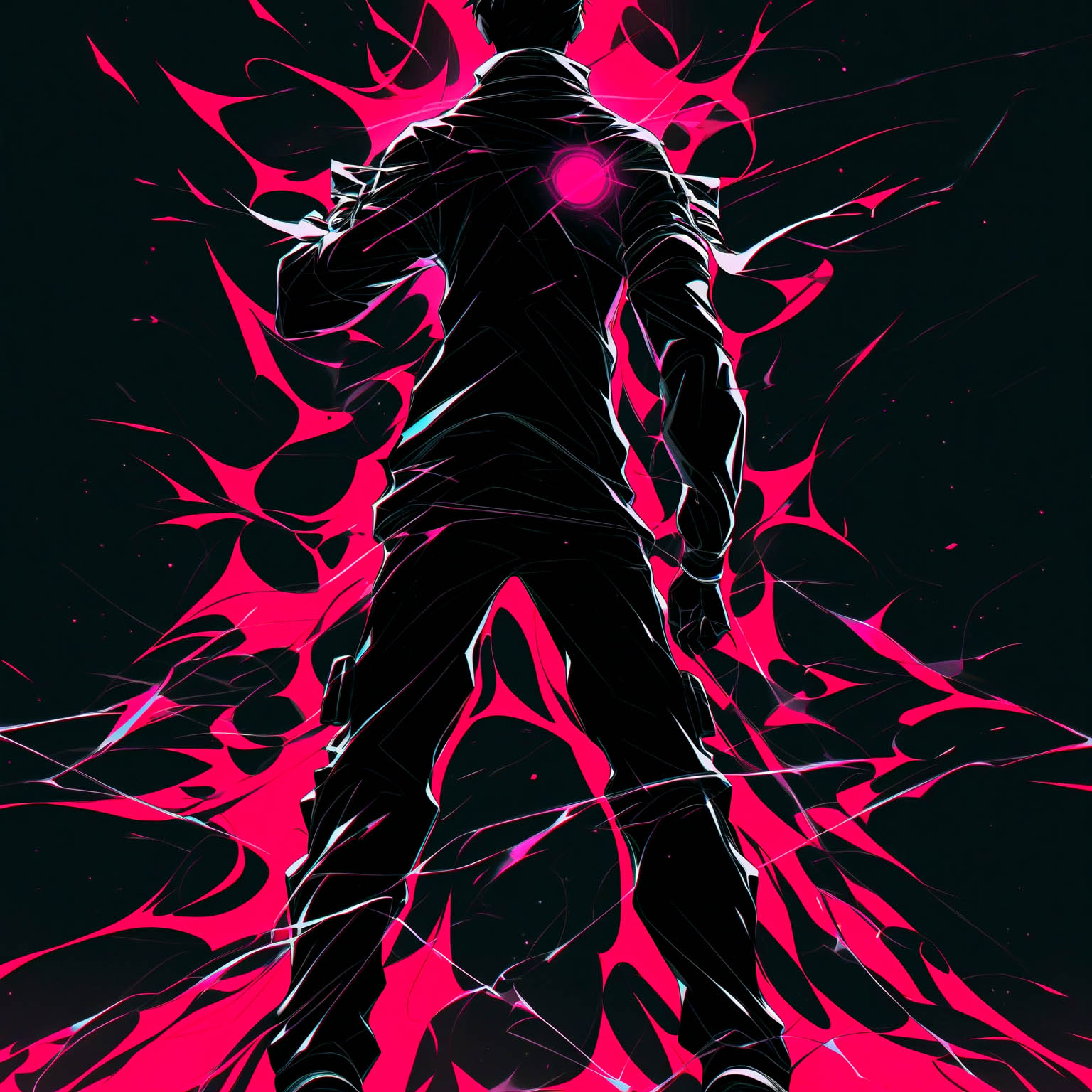}
    \end{minipage}

\caption{\textbf{High-quality samples from Illustrious v2.0.} Illustrious v2.0 can understand the natural language prompts.}
\label{fig:Samples_v2.0}
\end{figure*}

\begin{figure*}[htb]
    \centering
    \begin{minipage}{0.33\textwidth}
        \includegraphics[width=\textwidth,height=8cm,keepaspectratio]{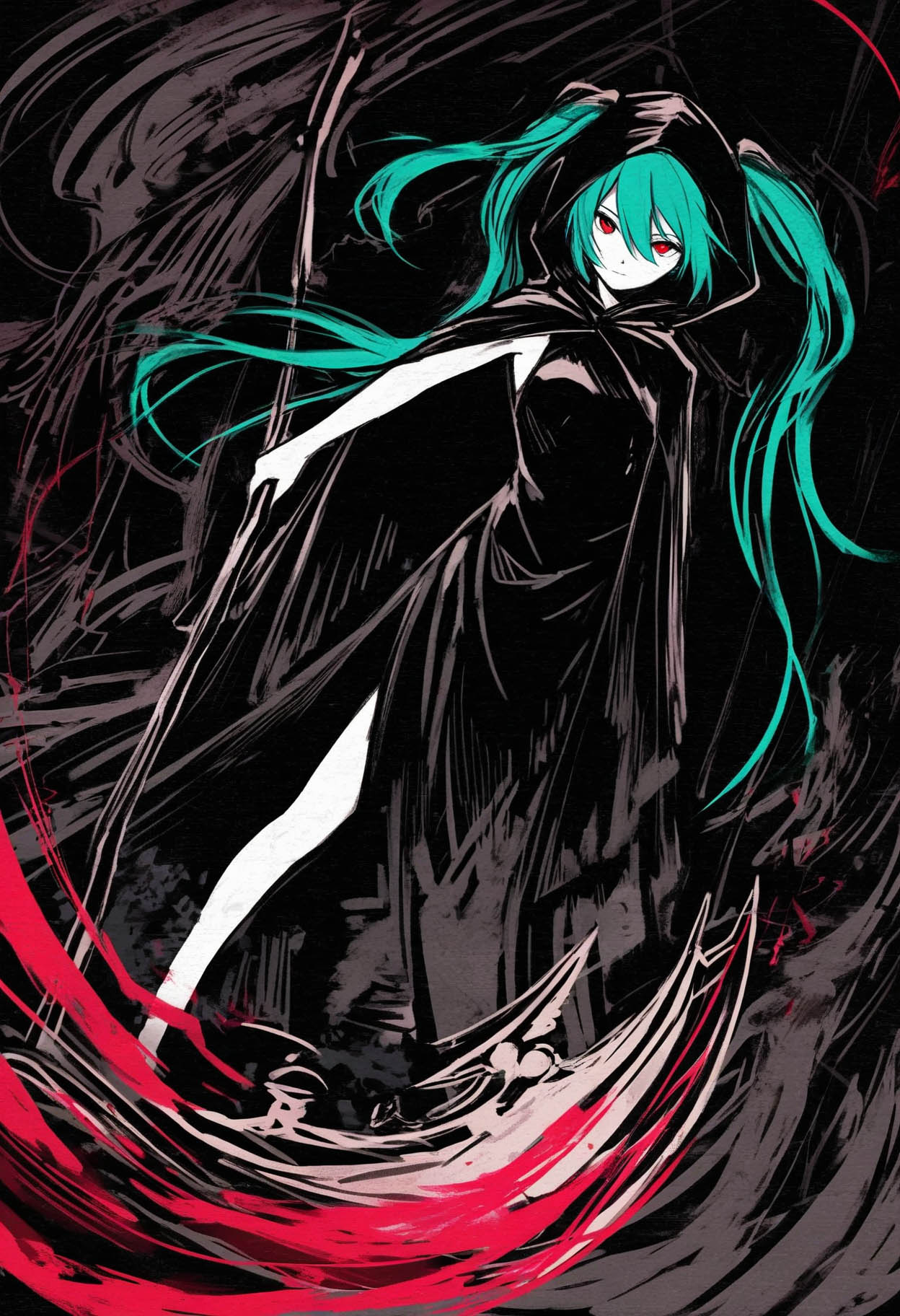}
    \end{minipage}\hfill
    \begin{minipage}{0.33\textwidth}
        \includegraphics[width=\textwidth,height=8cm,keepaspectratio]{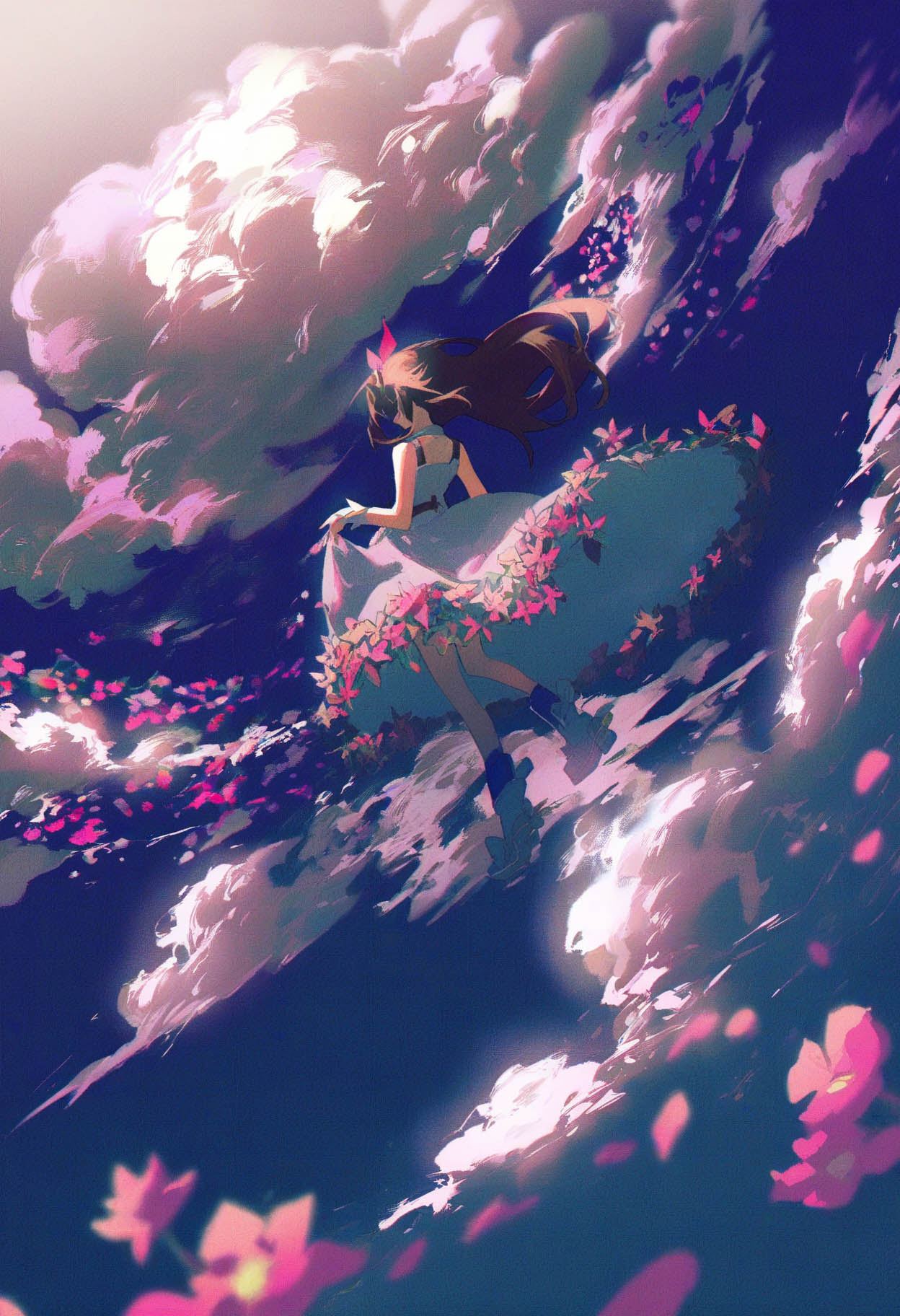}
    \end{minipage}\hfill
    \begin{minipage}{0.33\textwidth}
        \includegraphics[width=\textwidth,height=8cm,keepaspectratio]{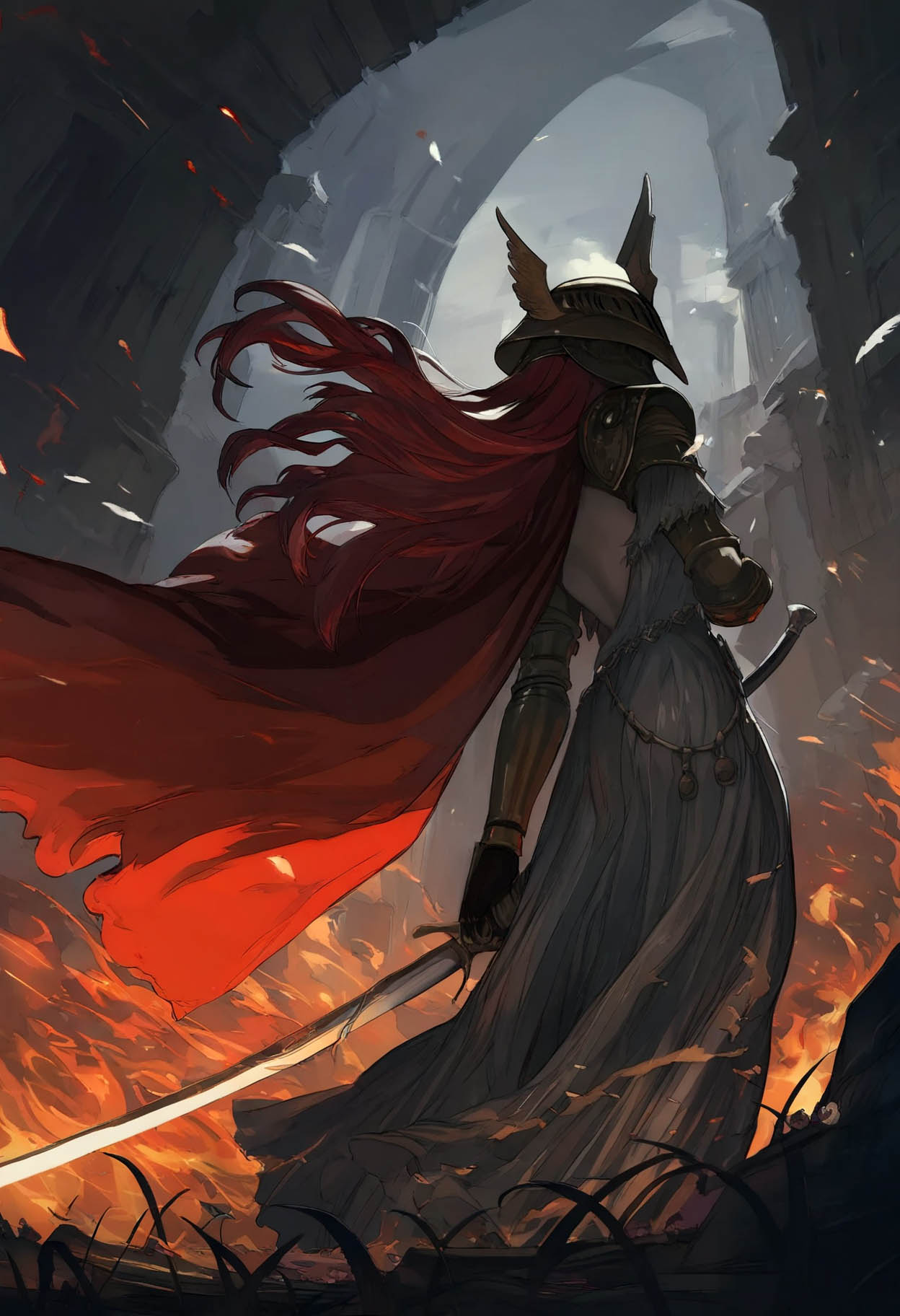}
    \end{minipage}
    
    
    \begin{minipage}{0.33\textwidth}
        \includegraphics[width=\textwidth,height=8cm,keepaspectratio]{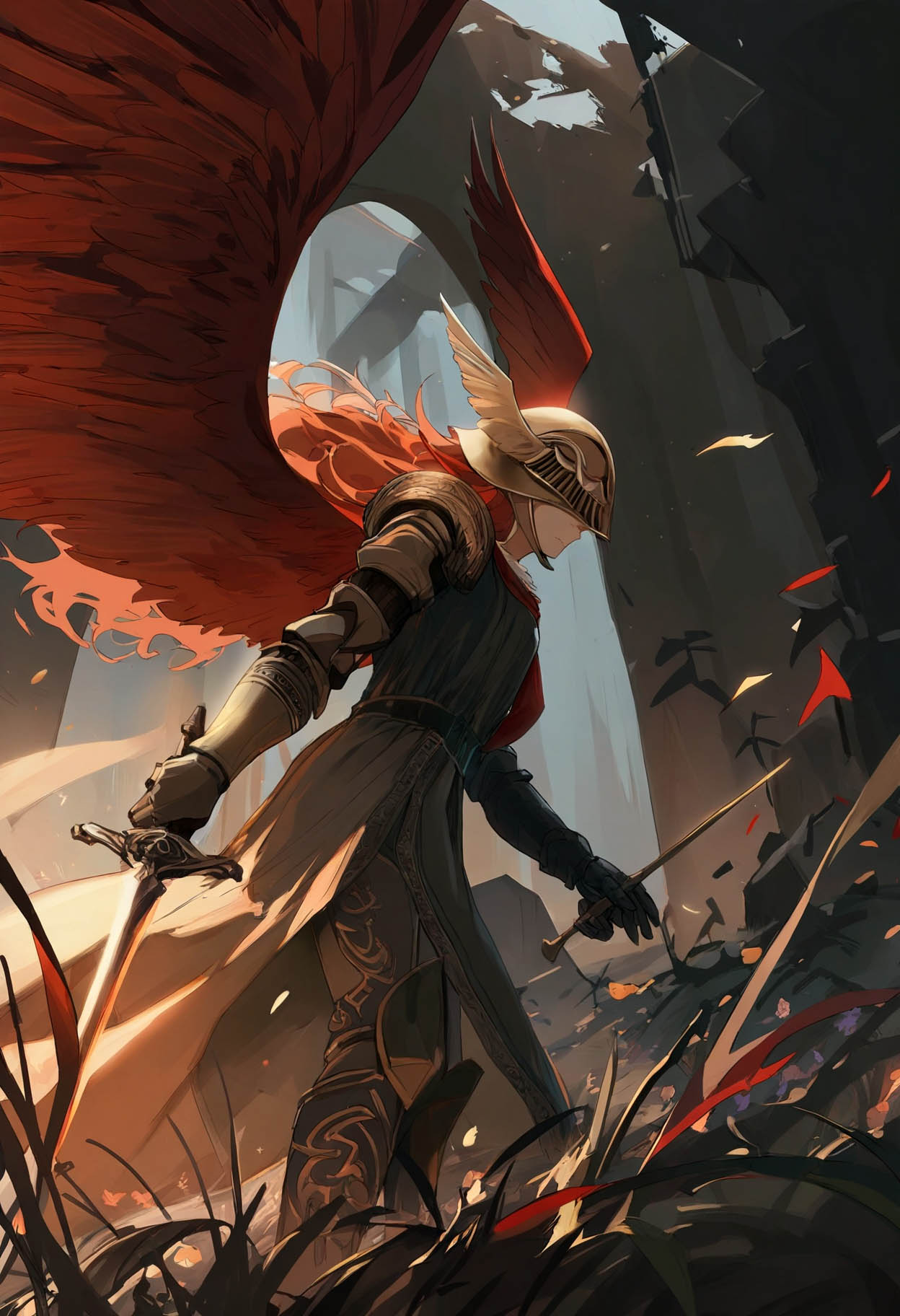}
    \end{minipage}\hfill
    \begin{minipage}{0.33\textwidth}
        \includegraphics[width=\textwidth,height=8cm,keepaspectratio]{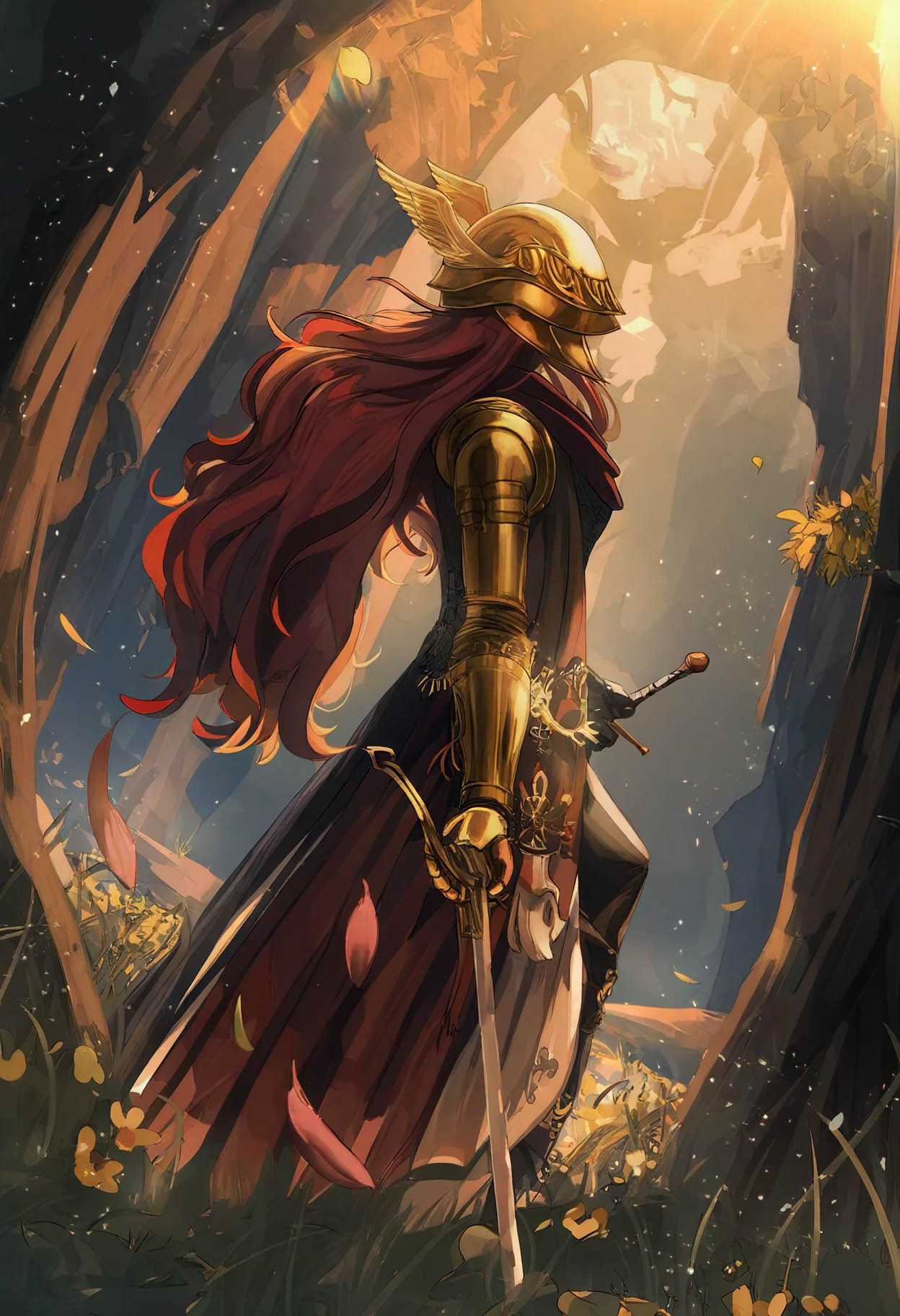}
    \end{minipage}\hfill
    \begin{minipage}{0.33\textwidth}
        \includegraphics[width=\textwidth,height=8cm,keepaspectratio]{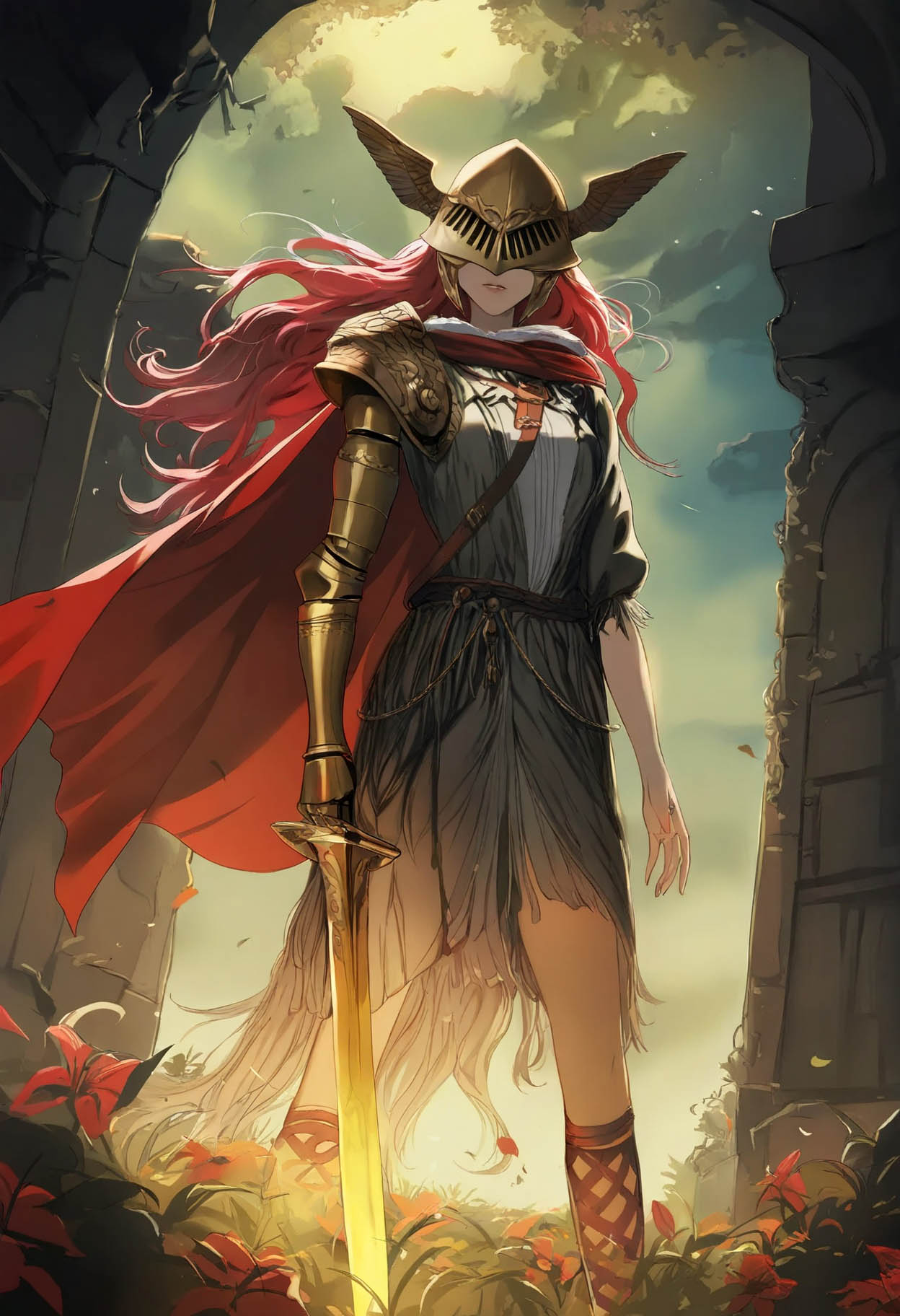}
    \end{minipage}


    \begin{minipage}{0.496\textwidth}
        \includegraphics[width=\textwidth,height=8cm,keepaspectratio]{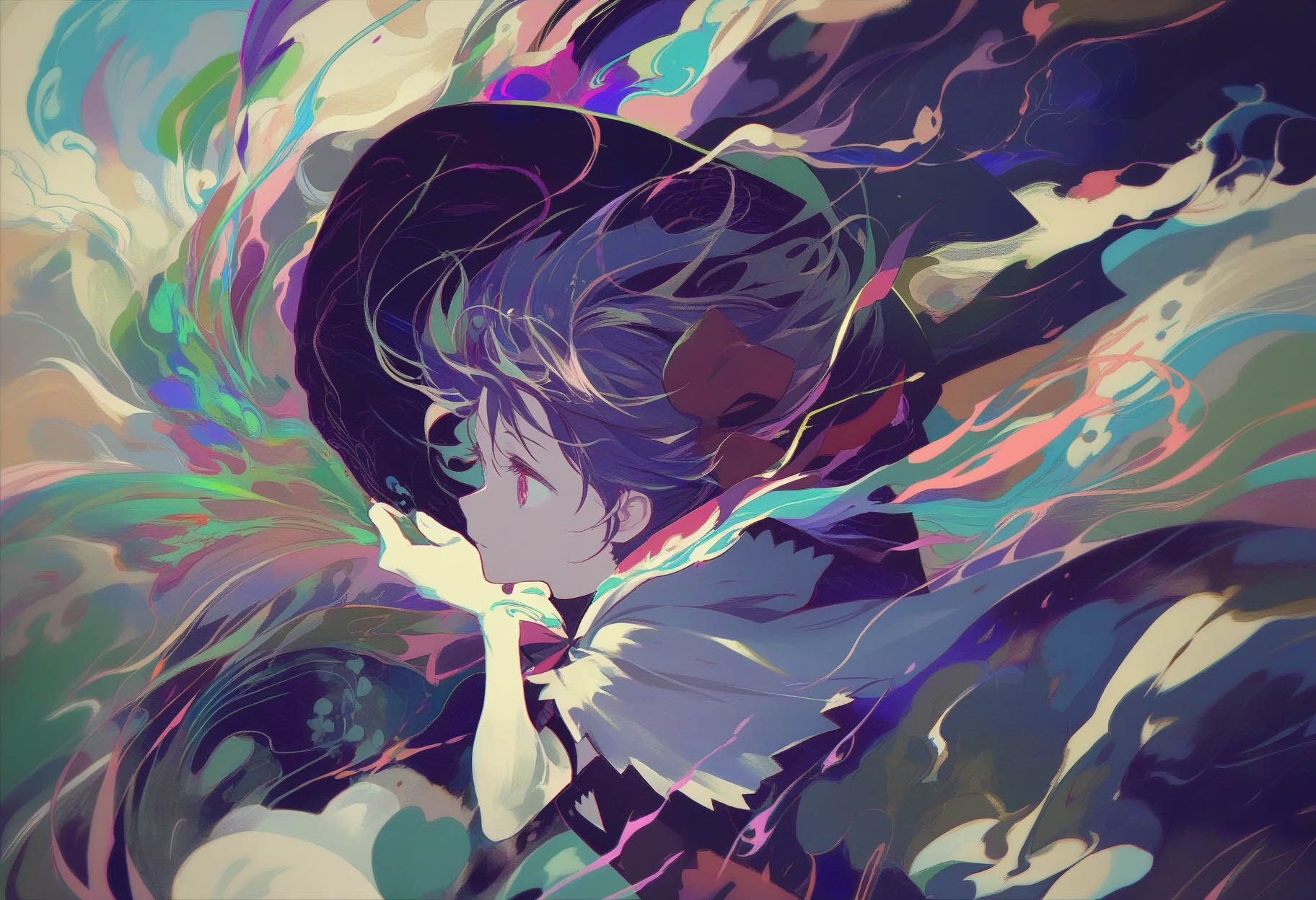}
    \end{minipage}\hfill
    \begin{minipage}{0.496\textwidth}
        \includegraphics[width=\textwidth,height=8cm,keepaspectratio]{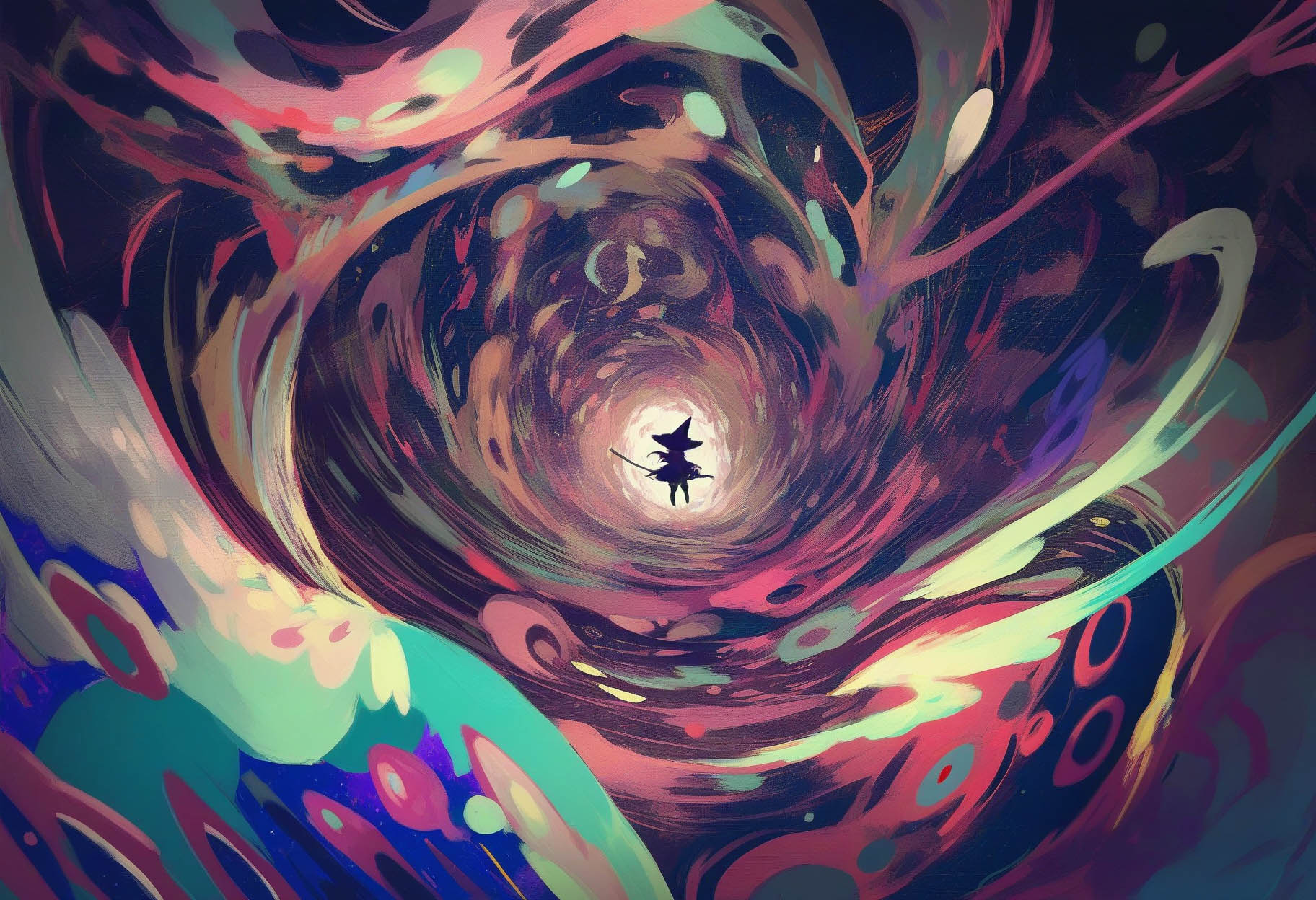}
    \end{minipage}\hfill
\caption{\textbf{Horizontal and Vertical High-quality samples from Illustrious v2.0.} Illustrious v2.0 can understand the natural language prompts.}
\label{fig:Samples_v2.0_HV}
\end{figure*}

Illustrious v2.0's sample image is depicted as Figure \ref{fig:Samples_v2.0} and \ref{fig:Samples_v2.0_HV}.

\begin{figure*}
    \centering
    \includegraphics[width=\linewidth]{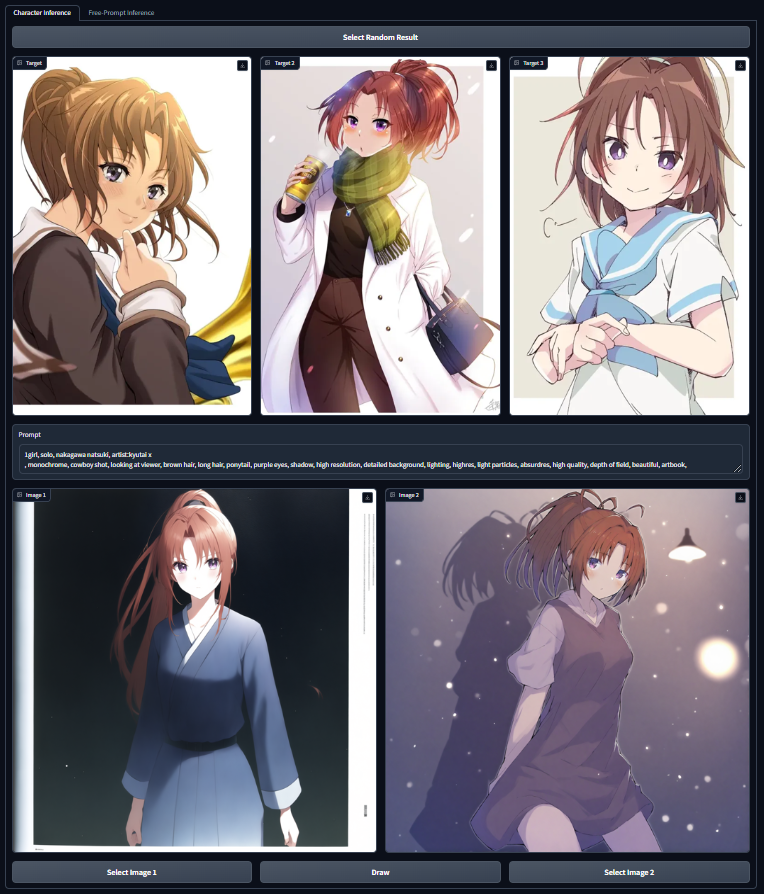}
    \caption{Model Compare Site Image}
    \label{fig:enter-label}
\end{figure*}

\begin{figure*}
    \centering
    \includegraphics[width=\textwidth]{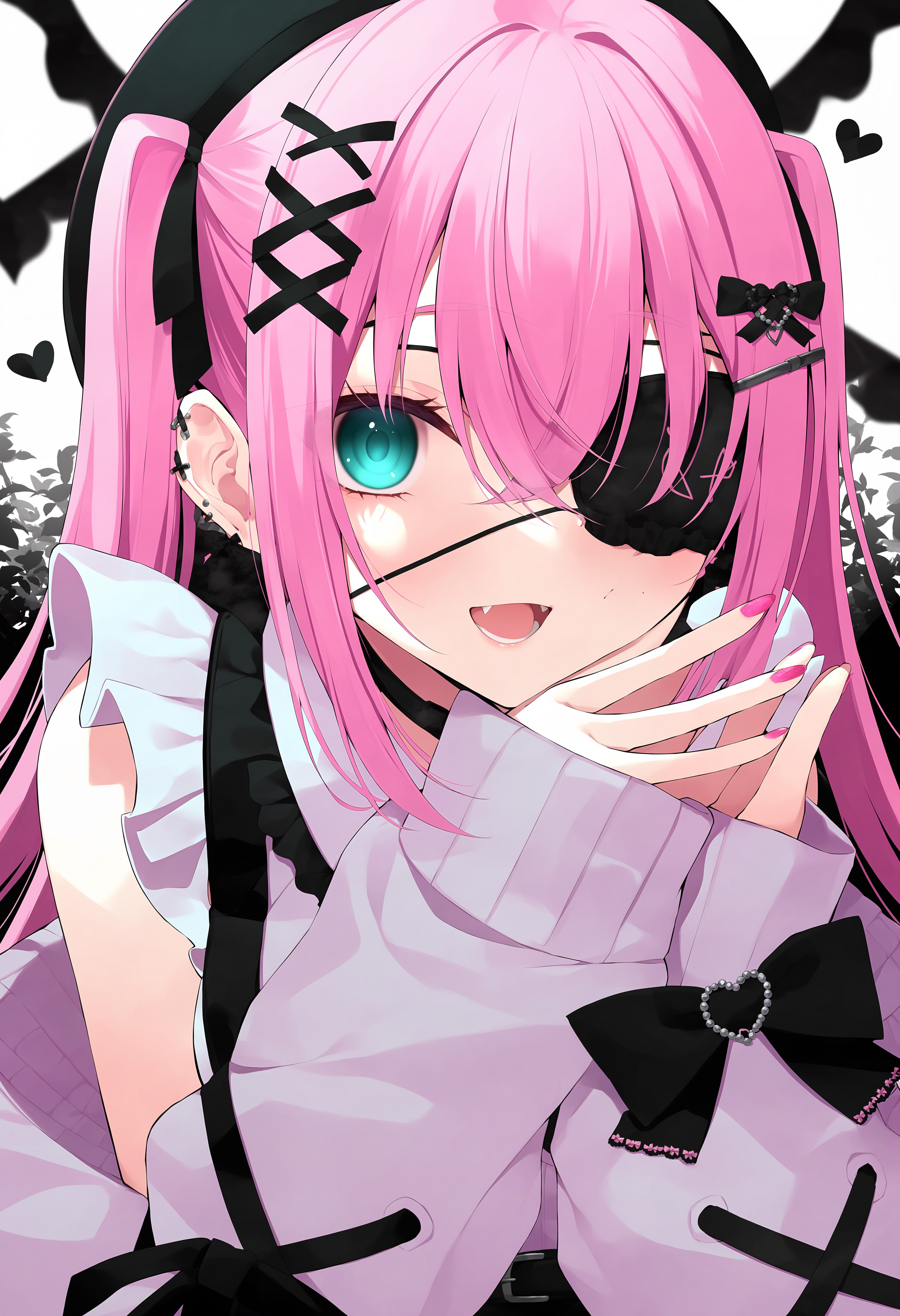}
    \caption{Illustrious v1.0+ can create the high resolution images. This image is 3744x5472 resolution by v1.0, firstly generated in 1248x1824, then upscaled toward 3744x5472 as same method using SDEdit \cite{meng2022sdeditguidedimagesynthesis}}.
    \label{fig:highresolution}
\end{figure*}

\newpage

\subsection {Recommended generation configuration}
We used Euler A Discrete sampler with step count >20, with CFG 5\textasciitilde7.5 for generation examples, however it may depend on styles, setups. For instance, we found that generating with DPM-based schedulers \cite{DPM} \cite{DPM++}, then piping through img2img pipeline with Euler discrete, works well for aesthetic / detailed image setups.
Illustrious v0.1 supports 1MP resolutions.
Illustrious v1.0+ supports native 1MP\textasciitilde2.25MP resolutions, up to 4MP with some loss.
All images, which exceeds 1:10 ratio, was not targetted and included in training.

\section{Thanks To}
\textbf{Kohaku (KBlueLeaf)}, with massive supports and initiatives to train the large scale base models,

\textbf{WDV team and community}, with initial thoughts and benchmarks over various prompts,

\textbf{DeepGHS team}, with open-minded datasets and tools, massive contributions fostering open source research,

and \textbf{OnomaAI}, supporting the research and training, allowing the model to exist.



\newpage


{
\small
\begin{thebibliography}{81}
    \providecommand{\natexlab}[1]{#1}
    \providecommand{\url}[1]{\texttt{#1}}
    \expandafter\ifx\csname urlstyle\endcsname\relax
      \providecommand{\doi}[1]{doi: #1}\else
      \providecommand{\doi}{doi: \begingroup \urlstyle{rm}\Url}\fi
    
    \bibitem[Alphanome.AI(2023)]{waluigi_effect_2023}
    Alphanome.AI.
    \newblock The waluigi effect in ai, 2023.
    \newblock URL \url{https://www.alphanome.ai/post/the-waluigi-effect-in-ai}.
    \newblock Accessed: 2024-09-28.
    
    \bibitem[Anonymous and community(2023)]{danbooru2023}
    Anonymous and Danbooru community.
    \newblock Danbooru2023: A large-scale crowdsourced \& tagged anime illustration dataset.
    \newblock \url{https://huggingface.co/datasets/nyanko7/danbooru2023}, 2023.
    \newblock URL \url{https://huggingface.co/datasets/nyanko7/danbooru2023}.
    
    \bibitem[Anonymous et~al.(2022)Anonymous, community, and Branwen]{danbooru}
    Anonymous, Danbooru community, and Gwern Branwen.
    \newblock Danbooru2021: A large-scale crowdsourced \& tagged anime illustration dataset.
    \newblock \url{https://gwern.net/danbooru2021}, January 2022.
    \newblock URL \url{https://gwern.net/danbooru2021}.
    \newblock Accessed: DATE.
    
    \bibitem[Azizi et~al.(2023)Azizi, Kornblith, Saharia, Norouzi, and Fleet]{syntheticdatadiffusionmodels}
    Shekoofeh Azizi, Simon Kornblith, Chitwan Saharia, Mohammad Norouzi, and David~J. Fleet.
    \newblock Synthetic data from diffusion models improves imagenet classification, 2023.
    \newblock URL \url{https://arxiv.org/abs/2304.08466}.
    
    \bibitem[Balaji et~al.(2023)Balaji, Nah, Huang, Vahdat, Song, Zhang, Kreis, Aittala, Aila, Laine, Catanzaro, Karras, and Liu]{ediff-i}
    Yogesh Balaji, Seungjun Nah, Xun Huang, Arash Vahdat, Jiaming Song, Qinsheng Zhang, Karsten Kreis, Miika Aittala, Timo Aila, Samuli Laine, Bryan Catanzaro, Tero Karras, and Ming-Yu Liu.
    \newblock ediff-i: Text-to-image diffusion models with an ensemble of expert denoisers, 2023.
    \newblock URL \url{https://arxiv.org/abs/2211.01324}.
    
    \bibitem[bdsqlsz(2024)]{safetycontrol}
    bdsqlsz.
    \newblock Adapter-based approach to control content safety.
    \newblock \url{https://huggingface.co/bdsqlsz/filter_nude}, 2024.
    
    \bibitem[black-forest labs(2024)]{blackforestlabs2024flux}
    black-forest labs.
    \newblock flux, 2024.
    \newblock Available: \url{https://github.com/black-forest-labs/flux}.
    
    \bibitem[Changpinyo et~al.(2021)Changpinyo, Sharma, Ding, and Soricut]{CC12M}
    Soravit Changpinyo, Piyush Sharma, Nan Ding, and Radu Soricut.
    \newblock Conceptual 12m: Pushing web-scale image-text pre-training to recognize long-tail visual concepts, 2021.
    \newblock URL \url{https://arxiv.org/abs/2102.08981}.
    
    \bibitem[Chen et~al.(2023)Chen, Yu, Ge, Yao, Xie, Wu, Wang, Kwok, Luo, Lu, and Li]{Pixart-alpha}
    Junsong Chen, Jincheng Yu, Chongjian Ge, Lewei Yao, Enze Xie, Yue Wu, Zhongdao Wang, James Kwok, Ping Luo, Huchuan Lu, and Zhenguo Li.
    \newblock Pixart-$\alpha$: Fast training of diffusion transformer for photorealistic text-to-image synthesis, 2023.
    \newblock URL \url{https://arxiv.org/abs/2310.00426}.
    
    \bibitem[Chiang et~al.(2024)Chiang, Zheng, Sheng, Angelopoulos, Li, Li, Zhang, Zhu, Jordan, Gonzalez, and Stoica]{chatbotarena}
    Wei-Lin Chiang, Lianmin Zheng, Ying Sheng, Anastasios~Nikolas Angelopoulos, Tianle Li, Dacheng Li, Hao Zhang, Banghua Zhu, Michael Jordan, Joseph~E. Gonzalez, and Ion Stoica.
    \newblock Chatbot arena: An open platform for evaluating llms by human preference, 2024.
    \newblock URL \url{https://arxiv.org/abs/2403.04132}.
    
    \bibitem[Darcet et~al.(2024)Darcet, Oquab, Mairal, and Bojanowski]{RegisterToken}
    Timothée Darcet, Maxime Oquab, Julien Mairal, and Piotr Bojanowski.
    \newblock Vision transformers need registers, 2024.
    \newblock URL \url{https://arxiv.org/abs/2309.16588}.
    
    \bibitem[Dehghani et~al.(2023)Dehghani, Djolonga, Mustafa, Padlewski, Heek, Gilmer, Steiner, Caron, Geirhos, Alabdulmohsin, Jenatton, Beyer, Tschannen, Arnab, Wang, Riquelme, Minderer, Puigcerver, Evci, Kumar, van Steenkiste, Elsayed, Mahendran, Yu, Oliver, Huot, Bastings, Collier, Gritsenko, Birodkar, Vasconcelos, Tay, Mensink, Kolesnikov, Pavetić, Tran, Kipf, Lučić, Zhai, Keysers, Harmsen, and Houlsby]{ScalingViT}
    Mostafa Dehghani, Josip Djolonga, Basil Mustafa, Piotr Padlewski, Jonathan Heek, Justin Gilmer, Andreas Steiner, Mathilde Caron, Robert Geirhos, Ibrahim Alabdulmohsin, Rodolphe Jenatton, Lucas Beyer, Michael Tschannen, Anurag Arnab, Xiao Wang, Carlos Riquelme, Matthias Minderer, Joan Puigcerver, Utku Evci, Manoj Kumar, Sjoerd van Steenkiste, Gamaleldin~F. Elsayed, Aravindh Mahendran, Fisher Yu, Avital Oliver, Fantine Huot, Jasmijn Bastings, Mark~Patrick Collier, Alexey Gritsenko, Vighnesh Birodkar, Cristina Vasconcelos, Yi~Tay, Thomas Mensink, Alexander Kolesnikov, Filip Pavetić, Dustin Tran, Thomas Kipf, Mario Lučić, Xiaohua Zhai, Daniel Keysers, Jeremiah Harmsen, and Neil Houlsby.
    \newblock Scaling vision transformers to 22 billion parameters, 2023.
    \newblock URL \url{https://arxiv.org/abs/2302.05442}.
    
    \bibitem[Deng et~al.(2009)Deng, Dong, Socher, Li, Li, and Fei-Fei]{ImageNet}
    Jia Deng, Wei Dong, Richard Socher, Li-Jia Li, Kai Li, and Li~Fei-Fei.
    \newblock Imagenet: A large-scale hierarchical image database.
    \newblock In \emph{2009 IEEE Conference on Computer Vision and Pattern Recognition}, pages 248--255, 2009.
    \newblock \doi{10.1109/CVPR.2009.5206848}.
    
    \bibitem[Devarakonda et~al.(2018)Devarakonda, Naumov, and Garland]{AdaBatch}
    Aditya Devarakonda, Maxim Naumov, and Michael Garland.
    \newblock Adabatch: Adaptive batch sizes for training deep neural networks, 2018.
    \newblock URL \url{https://arxiv.org/abs/1712.02029}.
    
    \bibitem[Dhariwal and Nichol(2021)]{diffusionbeatsgan}
    Prafulla Dhariwal and Alex Nichol.
    \newblock Diffusion models beat gans on image synthesis, 2021.
    \newblock URL \url{https://arxiv.org/abs/2105.05233}.
    
    \bibitem[Dong and narugo1992(2024)]{CCIP}
    Ziyi Dong and narugo1992.
    \newblock Contrastive anime character image pre-training.
    \newblock \url{https://huggingface.co/deepghs/ccip}, 2024.
    
    \bibitem[Du et~al.(2022)Du, Qian, Liu, Ding, Qiu, Yang, and Tang]{GLM}
    Zhengxiao Du, Yujie Qian, Xiao Liu, Ming Ding, Jiezhong Qiu, Zhilin Yang, and Jie Tang.
    \newblock Glm: General language model pretraining with autoregressive blank infilling, 2022.
    \newblock URL \url{https://arxiv.org/abs/2103.10360}.
    
    \bibitem[Esser et~al.(2024)Esser, Kulal, Blattmann, Entezari, Müller, Saini, Levi, Lorenz, Sauer, Boesel, Podell, Dockhorn, English, Lacey, Goodwin, Marek, and Rombach]{SD3}
    Patrick Esser, Sumith Kulal, Andreas Blattmann, Rahim Entezari, Jonas Müller, Harry Saini, Yam Levi, Dominik Lorenz, Axel Sauer, Frederic Boesel, Dustin Podell, Tim Dockhorn, Zion English, Kyle Lacey, Alex Goodwin, Yannik Marek, and Robin Rombach.
    \newblock Scaling rectified flow transformers for high-resolution image synthesis, 2024.
    \newblock URL \url{https://arxiv.org/abs/2403.03206}.
    
    \bibitem[fal(2024)]{fal2024auraflow}
    fal.
    \newblock Auraflow, 2024.
    \newblock Available: \url{https://huggingface.co/fal/AuraFlow?ref=blog.fal.ai}.
    
    \bibitem[{FreeDev Project}(2024)]{fpai2024}
    {FreeDev Project}.
    \newblock Fair public ai license 1.0-sd, 2024.
    \newblock Retrieved from \url{https://freedevproject.org/faipl-1.0-sd/}.
    
    \bibitem[Gadre et~al.(2023)Gadre, Ilharco, Fang, Hayase, Smyrnis, Nguyen, Marten, Wortsman, Ghosh, Zhang, Orgad, Entezari, Daras, Pratt, Ramanujan, Bitton, Marathe, Mussmann, Vencu, Cherti, Krishna, Koh, Saukh, Ratner, Song, Hajishirzi, Farhadi, Beaumont, Oh, Dimakis, Jitsev, Carmon, Shankar, and Schmidt]{DataComp}
    Samir~Yitzhak Gadre, Gabriel Ilharco, Alex Fang, Jonathan Hayase, Georgios Smyrnis, Thao Nguyen, Ryan Marten, Mitchell Wortsman, Dhruba Ghosh, Jieyu Zhang, Eyal Orgad, Rahim Entezari, Giannis Daras, Sarah Pratt, Vivek Ramanujan, Yonatan Bitton, Kalyani Marathe, Stephen Mussmann, Richard Vencu, Mehdi Cherti, Ranjay Krishna, Pang~Wei Koh, Olga Saukh, Alexander Ratner, Shuran Song, Hannaneh Hajishirzi, Ali Farhadi, Romain Beaumont, Sewoong Oh, Alex Dimakis, Jenia Jitsev, Yair Carmon, Vaishaal Shankar, and Ludwig Schmidt.
    \newblock Datacomp: In search of the next generation of multimodal datasets, 2023.
    \newblock URL \url{https://arxiv.org/abs/2304.14108}.
    
    \bibitem[Gal et~al.(2022)Gal, Alaluf, Atzmon, Patashnik, Bermano, Chechik, and Cohen-Or]{TextualInversion}
    Rinon Gal, Yuval Alaluf, Yuval Atzmon, Or~Patashnik, Amit~H. Bermano, Gal Chechik, and Daniel Cohen-Or.
    \newblock An image is worth one word: Personalizing text-to-image generation using textual inversion, 2022.
    \newblock URL \url{https://arxiv.org/abs/2208.01618}.
    
    \bibitem[Gandikota et~al.(2023)Gandikota, Materzynska, Zhou, Torralba, and Bau]{gandikota2023conceptslidersloraadaptors}
    Rohit Gandikota, Joanna Materzynska, Tingrui Zhou, Antonio Torralba, and David Bau.
    \newblock Concept sliders: Lora adaptors for precise control in diffusion models, 2023.
    \newblock URL \url{https://arxiv.org/abs/2311.12092}.
    
    \bibitem[Herbrich et~al.(2006)Herbrich, Minka, and Graepel]{TrueSkill}
    Ralf Herbrich, Tom Minka, and Thore Graepel.
    \newblock Trueskill™: a bayesian skill rating system.
    \newblock \emph{Advances in neural information processing systems}, 19, 2006.
    
    \bibitem[Ho and Salimans(2022)]{CFG}
    Jonathan Ho and Tim Salimans.
    \newblock Classifier-free diffusion guidance, 2022.
    \newblock URL \url{https://arxiv.org/abs/2207.12598}.
    
    \bibitem[Ho et~al.(2020)Ho, Jain, and Abbeel]{DDPM}
    Jonathan Ho, Ajay Jain, and Pieter Abbeel.
    \newblock Denoising diffusion probabilistic models, 2020.
    \newblock URL \url{https://arxiv.org/abs/2006.11239}.
    
    \bibitem[Hu et~al.(2021)Hu, Shen, Wallis, Allen-Zhu, Li, Wang, Wang, and Chen]{LoRA}
    Edward~J. Hu, Yelong Shen, Phillip Wallis, Zeyuan Allen-Zhu, Yuanzhi Li, Shean Wang, Lu~Wang, and Weizhu Chen.
    \newblock Lora: Low-rank adaptation of large language models, 2021.
    \newblock URL \url{https://arxiv.org/abs/2106.09685}.
    
    \bibitem[{Ilharco, Gabriel and Wortsman, Mitchell and Wightman, Ross and Gordon, Cade and Carlini, Nicholas and Taori, Rohan and Dave, Achal and Shankar, Vaishaal and Namkoong, Hongseok and Miller, John and Hajishirzi, Hannaneh and Farhadi, Ali and Schmidt, Ludwig}(2021)]{OpenCLIP}
    {Ilharco, Gabriel and Wortsman, Mitchell and Wightman, Ross and Gordon, Cade and Carlini, Nicholas and Taori, Rohan and Dave, Achal and Shankar, Vaishaal and Namkoong, Hongseok and Miller, John and Hajishirzi, Hannaneh and Farhadi, Ali and Schmidt, Ludwig}.
    \newblock Openclip, jul 2021.
    \newblock URL \url{https://doi.org/10.5281/zenodo.5143773}.
    
    \bibitem[Kim et~al.(2022)Kim, Park, Lee, Chung, Lee, and Choo]{AnimeCeleb}
    Kangyeol Kim, Sunghyun Park, Jaeseong Lee, Sunghyo Chung, Junsoo Lee, and Jaegul Choo.
    \newblock Animeceleb: Large-scale animation celebheads dataset for head reenactment, 2022.
    \newblock URL \url{https://arxiv.org/abs/2111.07640}.
    
    \bibitem[Koh et~al.(2024)Koh, Park, and Song]{improvingtextgenerationimages}
    Jun~Young Koh, Sang~Hyun Park, and Joy Song.
    \newblock Improving text generation on images with synthetic captions, 2024.
    \newblock URL \url{https://arxiv.org/abs/2406.00505}.
    
    \bibitem[Krasin et~al.(2017)Krasin, Duerig, Alldrin, Ferrari, Abu-El-Haija, Kuznetsova, Rom, Uijlings, Popov, Veit, Belongie, Gomes, Gupta, Sun, Chechik, Cai, Feng, Narayanan, and Murphy]{openimages}
    Ivan Krasin, Tom Duerig, Neil Alldrin, Vittorio Ferrari, Sami Abu-El-Haija, Alina Kuznetsova, Hassan Rom, Jasper Uijlings, Stefan Popov, Andreas Veit, Serge Belongie, Victor Gomes, Abhinav Gupta, Chen Sun, Gal Chechik, David Cai, Zheyun Feng, Dhyanesh Narayanan, and Kevin Murphy.
    \newblock Openimages: A public dataset for large-scale multi-label and multi-class image classification.
    \newblock \emph{Dataset available from https://github.com/openimages}, 2017.
    
    \bibitem[Krishna et~al.(2016)Krishna, Zhu, Groth, Johnson, Hata, Kravitz, Chen, Kalantidis, Li, Shamma, Bernstein, and Li]{VisualGenome}
    Ranjay Krishna, Yuke Zhu, Oliver Groth, Justin Johnson, Kenji Hata, Joshua Kravitz, Stephanie Chen, Yannis Kalantidis, Li-Jia Li, David~A. Shamma, Michael~S. Bernstein, and Fei-Fei Li.
    \newblock Visual genome: Connecting language and vision using crowdsourced dense image annotations, 2016.
    \newblock URL \url{https://arxiv.org/abs/1602.07332}.
    
    \bibitem[Lee et~al.(2023)Lee, Liu, Ryu, Watkins, Du, Boutilier, Abbeel, Ghavamzadeh, and Gu]{HFT2I}
    Kimin Lee, Hao Liu, Moonkyung Ryu, Olivia Watkins, Yuqing Du, Craig Boutilier, Pieter Abbeel, Mohammad Ghavamzadeh, and Shixiang~Shane Gu.
    \newblock Aligning text-to-image models using human feedback, 2023.
    \newblock URL \url{https://arxiv.org/abs/2302.12192}.
    
    \bibitem[Li et~al.(2024)Li, Zhang, Lin, Xiong, Long, Deng, Zhang, Liu, Huang, Xiao, Chen, He, Li, Li, Zhang, Quan, Lu, Huang, Yuan, Zheng, Li, Zhang, Zhang, Chen, Liu, Fang, Wang, Xue, Tao, Zhu, Liu, Lin, Sun, Li, Wang, Chen, Hu, Xiao, Chen, Liu, Liu, Wang, Yang, Jiang, and Lu]{Hunyuan-DiT}
    Zhimin Li, Jianwei Zhang, Qin Lin, Jiangfeng Xiong, Yanxin Long, Xinchi Deng, Yingfang Zhang, Xingchao Liu, Minbin Huang, Zedong Xiao, Dayou Chen, Jiajun He, Jiahao Li, Wenyue Li, Chen Zhang, Rongwei Quan, Jianxiang Lu, Jiabin Huang, Xiaoyan Yuan, Xiaoxiao Zheng, Yixuan Li, Jihong Zhang, Chao Zhang, Meng Chen, Jie Liu, Zheng Fang, Weiyan Wang, Jinbao Xue, Yangyu Tao, Jianchen Zhu, Kai Liu, Sihuan Lin, Yifu Sun, Yun Li, Dongdong Wang, Mingtao Chen, Zhichao Hu, Xiao Xiao, Yan Chen, Yuhong Liu, Wei Liu, Di~Wang, Yong Yang, Jie Jiang, and Qinglin Lu.
    \newblock Hunyuan-dit: A powerful multi-resolution diffusion transformer with fine-grained chinese understanding, 2024.
    \newblock URL \url{https://arxiv.org/abs/2405.08748}.
    
    \bibitem[Lin et~al.(2024)Lin, Liu, Li, and Yang]{lin2024common}
    Shanchuan Lin, Bingchen Liu, Jiashi Li, and Xiao Yang.
    \newblock Common diffusion noise schedules and sample steps are flawed.
    \newblock In \emph{Proceedings of the IEEE/CVF Winter Conference on Applications of Computer Vision (WACV)}, pages 5392--5399, 2024.
    \newblock \doi{10.1109/WACV57701.2024.00532}.
    \newblock URL \url{https://openaccess.thecvf.com/content/WACV2024/papers/Lin_Common_Diffusion_Noise_Schedules_and_Sample_Steps_Are_Flawed_WACV_2024_paper.pdf}.
    
    \bibitem[Lin et~al.(2015)Lin, Maire, Belongie, Bourdev, Girshick, Hays, Perona, Ramanan, Zitnick, and Dollár]{COCO}
    Tsung-Yi Lin, Michael Maire, Serge Belongie, Lubomir Bourdev, Ross Girshick, James Hays, Pietro Perona, Deva Ramanan, C.~Lawrence Zitnick, and Piotr Dollár.
    \newblock Microsoft coco: Common objects in context, 2015.
    \newblock URL \url{https://arxiv.org/abs/1405.0312}.
    
    \bibitem[Lipman et~al.(2023)Lipman, Chen, Ben-Hamu, Nickel, and Le]{flowmatching}
    Yaron Lipman, Ricky T.~Q. Chen, Heli Ben-Hamu, Maximilian Nickel, and Matt Le.
    \newblock Flow matching for generative modeling, 2023.
    \newblock URL \url{https://arxiv.org/abs/2210.02747}.
    
    \bibitem[Liu et~al.(2024)Liu, Akhgari, Visheratin, Kamko, Xu, Shrirao, Souza, Doshi, and Li]{PGv3}
    Bingchen Liu, Ehsan Akhgari, Alexander Visheratin, Aleks Kamko, Linmiao Xu, Shivam Shrirao, Joao Souza, Suhail Doshi, and Daiqing Li.
    \newblock Playground v3: Improving text-to-image alignment with deep-fusion large language models, 2024.
    \newblock URL \url{https://arxiv.org/abs/2409.10695}.
    
    \bibitem[Loshchilov and Hutter(2017)]{SGDR}
    Ilya Loshchilov and Frank Hutter.
    \newblock Sgdr: Stochastic gradient descent with warm restarts, 2017.
    \newblock URL \url{https://arxiv.org/abs/1608.03983}.
    
    \bibitem[Lu et~al.(2022)Lu, Zhou, Bao, Chen, Li, and Zhu]{DPM}
    Cheng Lu, Yuhao Zhou, Fan Bao, Jianfei Chen, Chongxuan Li, and Jun Zhu.
    \newblock Dpm-solver: A fast ode solver for diffusion probabilistic model sampling in around 10 steps.
    \newblock \emph{arXiv preprint arXiv:2206.00927}, 2022.
    
    \bibitem[Lu et~al.(2023)Lu, Zhou, Bao, Chen, Li, and Zhu]{DPM++}
    Cheng Lu, Yuhao Zhou, Fan Bao, Jianfei Chen, Chongxuan Li, and Jun Zhu.
    \newblock Dpm-solver++: Fast solver for guided sampling of diffusion probabilistic models, 2023.
    \newblock URL \url{https://arxiv.org/abs/2211.01095}.
    
    \bibitem[Ma et~al.(2023)Ma, Zhou, Rao, Zhang, and Sun]{Multicaptionsynthetic}
    Feipeng Ma, Yizhou Zhou, Fengyun Rao, Yueyi Zhang, and Xiaoyan Sun.
    \newblock Image captioning with multi-context synthetic data, 2023.
    \newblock URL \url{https://arxiv.org/abs/2305.18072}.
    
    \bibitem[Meng et~al.(2022)Meng, He, Song, Song, Wu, Zhu, and Ermon]{meng2022sdeditguidedimagesynthesis}
    Chenlin Meng, Yutong He, Yang Song, Jiaming Song, Jiajun Wu, Jun-Yan Zhu, and Stefano Ermon.
    \newblock Sdedit: Guided image synthesis and editing with stochastic differential equations, 2022.
    \newblock URL \url{https://arxiv.org/abs/2108.01073}.
    
    \bibitem[Minka et~al.(2018)Minka, Cleven, and Zaykov]{TrueSkill2}
    Tom Minka, Ryan Cleven, and Yordan Zaykov.
    \newblock Trueskill 2: An improved bayesian skill rating system.
    \newblock \emph{Technical Report}, 2018.
    
    \bibitem[Ning et~al.(2023)Ning, Sangineto, Porrello, Calderara, and Cucchiara]{ning2023inputperturbationreducesexposure}
    Mang Ning, Enver Sangineto, Angelo Porrello, Simone Calderara, and Rita Cucchiara.
    \newblock Input perturbation reduces exposure bias in diffusion models, 2023.
    \newblock URL \url{https://arxiv.org/abs/2301.11706}.
    
    \bibitem[Ossa et~al.(2024)Ossa, Doğan, Birch, and Johnson]{NAI3}
    Juan Ossa, Eren Doğan, Alex Birch, and F.~Johnson.
    \newblock Improvements to sdxl in novelai diffusion v3, 2024.
    \newblock URL \url{https://arxiv.org/abs/2409.15997}.
    
    \bibitem[Park et~al.(2024)Park, Park, Koh, Lee, and Song]{CAT}
    Jae~Wan Park, Sang~Hyun Park, Jun~Young Koh, Junha Lee, and Min Song.
    \newblock Cat: Contrastive adapter training for personalized image generation, 2024.
    \newblock URL \url{https://arxiv.org/abs/2404.07554}.
    
    \bibitem[Peebles and Xie(2023)]{scalableTransformer}
    William Peebles and Saining Xie.
    \newblock Scalable diffusion models with transformers, 2023.
    \newblock URL \url{https://arxiv.org/abs/2212.09748}.
    
    \bibitem[Podell et~al.(2023)Podell, English, Lacey, Blattmann, Dockhorn, Müller, Penna, and Rombach]{SDXL}
    Dustin Podell, Zion English, Kyle Lacey, Andreas Blattmann, Tim Dockhorn, Jonas Müller, Joe Penna, and Robin Rombach.
    \newblock Sdxl: Improving latent diffusion models for high-resolution image synthesis, 2023.
    \newblock URL \url{https://arxiv.org/abs/2307.01952}.
    
    \bibitem[Radford et~al.(2021)Radford, Kim, Hallacy, Ramesh, Goh, Agarwal, Sastry, Askell, Mishkin, Clark, Krueger, and Sutskever]{CLIP}
    Alec Radford, Jong~Wook Kim, Chris Hallacy, Aditya Ramesh, Gabriel Goh, Sandhini Agarwal, Girish Sastry, Amanda Askell, Pamela Mishkin, Jack Clark, Gretchen Krueger, and Ilya Sutskever.
    \newblock Learning transferable visual models from natural language supervision, 2021.
    \newblock URL \url{https://arxiv.org/abs/2103.00020}.
    
    \bibitem[Rafailov et~al.(2024)Rafailov, Sharma, Mitchell, Ermon, Manning, and Finn]{DPO}
    Rafael Rafailov, Archit Sharma, Eric Mitchell, Stefano Ermon, Christopher~D. Manning, and Chelsea Finn.
    \newblock Direct preference optimization: Your language model is secretly a reward model, 2024.
    \newblock URL \url{https://arxiv.org/abs/2305.18290}.
    
    \bibitem[Raffel et~al.(2023)Raffel, Shazeer, Roberts, Lee, Narang, Matena, Zhou, Li, and Liu]{T5}
    Colin Raffel, Noam Shazeer, Adam Roberts, Katherine Lee, Sharan Narang, Michael Matena, Yanqi Zhou, Wei Li, and Peter~J. Liu.
    \newblock Exploring the limits of transfer learning with a unified text-to-text transformer, 2023.
    \newblock URL \url{https://arxiv.org/abs/1910.10683}.
    
    \bibitem[Ramesh et~al.(2022)Ramesh, Dhariwal, Nichol, Chu, and Chen]{DALLE2}
    Aditya Ramesh, Prafulla Dhariwal, Alex Nichol, Casey Chu, and Mark Chen.
    \newblock Hierarchical text-conditional image generation with clip latents, 2022.
    \newblock URL \url{https://arxiv.org/abs/2204.06125}.
    
    \bibitem[Rios et~al.(2021)Rios, Cheng, and Lai]{DAF}
    Edwin~Arkel Rios, Wen-Huang Cheng, and Bo-Cheng Lai.
    \newblock Daf:re: A challenging, crowd-sourced, large-scale, long-tailed dataset for anime character recognition, 2021.
    \newblock URL \url{https://arxiv.org/abs/2101.08674}.
    
    \bibitem[Rombach et~al.(2022)Rombach, Blattmann, Lorenz, Esser, and Ommer]{SD1.5}
    Robin Rombach, Andreas Blattmann, Dominik Lorenz, Patrick Esser, and Björn Ommer.
    \newblock High-resolution image synthesis with latent diffusion models, 2022.
    \newblock URL \url{https://arxiv.org/abs/2112.10752}.
    
    \bibitem[Ronneberger et~al.(2015)Ronneberger, Fischer, and Brox]{UNet}
    Olaf Ronneberger, Philipp Fischer, and Thomas Brox.
    \newblock U-net: Convolutional networks for biomedical image segmentation, 2015.
    \newblock URL \url{https://arxiv.org/abs/1505.04597}.
    
    \bibitem[Ruiz et~al.(2023)Ruiz, Li, Jampani, Pritch, Rubinstein, and Aberman]{Dreambooth}
    Nataniel Ruiz, Yuanzhen Li, Varun Jampani, Yael Pritch, Michael Rubinstein, and Kfir Aberman.
    \newblock Dreambooth: Fine tuning text-to-image diffusion models for subject-driven generation, 2023.
    \newblock URL \url{https://arxiv.org/abs/2208.12242}.
    
    \bibitem[Saharia et~al.(2022)Saharia, Chan, Saxena, Li, Whang, Denton, Ghasemipour, Ayan, Mahdavi, Lopes, Salimans, Ho, Fleet, and Norouzi]{IMAGEN}
    Chitwan Saharia, William Chan, Saurabh Saxena, Lala Li, Jay Whang, Emily Denton, Seyed Kamyar~Seyed Ghasemipour, Burcu~Karagol Ayan, S.~Sara Mahdavi, Rapha~Gontijo Lopes, Tim Salimans, Jonathan Ho, David~J Fleet, and Mohammad Norouzi.
    \newblock Photorealistic text-to-image diffusion models with deep language understanding, 2022.
    \newblock URL \url{https://arxiv.org/abs/2205.11487}.
    
    \bibitem[Schuhmann et~al.(2022)Schuhmann, Beaumont, Vencu, Gordon, Wightman, Cherti, Coombes, Katta, Mullis, Wortsman, Schramowski, Kundurthy, Crowson, Schmidt, Kaczmarczyk, and Jitsev]{LAION-5B}
    Christoph Schuhmann, Romain Beaumont, Richard Vencu, Cade~W Gordon, Ross Wightman, Mehdi Cherti, Theo Coombes, Aarush Katta, Clayton Mullis, Mitchell Wortsman, Patrick Schramowski, Srivatsa~R Kundurthy, Katherine Crowson, Ludwig Schmidt, Robert Kaczmarczyk, and Jenia Jitsev.
    \newblock {LAION}-5b: An open large-scale dataset for training next generation image-text models.
    \newblock In \emph{Thirty-sixth Conference on Neural Information Processing Systems Datasets and Benchmarks Track}, 2022.
    \newblock URL \url{https://openreview.net/forum?id=M3Y74vmsMcY}.
    
    \bibitem[Shallue et~al.(2019)Shallue, Lee, Antognini, Sohl-Dickstein, Frostig, and Dahl]{batchsize1}
    Christopher~J. Shallue, Jaehoon Lee, Joseph Antognini, Jascha Sohl-Dickstein, Roy Frostig, and George~E. Dahl.
    \newblock Measuring the effects of data parallelism on neural network training, 2019.
    \newblock URL \url{https://arxiv.org/abs/1811.03600}.
    
    \bibitem[Shipard et~al.(2023)Shipard, Wiliem, Thanh, Xiang, and Fookes]{diversitydefinitelyneededimproving}
    Jordan Shipard, Arnold Wiliem, Kien~Nguyen Thanh, Wei Xiang, and Clinton Fookes.
    \newblock Diversity is definitely needed: Improving model-agnostic zero-shot classification via stable diffusion, 2023.
    \newblock URL \url{https://arxiv.org/abs/2302.03298}.
    
    \bibitem[Song et~al.(2022)Song, Meng, and Ermon]{DDIM}
    Jiaming Song, Chenlin Meng, and Stefano Ermon.
    \newblock Denoising diffusion implicit models, 2022.
    \newblock URL \url{https://arxiv.org/abs/2010.02502}.
    
    \bibitem[Song et~al.(2021)Song, Sohl-Dickstein, Kingma, Kumar, Ermon, and Poole]{SDE}
    Yang Song, Jascha Sohl-Dickstein, Diederik~P. Kingma, Abhishek Kumar, Stefano Ermon, and Ben Poole.
    \newblock Score-based generative modeling through stochastic differential equations, 2021.
    \newblock URL \url{https://arxiv.org/abs/2011.13456}.
    
    \bibitem[Srinivasan et~al.(2021)Srinivasan, Raman, Chen, Bendersky, and Najork]{WIT}
    Krishna Srinivasan, Karthik Raman, Jiecao Chen, Michael Bendersky, and Marc Najork.
    \newblock Wit: Wikipedia-based image text dataset for multimodal multilingual machine learning.
    \newblock \emph{arXiv preprint arXiv:2103.01913}, 2021.
    
    \bibitem[Su et~al.(2023)Su, Lu, Pan, Murtadha, Wen, and Liu]{Roformer}
    Jianlin Su, Yu~Lu, Shengfeng Pan, Ahmed Murtadha, Bo~Wen, and Yunfeng Liu.
    \newblock Roformer: Enhanced transformer with rotary position embedding, 2023.
    \newblock URL \url{https://arxiv.org/abs/2104.09864}.
    
    \bibitem[Team(2024)]{kolors}
    Kolors Team.
    \newblock Kolors: Effective training of diffusion model for photorealistic text-to-image synthesis.
    \newblock \emph{arXiv preprint}, 2024.
    
    \bibitem[Thrush et~al.(2022)Thrush, Jiang, Bartolo, Singh, Williams, Kiela, and Ross]{limitationCLIP2}
    Tristan Thrush, Ryan Jiang, Max Bartolo, Amanpreet Singh, Adina Williams, Douwe Kiela, and Candace Ross.
    \newblock Winoground: Probing vision and language models for visio-linguistic compositionality, 2022.
    \newblock URL \url{https://arxiv.org/abs/2204.03162}.
    
    \bibitem[Wang et~al.(2024)Wang, Lv, Yu, Hong, Qi, Wang, Ji, Yang, Zhao, Song, Xu, Xu, Li, Dong, Ding, and Tang]{CogVLM}
    Weihan Wang, Qingsong Lv, Wenmeng Yu, Wenyi Hong, Ji~Qi, Yan Wang, Junhui Ji, Zhuoyi Yang, Lei Zhao, Xixuan Song, Jiazheng Xu, Bin Xu, Juanzi Li, Yuxiao Dong, Ming Ding, and Jie Tang.
    \newblock Cogvlm: Visual expert for pretrained language models, 2024.
    \newblock URL \url{https://arxiv.org/abs/2311.03079}.
    
    \bibitem[Wang et~al.(2004)Wang, Bovik, Sheikh, and Simoncelli]{SSIM}
    Zhou Wang, A.C. Bovik, H.R. Sheikh, and E.P. Simoncelli.
    \newblock Image quality assessment: from error visibility to structural similarity.
    \newblock \emph{IEEE Transactions on Image Processing}, 13\penalty0 (4):\penalty0 600--612, 2004.
    \newblock \doi{10.1109/TIP.2003.819861}.
    
    \bibitem[Yang et~al.(2023)Yang, Tao, Lyu, Ge, Chen, Li, Shen, Zhu, and Li]{D3PO}
    Kai Yang, Jian Tao, Jiafei Lyu, Chunjiang Ge, Jiaxin Chen, Qimai Li, Weihan Shen, Xiaolong Zhu, and Xiu Li.
    \newblock Using human feedback to fine-tune diffusion models without any reward model.
    \newblock \emph{arXiv preprint arXiv:2311.13231}, 2023.
    
    \bibitem[Ye et~al.(2023)Ye, Zhang, Liu, Han, and Yang]{IPAdapter}
    Hu~Ye, Jun Zhang, Sibo Liu, Xiao Han, and Wei Yang.
    \newblock Ip-adapter: Text compatible image prompt adapter for text-to-image diffusion models, 2023.
    \newblock URL \url{https://arxiv.org/abs/2308.06721}.
    
    \bibitem[Yeh(2024)]{yeh2024tipo}
    Shih-Ying Yeh.
    \newblock Tipo: Text to image with text presampling for prompt optimization, 9 2024.
    \newblock Technical report available at \url{https://hackmd.io/@KBlueLeaf/BJULOQBR0}. Model available at \url{https://huggingface.co/KBlueLeaf/TIPO-500M}. Source code available at \url{https://github.com/KohakuBlueleaf/KGen}.
    
    \bibitem[Yeh et~al.(2024)Yeh, Hsieh, Gao, Yang, Oh, and Gong]{lycoris}
    Shih-Ying Yeh, Yu-Guan Hsieh, Zhidong Gao, Bernard B~W Yang, Giyeong Oh, and Yanmin Gong.
    \newblock Navigating text-to-image customization: From lycoris fine-tuning to model evaluation, 2024.
    \newblock URL \url{https://arxiv.org/abs/2309.14859}.
    
    \bibitem[Young et~al.(2014)Young, Lai, Hodosh, and Hockenmaier]{Flickr30k}
    Peter Young, Alice Lai, Micah Hodosh, and Julia Hockenmaier.
    \newblock From image descriptions to visual denotations: New similarity metrics for semantic inference over event descriptions.
    \newblock \emph{Transactions of the Association for Computational Linguistics}, 2:\penalty0 67--78, 2014.
    \newblock \doi{10.1162/tacl_a_00166}.
    \newblock URL \url{https://aclanthology.org/Q14-1006}.
    
    \bibitem[Yu et~al.(2024)Yu, Shen, Huang, Li, and Zhao]{yu2024unmaskingbiasdiffusionmodel}
    Hu~Yu, Li~Shen, Jie Huang, Hongsheng Li, and Feng Zhao.
    \newblock Unmasking bias in diffusion model training, 2024.
    \newblock URL \url{https://arxiv.org/abs/2310.08442}.
    
    \bibitem[Yuksekgonul et~al.(2023)Yuksekgonul, Bianchi, Kalluri, Jurafsky, and Zou]{limitationCLIP1}
    Mert Yuksekgonul, Federico Bianchi, Pratyusha Kalluri, Dan Jurafsky, and James Zou.
    \newblock When and why vision-language models behave like bags-of-words, and what to do about it?, 2023.
    \newblock URL \url{https://arxiv.org/abs/2210.01936}.
    
    \bibitem[Zhang et~al.(2019)Zhang, Li, Nado, Martens, Sachdeva, Dahl, Shallue, and Grosse]{batchsize2}
    Guodong Zhang, Lala Li, Zachary Nado, James Martens, Sushant Sachdeva, George~E. Dahl, Christopher~J. Shallue, and Roger Grosse.
    \newblock Which algorithmic choices matter at which batch sizes? insights from a noisy quadratic model, 2019.
    \newblock URL \url{https://arxiv.org/abs/1907.04164}.
    
    \bibitem[Zhang et~al.(2023)Zhang, Rao, and Agrawala]{ControlNet}
    Lvmin Zhang, Anyi Rao, and Maneesh Agrawala.
    \newblock Adding conditional control to text-to-image diffusion models, 2023.
    \newblock URL \url{https://arxiv.org/abs/2302.05543}.
    
    \bibitem[Zhang et~al.(2018)Zhang, Isola, Efros, Shechtman, and Wang]{LPIPS}
    Richard Zhang, Phillip Isola, Alexei~A. Efros, Eli Shechtman, and Oliver Wang.
    \newblock The unreasonable effectiveness of deep features as a perceptual metric, 2018.
    \newblock URL \url{https://arxiv.org/abs/1801.03924}.
    
    \bibitem[Zheng et~al.(2023)Zheng, Chiang, Sheng, Zhuang, Wu, Zhuang, Lin, Li, Li, Xing, Zhang, Gonzalez, and Stoica]{LLMasaJudge}
    Lianmin Zheng, Wei-Lin Chiang, Ying Sheng, Siyuan Zhuang, Zhanghao Wu, Yonghao Zhuang, Zi~Lin, Zhuohan Li, Dacheng Li, Eric~P. Xing, Hao Zhang, Joseph~E. Gonzalez, and Ion Stoica.
    \newblock Judging llm-as-a-judge with mt-bench and chatbot arena, 2023.
    \newblock URL \url{https://arxiv.org/abs/2306.05685}.
    
    \bibitem[Zheng et~al.(2020)Zheng, Zhao, Ren, Yan, Lu, Liu, and Li]{CartoonFaceRecognition}
    Yi~Zheng, Yifan Zhao, Mengyuan Ren, He~Yan, Xiangju Lu, Junhui Liu, and Jia Li.
    \newblock Cartoon face recognition: A benchmark dataset.
    \newblock In \emph{Proceedings of the 28th ACM International Conference on Multimedia}, MM '20, page 2264–2272, New York, NY, USA, 2020. Association for Computing Machinery.
    \newblock ISBN 9781450379885.
    \newblock \doi{10.1145/3394171.3413726}.
    \newblock URL \url{https://doi.org/10.1145/3394171.3413726}.
    
    \end{thebibliography}
    
}
\end{document}